\RequirePackage{fix-cm} 
\documentclass[twocolumn]{svjour3}          
\smartqed  

\usepackage{moreverb,url}
\usepackage[colorlinks = true,
			linkcolor = blue,
			urlcolor  = blue,
			citecolor = blue,
			anchorcolor = blue]{hyperref}

\usepackage{dblfloatfix} 
\usepackage{subfigure} 
\usepackage{graphicx} 
\usepackage{algorithmic} 
\usepackage{algorithm} 
\usepackage{verbatim}
\usepackage{amsfonts} 
\usepackage{amssymb} 
\usepackage{amsmath}

\usepackage{graphicx} 

\usepackage{subfigure} 

\usepackage{caption} 

\usepackage{amssymb}
\usepackage{pifont}

\newcommand{\cmark}{\ding{51}}%
\newcommand{\bcmark}{\ding{52}}%
\newcommand{\xmark}{\ding{55}}%

\usepackage{url} 
\usepackage{colortbl} 

\definecolor{lineGray}{cmyk}{0,0,0,.1} 
\definecolor{rowGreen}{cmyk}{.07,0,.12,0} 

\usepackage[misc]{ifsym}
\begin{document} 
	
	\title{Cooperative Multi-UAV Coverage Mission Planning Platform for Remote Sensing Applications} 
	
	
	\author{Savvas D. Apostolidis$^{1,2}$ \and Pavlos Ch. Kapoutsis$^{3}$ \and Athanasios Ch. Kapoutsis$^{2}$ \and Elias B. Kosmatopoulos$^{1,2}$} 
	
	\authorrunning{Savvas D. Apostolidis et al.} 
	
	\institute{\Letter { } S. D. Apostolidis \at 
		sapostol@ee.duth.gr      
		\and
		\Letter { } P. Ch. Kapoutsis \at 
		pavloskapoutsis@gmail.com      
		\and 
		\Letter { } A. Ch. Kapoutsis \at 
		athakapo@iti.gr      
		\and 
		\Letter { } E. B. Kosmatopoulos \at 
		kosmatop@iti.gr\\ \\ 
		$^{1}$ Department of Electrical and Computer Engineering, Democritus University of Thrace, Xanthi, Greece\\ 
		$^{2}$ Information Technologies Institute, The Centre for Research \& Technology, Hellas, Thessaloniki, Greece\\ 
		$^{3}$ School of Electrical and Computer Engineering, National Technical University, Athens, Greece 
	} 
	
	\date{Received: date / Accepted: date} 

	\maketitle 
	
	\begin{abstract} 
		This paper proposes a novel mission planning platform, capable of efficiently deploying a team of UAVs to cover complex-shaped areas, in various remote sensing applications. Under the hood lies a novel optimization scheme for grid-based methods, utilizing Simulated Annealing algorithm, that significantly increases the achieved percentage of coverage and improves the qualitative features of the generated paths. Extensive simulated evaluation in comparison with a state-of-the-art alternative methodology, for coverage path planning (CPP) operations, establishes the performance gains in terms of achieved coverage and overall duration of the generated missions. On top of that, DARP algorithm is employed to allocate sub-tasks to each member of the swarm, taking into account each UAV's sensing and operational capabilities, their initial positions and any no-fly-zones possibly defined inside the operational area. This feature is of paramount importance in real-life applications, as it has the potential to achieve tremendous performance improvements in terms of time demanded to complete a mission, while at the same time it unlocks a wide new range of applications, that was previously not feasible due to the limited battery life of UAVs. In order to investigate the actual efficiency gains that are introduced by the multi-UAV utilization, a simulated study is performed as well. All of these capabilities are packed inside an end-to-end platform that eases the utilization of UAVs' swarms in remote sensing applications. Its versatility is demonstrated via two different real-life applications: (i) a photogrametry for precision agriculture and (ii) an indicative search and rescue for first responders missions, that were performed utilizing a swarm of commercial UAVs. An implementation of the mCPP methodology introduced in this work, as well as a link for a demonstrative video and a link for a fully functional, on-line hosted instance of the presented platform can be found here: \url{https://github.com/savvas-ap/mCPP-optimized-DARP}.
	\end{abstract}

\section{Introduction}
\label{sec:intro}

For the past few years UAVs have attracted the interest of many enterprise fields, becoming a powerful tool for professionals to acquire data in a fast and efficient way. Some of the fields that already exploit the use of UAVs for acquiring data are: precision agriculture \cite{deng2018uav}, \cite{maes2019perspectives}, \cite{comba2018unsupervised}, infrastructure inspection \cite{ham2016visual}, \cite{al2017vbii}, exploration \cite{renzaglia2020common}, search and rescue \cite{sun2016camera}, \cite{qi2016search} and monitoring \cite{koutras2020autonomous}, \cite{kapoutsis2019distributed}. As a result of this all the more growing interest on this direction, there are now available a lot of commercial UAVs, specialized for enterprise use. To cope with all these aerial devices, a range of software platforms that automate the procedure of flying, acquiring and processing data has been developed as well.

Integrated end-to-end platforms, that offer an automated coverage procedure with UAVs, facilitate data gathering and make remote sensing with UAVs accessible to a wide range of professionals and hobbyists. Thanks to such platforms, common, relatively cheap, commercial UAVs can be turned to powerful tools for inspection, mapping, monitoring and search and rescue operations. The introduction of UAVs in the aforementioned operations has started to radically change many professions, making the conduction of different tasks easier, safer and more efficient. Despite the advances that were introduced by these platforms, there is still way to go, to make them more efficient and make the most out of the capabilities offered by UAVs.

For a wide range of remote sensing operations with UAVs, the objective of a mission is to completely cover a user-defined area and gather data. In general, the path generation problem, with the objective to completely cover a region of interest (ROI), is usually referred as Coverage Path Planning (CPP) \cite{Galceran_2013} in literature. The CPP problem can be stated as follows: given a \textit{defined ROI} and \textit{specific coverage capabilities of the sensors} used, \textit{provide paths}, for one or more mobile robots, in order to \textit{completely cover the ROI}, taking into account \textit{spatial and motion constraints}. 

This paper addresses the problem of designing multi-UAV missions to cooperatively cover a single region of interest, oriented to the real world remote sensing application. Such a problem set-up contains both a NP-hard variation \cite{rekleitis2008efficient}  of the original CPP, called mCPP (multi-robot Coverage Path Planning), and the challenging task of translating grid-based solutions to real-world navigation paths. The following subsection summarizes the related works that have been developed around these axes.

\subsection{Related Works}
\label{subsec:related}

Given that the CPP problem provides a powerful tool for a wide range of domains and applications, it has attracted a great interest over the years and there are many research works about it. The state-of-the-art on CPP methods for robotics in general is summarized in \cite{Galceran_2013} and works tailored to UAVs are presented in \cite{Cabreira_2019}. CPP methods can be divided in categories according to different criteria that focus on specific aspects of the problem. The most important of them are:

\begin{enumerate}
	\item \textit{cellular decomposition}/\textit{grid-based} methods, based on how the method discretizes the ROI in order to calculate paths,
	\item \textit{single}/\textit{multi-robot} methods, regarding the number of robots that can participate in a mission,
	\item \textit{on-line}/\textit{off-line} methods, based on whether or not an on-going mission can be adjusted by getting real-time feedback from the operational area
	\item and \textit{energy-aware} or \textit{not}, based on whether some actions are taken in order to provide energy-efficient paths (e.g., reduce turns, avoid redundant scans etc.).
\end{enumerate}
In addition to them, there are also different categories regarding the patterns of paths that come out. The most dominant of them are:
\begin{enumerate}
	\item the ones using \textit{lawnmower patterns}, that are simple back and forth patterns with a defined distance between them,
	\item \textit{Spanning Tree Coverage} (STC) \cite{gabriely2001spanning} \textit{patterns}, where a Minimum Spanning Tree (MST) \cite{Gower_1969} is constructed at first and a path that circumnavigates the MST is generated afterwards,
	\item and methods using \textit{spiral patterns}, navigating from a sided starting point to a centrally-placed ending point of the ROI, or reverse.
\end{enumerate}

Table \ref{table:RelWorkOvrv} presents an overview of the following related works, comparing the support and implementation of some major features for a CPP method, in a succinct way.

\begin{table*}[!t]
	\def\arraystretch{1.3}
	\centering
	\resizebox{\textwidth}{!}{\begin{tabular}{ c | c  c  c  c  c  c } 
			\hline
			\textbf{Method} & \begin{tabular}{@{}c@{}}\textbf{Real-life} \\ \textbf{experiments}\end{tabular}  & \begin{tabular}{@{}c@{}} \textbf{Multi-robot} \\ \textbf{support} \end{tabular}  & \textbf{Task allocation} & \textbf{Integrated platform} & \textbf{Supported shape of ROI} & \begin{tabular}{@{}c@{}} \textbf{Support of} \\ \textbf{NFZ/Obstacles} \end{tabular} \\ 
			\hline\hline\arrayrulecolor{lineGray}\hline
			\cite{choset1998coverage}, \cite{li2011coverage} & \xmark & \xmark & - & \xmark & Convex \& Concave polygons & \cmark \\

			\hline
			\cite{coombes2017boustrophedon} & \xmark & \xmark & - & \xmark & Convex \& Concave polygons & \cmark* \\
			
			\hline
			\cite{lewis2017semi} & \xmark & \xmark & - & \xmark & \begin{tabular}{@{}c@{}} Convex \& Concave \\ Non-connective ROIs \end{tabular} & \cmark \\
			
			\hline
			\cite{bahnemann2019revisiting} & \cmark & \xmark & - & ROS implementation & Convex \& Concave polygons & \cmark \\
			
			\hline
			\cite{di2016coverage} & \cmark & \xmark & - & \xmark & Convex polygons & \xmark \\
			
			\hline
			\cite{maza2007multiple} & \xmark & \cmark & \begin{tabular}{@{}c@{}} Exclusive \& Proportional \end{tabular} & \xmark & Convex polygons & \xmark \\
			
			\hline
			\cite{gabriely2001spanning}, \cite{guruprasad2019x} & \xmark & \xmark & - & \xmark & Grid with obstacle cells & \cmark - obstacle cells \\
			
			\hline
			\cite{hazon2005redundancy}, \cite{agmon2006constructing} & \xmark & \cmark & \begin{tabular}{@{}c@{}} Single path allocation \\ based on initial positions \end{tabular} & \xmark & Grid with obstacle cells & \cmark - obstacle cells\\
			
			\hline
			\cite{zheng2005multi} & \xmark & \cmark & \begin{tabular}{@{}c@{}} Non-exclusive \\ single path allocation \end{tabular} & \xmark & Grid with obstacle cells & \cmark - obstacle cells\\
			
			\hline
			\cite{kapoutsis2017darp} & \xmark & \cmark & \begin{tabular}{@{}c@{}} Exclusive \\ Equal \end{tabular} & \xmark & Grid with obstacle cells & \cmark - obstacle cells\\
			
			\hline
			\cite{huang2020multi} & \xmark & \cmark & Exclusive \& Equal & \xmark & Multi-scale grid with obstacles & \cmark - obstacle cells \\
			
			\hline
			\cite{barrientos2011aerial} & \cmark & \cmark & \begin{tabular}{@{}c@{}} Minimum overlap \& \\ Proportional after negotiation \end{tabular} & \cmark & Convex \& Concave polygons & \cmark \\
			
			\hline
			\cite{shah2020multidrone} & \cmark & \cmark & Minimum overlap & \xmark & Convex \& Concave polygons & \cmark** \\
			
			\rowcolor{rowGreen}
			\hline
			\textbf{Proposed} & \bcmark & \bcmark & \begin{tabular}{@{}c@{}} \textbf{Exclusive \&} \\ \textbf{Equal/Proportional} \end{tabular}  & \bcmark &  \textbf{Convex \& Concave polygons} & \bcmark \\
			\arrayrulecolor{black}\hline
	\end{tabular}}
	\caption{Overview of Related Works}
	
	\begin{flushleft}{*Not explicitly mentioned in the paper, but this feature is supported inherently by the method used.}\end{flushleft}
	\begin{flushleft}{**In the paper NFZs are only included between the home position of the UAVs and the first mission's waypoint, but not inside the ROI.}\end{flushleft}
	\label{table:RelWorkOvrv}
\end{table*}

One of the most popular approaches in the CPP problem is the boustrophedon cellular decomposition \cite{choset1998coverage}. This method generates lawnmower patterns and was initially designed in an attempt to minimize the number of excess lengthwise motions of the robot. An important advantage of the method, making it so popular, is the fact that it can be used for complex-shaped ROIs with no-fly-zone (NFZs) inside them, providing very high percentage of coverage. Many works so far are based on it, exploiting certain aspects, improving and extending this approach. Some recent examples that exploit boustrophedon cellular decomposition are: \cite{coombes2017boustrophedon} where the authors present a method to define and calculate flight times in a boustrophedon aerial survey coverage path in wind, for precision agriculture applications and \cite{lewis2017semi} which is introduced by the authors as a semi-boustrophedon method, because, while they use boustrophedon cellular decomposition, they do not limit coverage strictly to boustrophedon paths in order to reduce run-time, while maintaining near-optimal path length. This family of approaches is the most commonly used for real-life operations, at the moment, because of its incorporated advantages and its out-of-the-box adaptability to complex-shaped ROIs. However, it is not the most energy-efficient one, as it many times includes redundant movements that do not contribute to the scanning procedure (e.g., from the starting/ending points to the home position of the vehicles) and multiple times re visitation of certain points, to ensure complete coverage.

In \cite{li2011coverage} exact cellular decomposition method is used in order to provide turn-minimized paths for UAVs inside a polygon ROI. It is shown that this kind of paths is more efficient, as turns are time and energy consuming for UAV missions. In this work, convex or concave polygon regions are decomposed to convex sub-regions. After that, the optimal path orientation for every sub-region is calculated and sub-regions get connected again, to provide the overall path. While this work becomes a lot more energy-efficient by significantly reducing the number of turns, it still inherits the issues mentioned in the previous paragraph.

\cite{bahnemann2019revisiting} presents one of the most recent works on the CPP problem. In this work, boustrophedon CPP \cite{choset1998coverage} is extended in order to find the optimal sweeping direction for every sub-region of a user-defined ROI, including obstacles and NFZs. This method exploits already existing approaches on combination and manages to attract the interest as one of the most appropriate for real-life use implementations of coverage mission planners. It was developed for low-altitude flights of UAVs, and is available in an open source robot operating system (ROS) implementation. The mission planner presented in this work, outperforms in terms of time-efficiency and ability to handle NFZs many existing, commercial or not, planners available at the moment (such as Ardupilot Mission Planner\footnote{\url{http://ardupilot.org/planner}}, Pix4DCapture\footnote{\url{https://www.pix4d.com/product/pix4dcapture}}, Drone Harmony\footnote{\url{https://droneharmony.com}} and DJI Flight Planner\footnote{\url{https://www.djiflightplanner.com}}). A point that this work falls short, is that it cannot handle multiple UAVs missions.

In \cite{di2016coverage} a CPP method tailored to UAVs, having in mind real-life problems and limitations faced in photogrammetry applications, is presented. In this work an energy model derived from real measurements is introduced, that is used to provide a power consumption estimation for the UAV. After that, an energy aware algorithm is proposed, that provides path with reduced energy consumption, while taking into account image resolution constraints. On the top of that, two safety mechanisms are presented. The first one, executed off-line, checks if the available energy is enough to complete a calculated mission, while the second, on-line mechanism, reassures safe return for the UAV to the take-off location, in case the remaining battery energy is not enough to complete a mission and return to home. While this work manages to efficiently face some of the challenges that are met in real-life operations, it still remains limited by the UAV's battery life, as it cannot handle multiple UAVs in the same mission.

\cite{maza2007multiple} presents a mCPP method for UAVs. The presented algorithm divides the ROI into smaller, exclusive sub-regions for every UAV and computes a turn-minimized back and forth path for every UAV, inside their exclusive sub-regions. Although this work presents an interesting approach, it can only handle convex polygon areas and does not support NFZs inside them.

Another popular approach in the CPP domain is the STC method \cite{gabriely2001spanning}. The proposed algorithm sub-divides the operational area into disjoint cells, which are used for the generation of a spanning tree. As a next step, a path is generated around this spanning tree and the method guarantees that every point of the grid is covered precisely once. This circumnavigating path has the advantage of avoiding redundant movements of the robot, that do not contribute in the scanning procedure. A mission starts and ends in the same position and every cell is visited only once. This way, the paths generated are highly appropriate for energy-efficient applications. Defining obstacles inside a ROI is also supported, by removing connections of nodes that represent them, during the MST generation. However, being a grid based method, it can turn out to be inefficient for covering complex-shaped ROIs, because of common discretization issues \cite{choset2001coverage}, \cite{ghaemi2009evaluation}. As in boustrophedon approach, there are also a lot of works based on STC, exploiting certain aspects and extending this approach, such as \cite{guruprasad2019x} where the authors attempt to extended spanning tree based coverage (X-STC) algorithm, in order to cover even the partially occupied cells, which helps in the better coverage of complex-shaped polygon regions.

In addition, \cite{hazon2005redundancy}, \cite{agmon2006constructing}, \cite{zheng2005multi} present some approaches expanding the STC for multi-robot use. \cite{hazon2005redundancy} is the first approach utilizing STC for the mCPP problem, and \cite{agmon2006constructing}, \cite{zheng2005multi} introduce ideas to solve it more efficiently. In all of these works, a single MST is constructed and the generated path around it gets allocated proportionally to the multiple robots. This approach for multi-robot coverage though, is not the most efficient one, as the portion of the area that will be allocated to each robot is dependent on their initial positions, ignoring the operational capabilities of each member of the team. Moreover, there are cases where the same cell of the grid has to be scanned multiple times by different members. Finally, it should be noted that none of these works include real-life applications.

\cite{kapoutsis2017darp} is also a STC-based approach that deals with the mCPP problem, for grids with prior-defined obstacles. However, this work approaches the problem from a completely different point of view, compared to \cite{hazon2005redundancy},\cite{zheng2005multi}, \cite{agmon2006constructing}. Specifically, \cite{kapoutsis2017darp} deals with the multiple vehicles, by reducing the mCPP problem to, as-many-as-the-number-of-robots single-robot CPP problems. This is achieved by the division of the overall ROI to equal, exclusive sub-regions for each vehicle, based on their initial positions. The proposed methodology promises to always converge to a solution, at least in cases where one exists and also offers a significantly reduced complexity compared to other state-of-the-art-approaches of that time. While \cite{kapoutsis2017darp} presents a really promising approach, it does not contain any real-life experiments that would validate its real-world efficiency.

Based on a different approach, where an overall region is divided in exclusive sub-regions and then allocated to a group of UAVs, \cite{huang2020multi} identifies the gap of coverage path planning in complex geographic environments and proposes an STC-based mCPP method, called MCCP-MLCT. This method constructs a grid with multiple scales (different sizes of cells), based on the complexity and topology of the environment that will be applied. The key idea is that different parts of a ROI that should be covered, should be faced in different ways, e.g., in a search and rescue task, more complex parts of the region should be scanned in more detail, so smaller-sized cells should be used for the construction of MSTs. This work introduces an idea that makes it highly appropriate for specific types of missions, where the detail of collected data should vary depending on the environment characteristics. While an application of the CPP method is presented for a real remote sensing image, this work does not include any real-life experiments.

An integrated platform for managing missions for precision agriculture, utilizing multiple UAVs, is presented in \cite{barrientos2011aerial}. This work contains a complete top to down approach, as it describes a tool containing everything from a graphic user interface (GUI) on a ground station, to a robust flight controller for the UAVs, to improve their maneuverability. The main contribution here is the practical experimentation with an integrated tool, however a new one-phase task partitioning manager is presented and a CPP method that minimizes turns and times of revisiting a specific cell is used as well. It is worth pointing out that this setup is able to handle non-convex ROIs with NFZs inside them as well. Even if there are available more energy-efficient approaches for the coverage paths generation, this work is one of the most complete, in terms of integration level for the presented platform.

A recent work, including real-life, multi-UAV operations, under very harsh weather conditions, is presented in \cite{shah2020multidrone}. The authors have developed a mCPP methodology based on a completely different approach, in order to face a set of challenges that are faced in surveys of penguin colonies in Antarctica. The main objective of this work is to minimize the time duration of the missions, a factor that is critical for this type of operations. The proposed methodology, named Path Optimization for Population Counting with Overhead Robotic Networks (POPCORN), generates paths for multiple UAVs by solving a series of satisfiability modulo theory (SMT) instances (high-level objectives are defined at this step, e.g., same or almost same starting and ending positions for the paths). By iteratively reducing the maximum allowed path's length, this methodology manages to calculate the shortest possible paths to completely cover a ROI. The backtracking of paths and the places revisited by multiple UAVs is minimal, but still not null, however in missions where detecting moving targets is desired, this could not be considered as a disadvantage.

From the works presented above, many suffer in handling concave ROIs and NFZs, not all of them are capable of utilizing multiple robots and only \cite{bahnemann2019revisiting}, \cite{barrientos2011aerial} are presented as an integrated platform for real-life use. In addition, especially for the STC-based methods, none of them presents real-life experiments.

On the other hand, regarding the majority of the software platforms for mission planning, that are currently available, they mostly use simple back and forth patterns. Some of them, also allow the user to select a rotation angle for the generated path. While these patterns are efficient in terms of computation and provide a very high percentage of coverage for a defined ROI, they usually suffer in managing efficiently mission's resources. Most of high-level, ready-to-use mission planners available, do not support multi-robot coverage path generation, or the definition of NFZs inside the user-defined ROI as well.

Overall, in the CPP domain there is a clear gap between the integrated end-to-end platforms that can easily deploy coverage missions in real-world conditions and the state-of-the-art, literature approaches that incorporate novel, efficient functionalities. On the one hand, most of the integrated platforms manage to successfully deploy coverage missions, easing the whole procedure by offering user-friendly environments and robust implementations. On the other hand, while various literature approaches manage to solve very efficiently specific problems faced in the CPP domain, most of them are not developed with the intention of getting applied in real-life operations and thus cannot easily be used in real-world conditions.

\subsection{Contributions}	
\label{sec:problemForm}	

\begin{figure*}[!t]
	\captionsetup{justification=centering,margin=2cm}
	\includegraphics[width=\linewidth]{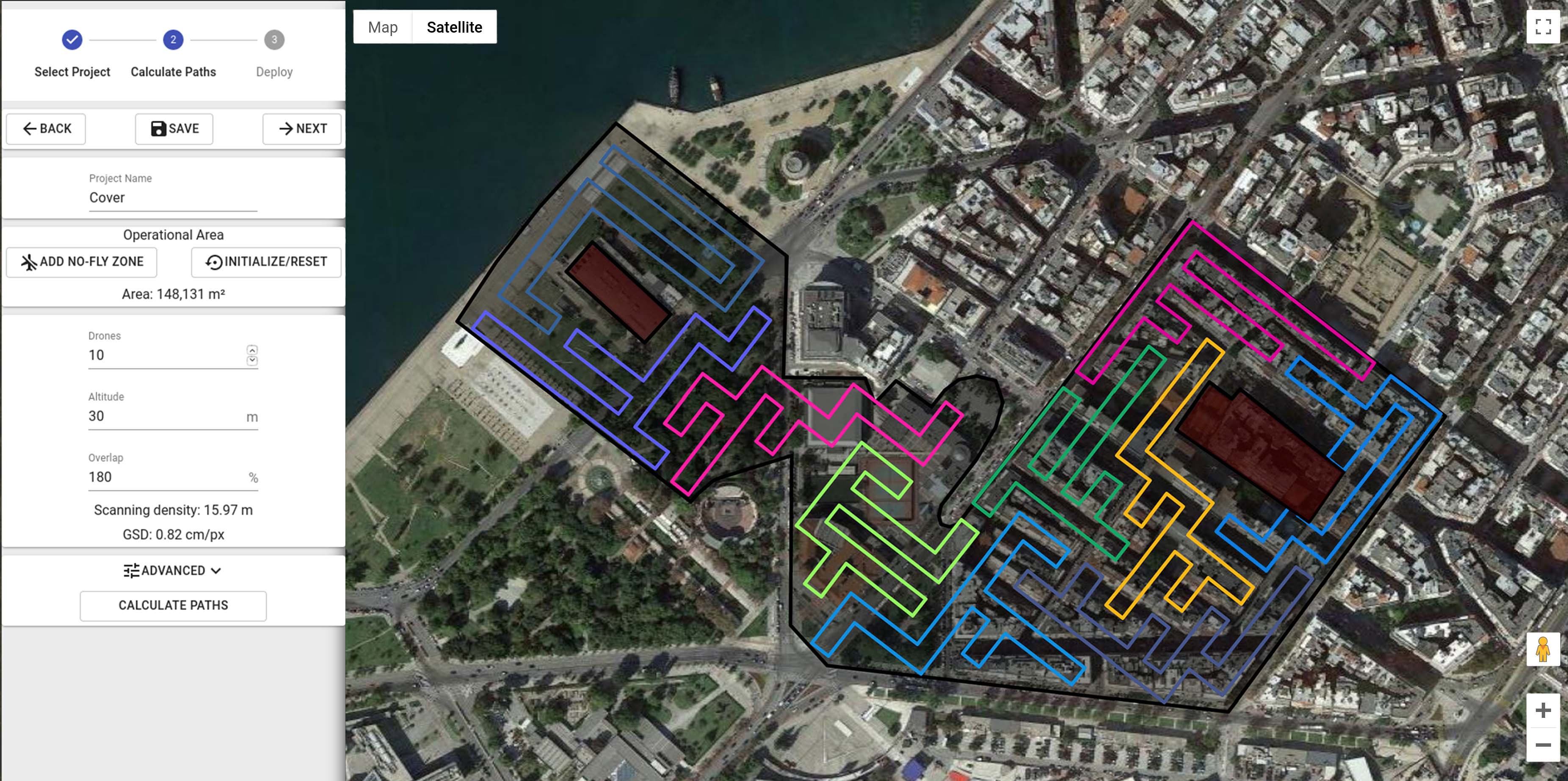}
	\caption{Coverage mission with the proposed mCPP platform}
	\centering{The objective of this mission is to completely cover a non-convex, complex-shaped polygon ROI, that includes two separate NFZs inside it (marked with light-red color), involving 10 UAVs, that all contribute equally to the scanning procedure.}
	\label{fig:cover}
\end{figure*}

In this work is presented a high-level, user-friendly platform, developed to perform multi-UAV coverage missions, optimized for real-life use and integrating energy-aware features. The proposed platform, as depicted in the last row of table \ref{table:RelWorkOvrv}, intends to provide a tool that can ease the planning and execution of coverage missions with multiple commercial, or custom UAVs, while at the same time provide state-of-the-art performance guarantees. For this purpose:

\begin{enumerate}
	\item A previous work from our lab \cite{kapoutsis2017darp} for multi-robot task allocation and coverage path planning, is adjusted and applied for real-life use with UAVs. Specifically, \cite{kapoutsis2017darp} is extended to also support proportional allocation of the ROI to the vehicles, and great effort has been given to address some of the efficiency issues that arise, when applying grid-based CPP methodologies in real-life operations.
	\item A novel optimization procedure that significantly increases the percentage of coverage and improves the qualitative features of the generated paths, for STC-based methods, is implemented and evaluated.
	\item An approach to further reduce the path turns and increase the energy-awareness of the proposed method is applied.
	\item A comparison with a recent, powerful in terms of complete coverage and support of complex-shaped ROIs, state-of-the-art approach is performed, in order to quantify and evaluate the achieved performance of the resulting path planning scheme.
	\item A multi-robot marginal utility study is performed, in order to investigate the actual efficiency gains that are introduced by the utilization of multiple UAVs, in coverage missions.
	\item A new, end-to-end interface is developed, to provide a user-friendly interaction layer to create, perform and manage missions.
	\item A robust communication interface to integrate with ease commercial UAVs to the platform is developed.
	\item Two real-life experiments are performed, to demonstrate the applicability and robustness of the developed platform in some indicative use cases.
\end{enumerate}

The result of this work is a powerful multi-UAV mission planner, that supports no-fly-zones and obstacles, manages wisely the operational resources, supports both fair and proportional task allocation for the UAVs, depending on their operational capabilities, is easy to use and is compatible with almost any commercial UAV, not demanding special and expensive equipment. Based on the categories that were mentioned in \ref{subsec:related}, the proposed CPP methodology can be described as a grid-based, off-line, STC-pattern, multi-robot, energy aware solution. Figure  \ref{fig:cover} shows a mission example from the proposed platform, involving 10 UAVs, in a non-convex, complex-shaped polygon ROI with 2 NFZs and the objective to completely cover the defined region, utilizing all UAVs equally.

\subsection{Paper outline}
\label{subsec:outline}
The work in this paper is organized as follows: section \ref{sec:proposedSolution} presents the task allocation and coverage path planning methods, along with the optimization procedure that is implemented, in section \ref{sec:simulation} is performed a simulated evaluation of the proposed methodology, to assess the optimization procedure and the overall CPP efficiency, as well as a simulated multi-robot study, section \ref{sec:platform} presents an overview of the overall software platform, section \ref{sec:experiments} includes two different applications of the proposed platform in real-life experiments, along with visual results, and finally in section \ref{sec:conclusions} this work is summarized and future plans to improve it are discussed.

\begin{figure*}[!t]
	\captionsetup{justification=centering,margin=2cm}
	\includegraphics[width=0.87\paperwidth]{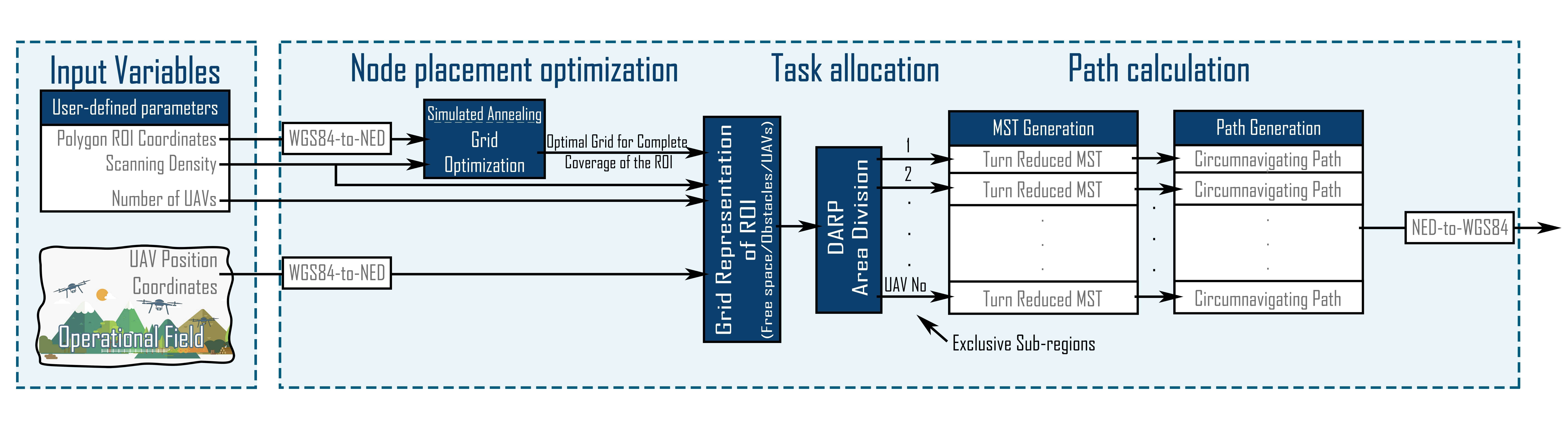}
	\caption{Data-flow of the mCPP method}
	\label{ROI2paths}
\end{figure*}

\section{Multi-robot Coverage Path Planning}
\label{sec:proposedSolution}

In order to cope with the challenges that are faced in real-life, multi-UAV, coverage missions, it is essential to utilize an approach that (i) provides safe and efficient paths, (ii) respects the operational capabilities and motion-limitations of the vehicles and (iii) ensures a high percentage of coverage, so as to gather sufficient information for any operational area. To efficiently address these issues, an optimized for real-life use mCPP method, based on a previous work of our lab \cite{kapoutsis2017darp}, is proposed. In this section, the overall proposed method is described in detail, along with the optimizations that take place to make it appropriate and efficient for the multi-UAV domain.

The proposed mCPP method deals with the problem of designing paths for a team of UAVs, participating in a mission, given:
\begin{itemize}
	\item a \textit{user-defined polygon ROI}, probably including smaller polygon sub-regions representing NFZs/obstacles ($O$), formatted in the WGS84 (World Geodetic System - 1984) coordinate system \cite{cai2011coordinate}, as follows:
	\begin{equation}
	ROI = \{(lat_1, lon_1), ... (lat_n, lon_n)\}
	\label{eq:polygon}
	\end{equation}
	\begin{equation}
	\begin{split}
	O = \{\{(lat_{ob_1|1}, lon_{ob_1|1}), ... (lat_{ob_1|n_1}, lon_{ob_1|n_1})\},\\
	... \{(lat_{ob_m|1}, lon_{ob_m|1}), ... (lat_{ob_m|n_m}, lon_{ob_m|n_m})\}\}
	\end{split}
	\label{eq:obstacles}
	\end{equation}
	where $n$ is the count of polygon vertices, $ob_i$ $(i = 0, ... m)$ stands for the obstacle number (pointer), $n_i$ $(i = 0, ... m)$ is the count of vertices for each NFZ region and $m$ the count of NFZs/obstacles.
	\item the \textit{desired scanning density} ($d_s$) for the mission, in meters, representing the distance between two sequential trajectories,
	\item the \textit{number of UAVs} ($v_n$) that will participate in the mission
	\item and the \textit{initial positions of the UAVs} in the operational area (could be real, user-defined, or random), formatted in WGS84 as well:
	\begin{equation}
	init_{pos} = \{(lat_{ v_1}, lon_{ v_1}), ... (lat_{ v_{v_n}}, lon_{ v_{v_n}}\}
	\label{eq:init_pos}
	\end{equation}
	where $v_i$ $(i = 1, ... v_n)$ stands for the UAV number (pointer).
\end{itemize}
As an output of the method, a set of paths, one path for each UAV participating in the mission, is produced, so as to provide complete coverage of the ROI:
\begin{equation}
\begin{split}
paths = \{\{(lat_{{ v_1}|1}, lon_{{ v_1}|1}), ... (lat_{{ v_1}|{m_1}}, lon_{{ v_1}|{m_1}})\},\\
... \{(lat_{ v_{{v_n}}|1}, lon_{{ v_{v_n}}|1}), ... (lat_{{ v_{v_n}}|{m_{v_n}}}, lon_{ v_{{v_n}}|{m_{v_n}}})\}\}
\end{split}
\label{eq:paths}
\end{equation}
where $m_i$ $(i = 1, ... v_n)$ is the count of waypoints for each UAV. Figure \ref{ROI2paths} presents the data-flow of the method.

\subsection{Field Representation on Grid}
\label{subsec:field2grid}
\subsubsection{Transformation of Coordinates}
\label{subsec:coordTrans}
As a first step, the input coordinates of the polygon ROI (\ref{eq:polygon}), the obstacles (\ref{eq:obstacles}) and the UAVs' positions (\ref{eq:init_pos}), are transformed from the WGS84 system, to a local NED (North East Down) system \cite{cai2011coordinate}, with a common reference point inside the ROI. The reason why NED coordinate system is preferable, is because coordinates are represented in a local Cartesian system with the metric system being straight in meters. NED coordinates facilitate in the node placement and optimization procedures that are described below. The paths are also calculated in NED coordinates, and have to be transformed back to the WGS84 coordinate system, before given as output.

\subsubsection{Nodes Placement}
\label{nodePlacement}
As a first step, the user-defined ROI must be represented on a grid. The grid size needs to be proportional to the user-defined scanning density - $d_s$. The center of every grid's cell represents a node which will be later used for the MST construction, with the actual path being laid around the constructed spanning tree. Therefore, the nodes in the grid should be placed with a distance of $ d_n = 2 \times d_s $ between each other, in order to achieve the desired distance between trajectories.

\begin{figure}[!b]
	\centering
	\subfigure[Standard and augmented bounding boxes]{%
		\label{fig:Poly2Grid_aug_box}%
		\includegraphics[height=2.5in]{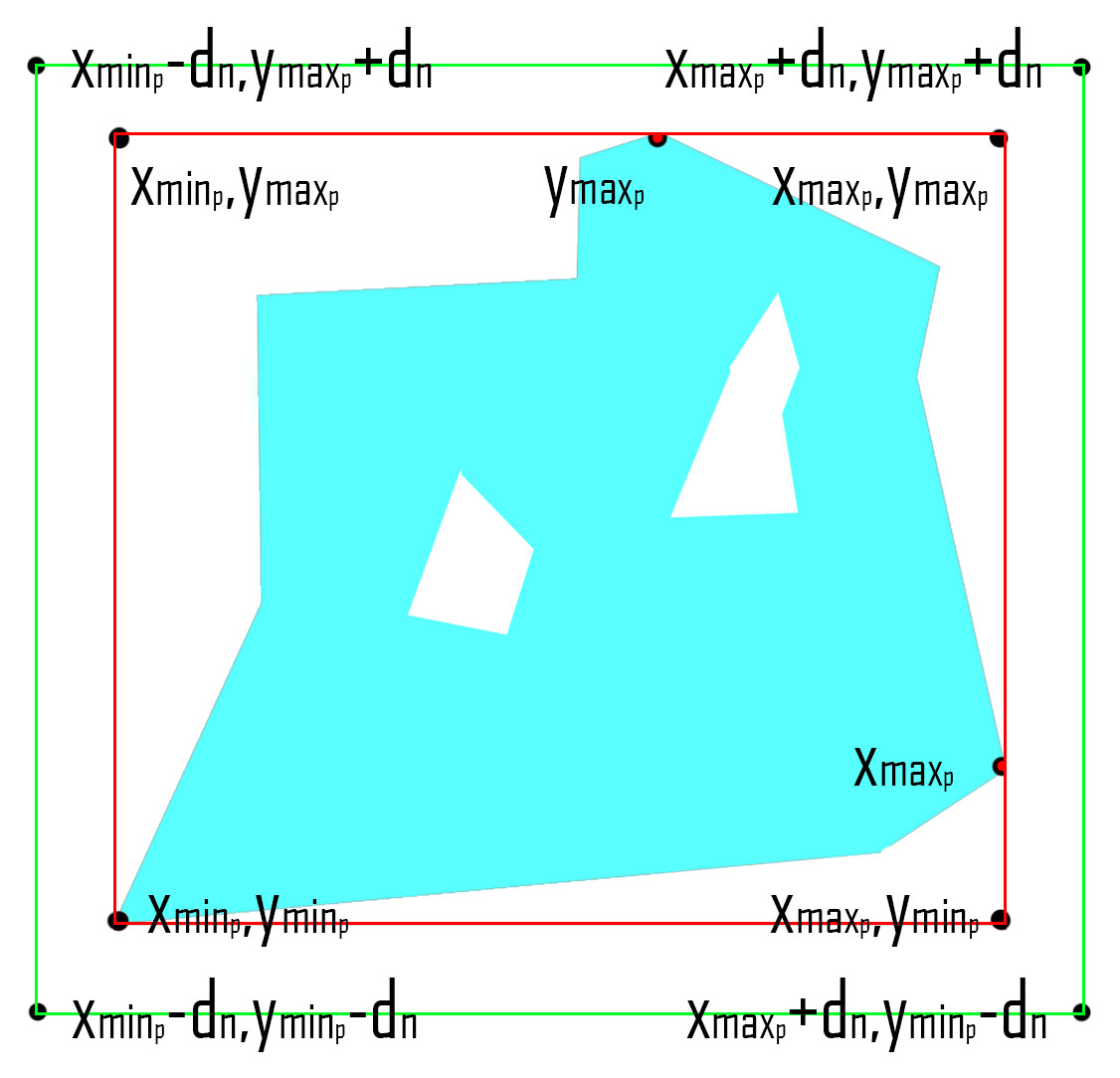}}%
	\qquad
	\subfigure[Nodes placed over polygon - Paths Strictly in Polygon approach]{%
		\label{fig:Poly2Grid}%
		\includegraphics[height=1.4in]{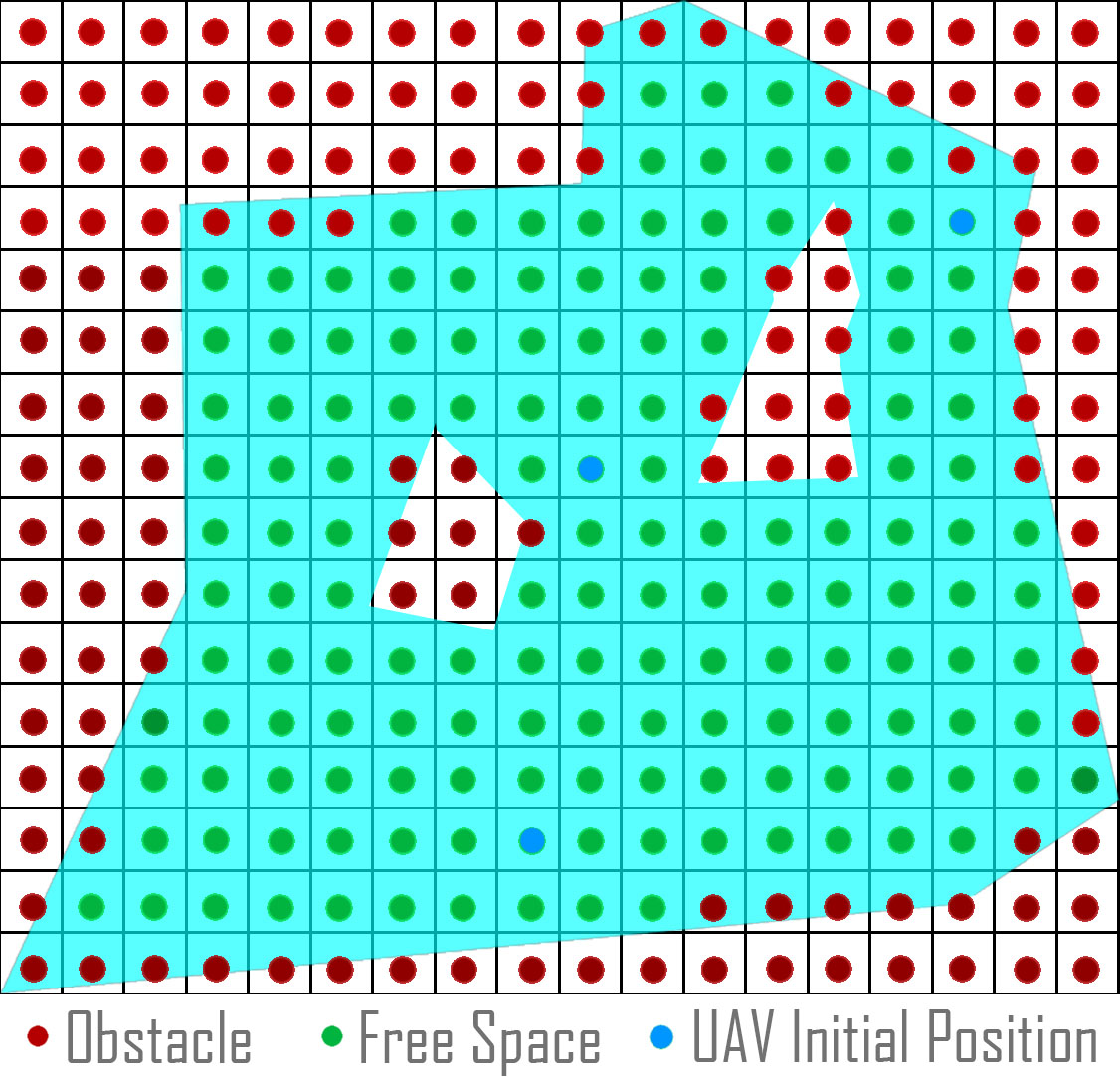}}%
	\qquad
	\subfigure[Nodes placed over polygon - Provide Better Coverage approach]{%
		\label{fig:Poly2Grid_bttrCov}%
		\includegraphics[height=1.4in]{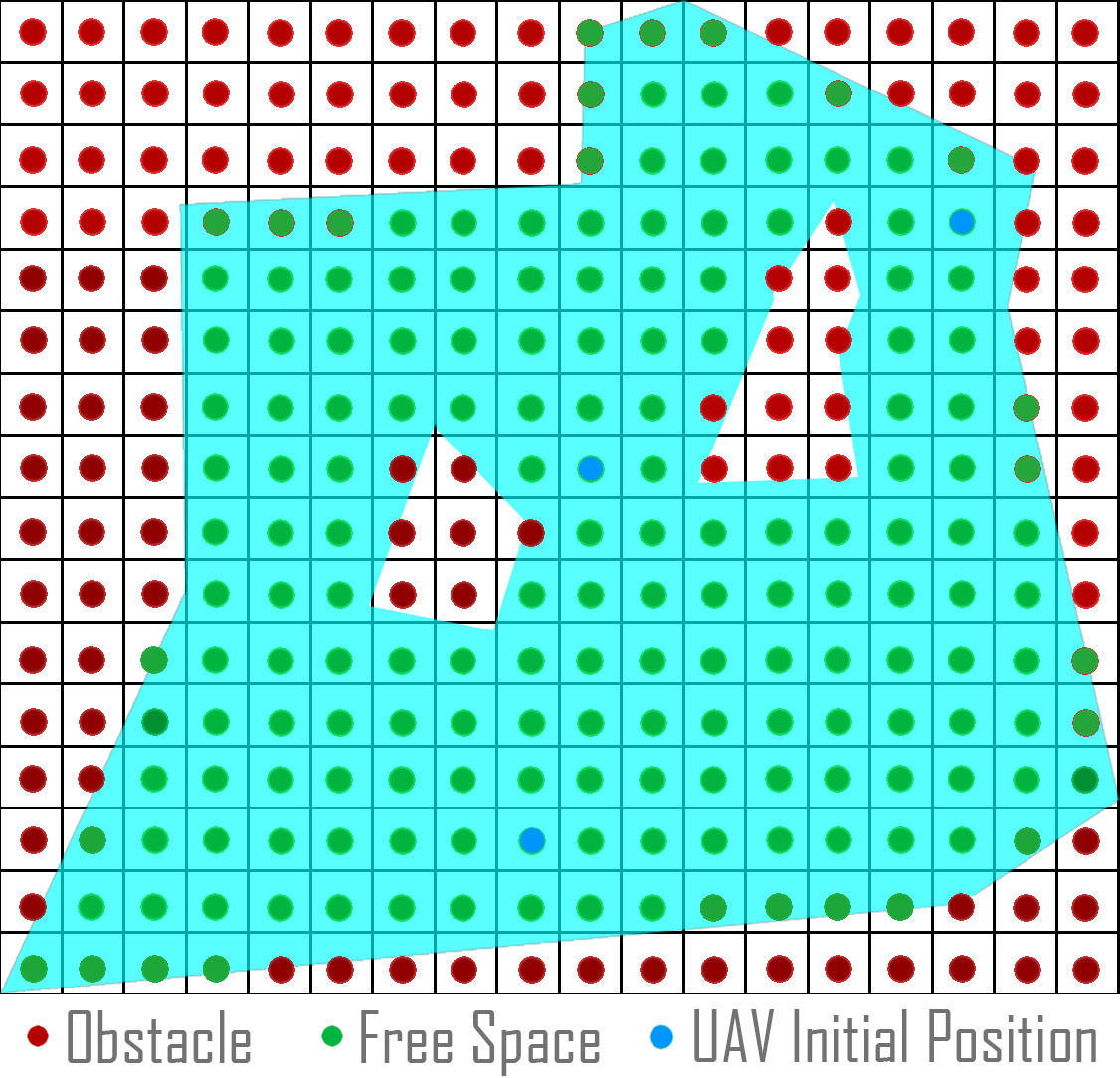}}%
	\caption{Representation of polygon ROI on grid}
	\label{poly2grid_complete}
\end{figure}

To define the size of the grid, a bounding box around the polygon is constructed, as shown in \ref{fig:Poly2Grid_aug_box} (red box). According to this, the grid size will be $x \times y$, where: 
\begin{equation}
x = \lfloor \frac{x_{max_p}-x_{min_p}}{d_n}  \rfloor
\end{equation}
\begin{equation}
y = \lfloor \frac{y_{max_p}-y_{min_p}}{d_n}  \rfloor
\end{equation}
and $\lfloor r \rfloor$ denotes the floor of $r$ (largest integer below the given real number).

After that, the $x \times y$ nodes are placed inside the bounding box with a $d_n$ distance between each other.

\begin{figure*}[!t]
	\centering
	\subfigure[Initial placement]{%
		\label{fig:sub_init}%
		\includegraphics[height=2.1in]{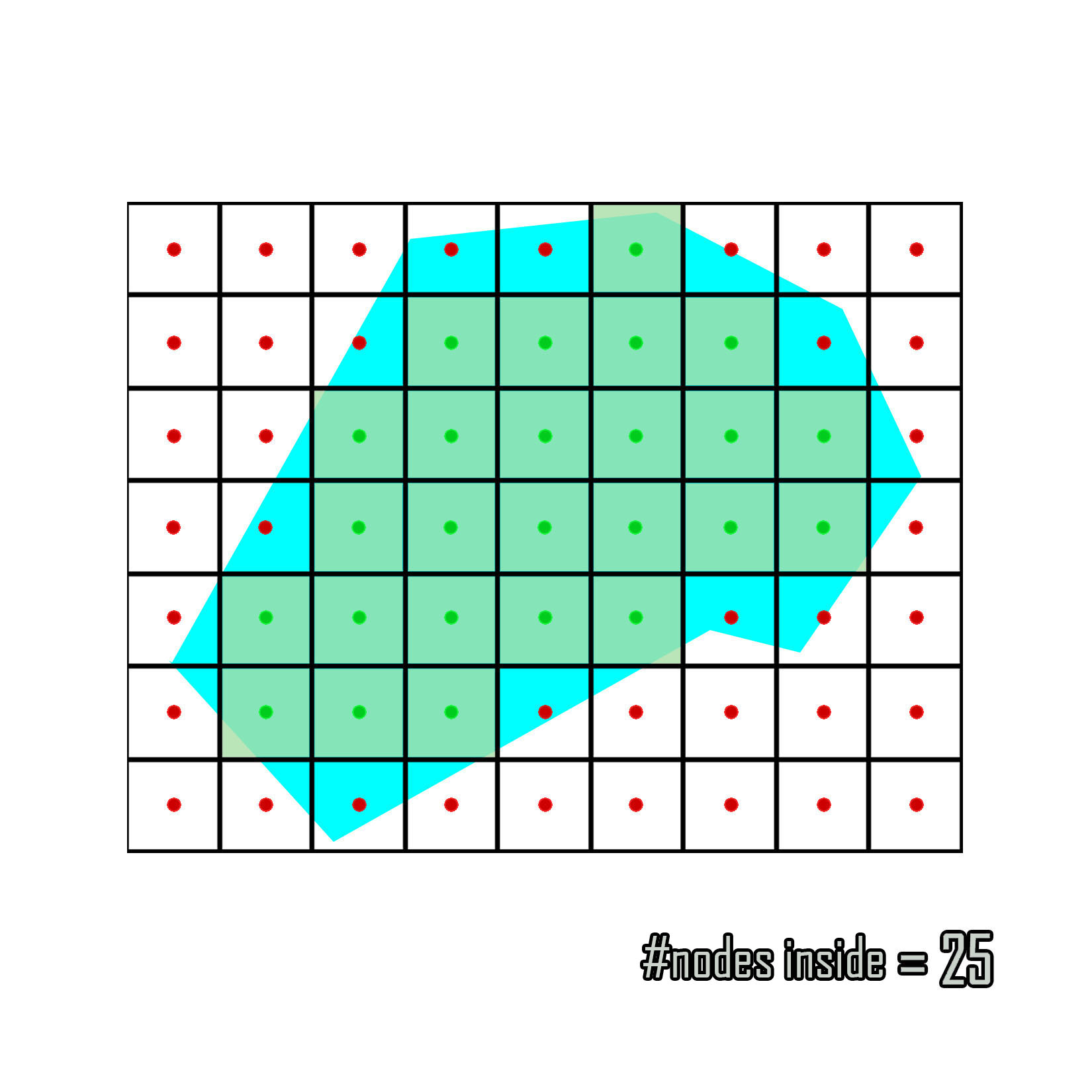}}%
	\qquad
	\subfigure[Rotation over grid]{%
		\label{fig:sub_rot}%
		\includegraphics[height=2.1in]{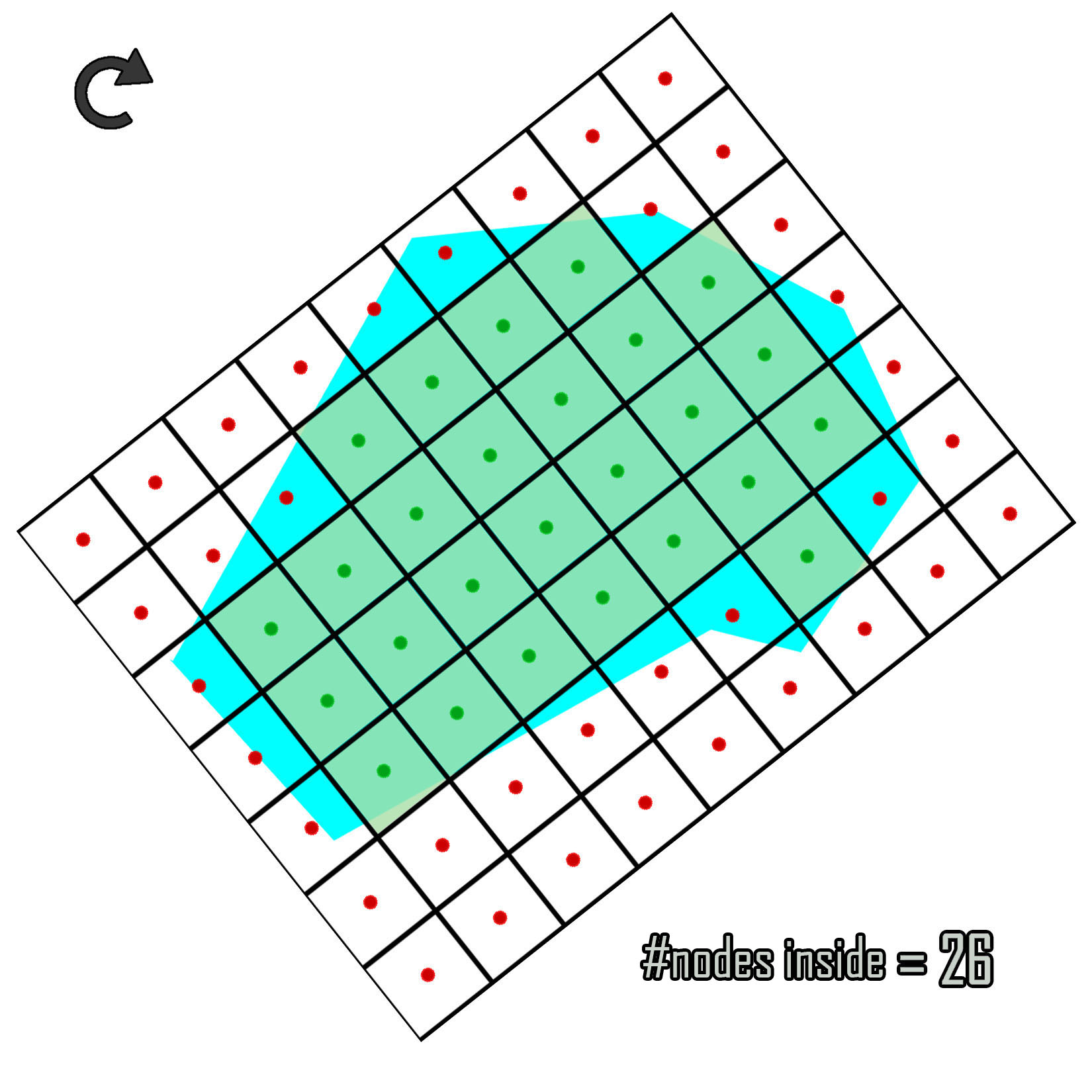}}%
	\qquad
	\subfigure[Shift and rotation over grid]{%
		\label{fig:sub_rotShift}%
		\includegraphics[height=2.1in]{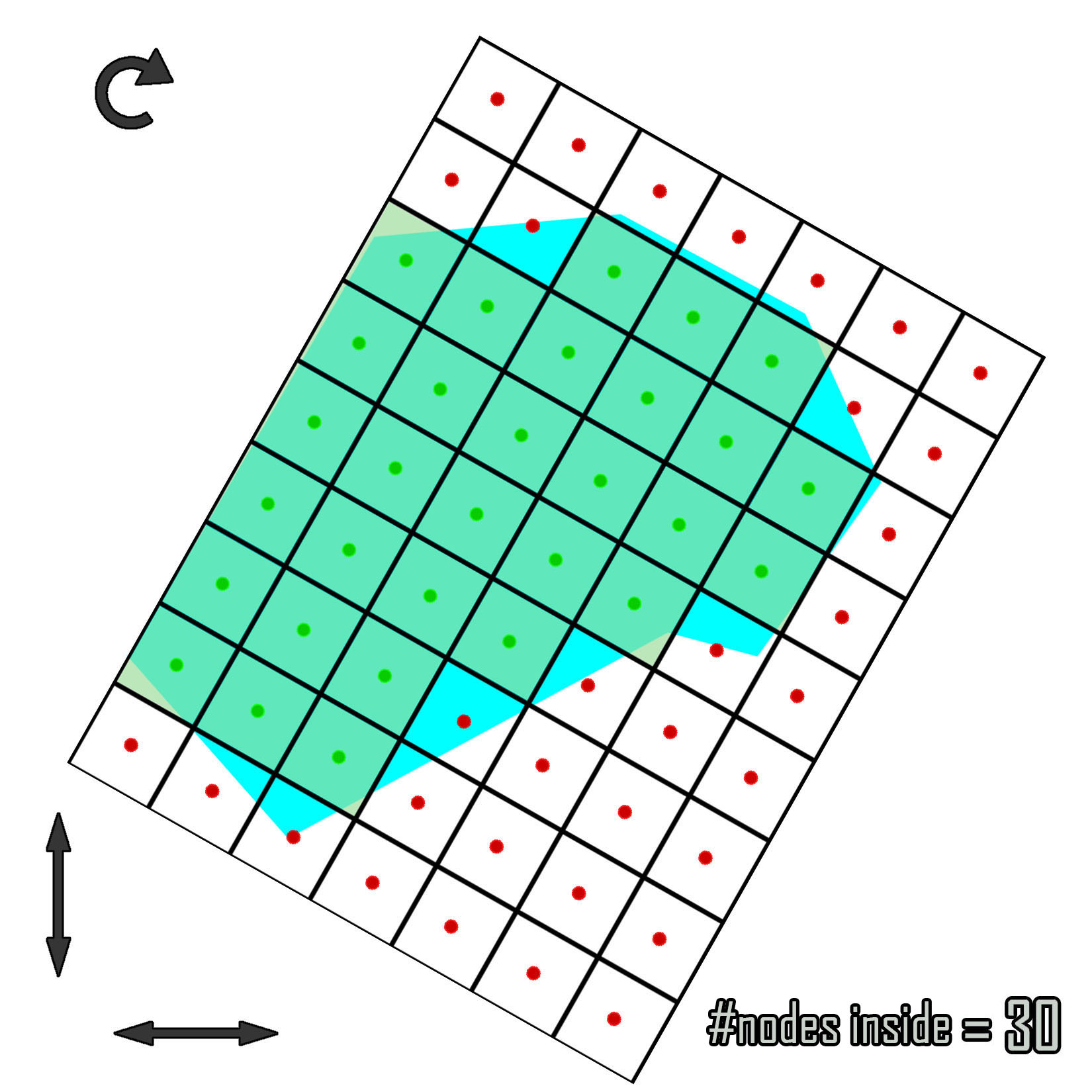}}%
	\caption{Node placement optimization}
	\label{fig:optimization}
\end{figure*}

\subsubsection{Grid Representation}
\label{grid}
At this point, each one of the previously placed nodes should obtain a state that describes how it should be treated during the path generation procedure. The three possible states for every node are:

\begin{itemize}
	\item \textit{Obstacle}, used to represent areas that are outside of polygon, or inside NFZs. These nodes will not be used for the MST construction and no path will be laid around them.
	\item \textit{Free Space}, used to represent nodes that will be used for MST construction and path generation.
	\item \textit{UAV}, used to represent the initial UAVs' position.
\end{itemize}

Two different approaches are implemented to decide whether a node will be labeled as \textit{Obstacle} or \textit{Free Space}. Figures \ref{fig:Poly2Grid}, \ref{fig:Poly2Grid_bttrCov} shown a example of a polygon ROI representation on grid, using both approaches. The first one intends to provide \textit{paths strictly inside the polygon} handling the boundaries of the polygon as a strict geo-fencing zone, while the second one allows paths to get outside of polygon, in order to provide even better coverage of the ROI. For the first approach, the characterization of a node comes out of the area around the node. Specifically, a check is performed, regarding whether the centers of all sub-cells, that will be described in subsection \ref{subsec:MSTpaths}, are inside the polygon, or not. For the second approach, the same check is performed, however, only for nodes themselves, ignoring the area around them.

\subsection{Node Placement Optimization}
\label{subsec:optimization}

One of the main issues that grid-based methods usually face when used for complex-shaped polygons, is that due to discretization issues the representation of the ROI ends up being a lot different and smaller in comparison with the actual region. That gets translated to incomplete coverage of the ROI. A common practice to overcome this problem is to use a smaller discretization scale. However, as mentioned above, the grid size should be proportional to the user-defined $d_s$. Forcing a $d_s$ that is smaller than the indicated one, in order to provide a descent coverage is a waste of resources, as it increases time and energy demands for a mission to be carried out. Moreover, large grid dimensions are translated to bigger demands in memory and computational power for the path planning method. In this work, in order to overcome this issue an optimization procedure is proposed, that intends to optimize the nodes' placement, given the user-defined $d_s$.

The key idea is that by \textit{rotating} and \textit{shifting} the polygon over a grid that respects the defined $d_s$, more favorable configurations may appear, where the count of nodes that are annotated as \textit{Free Space} is greater. An example where these transformations are applied to a specific ROI is presented in figure \ref{fig:optimization}, for the better understanding of this idea. Such configurations are more likely to provide paths that achieve a high percentage of coverage for the given ROI. Based on this idea, an optimization process utilizing the \textit{Simulated Annealing} \cite{Kirkpatrick1983OptimizationBS} algorithm is proceeded, in order to maximize the overall coverage of the ROI. The control variables for the aforementioned optimization problem are selected as follows:

\begin{itemize}
	\item $s_x$, which controls the shifting of the polygon over the grid in X axis, in a range $[0,d_n]$,
	\item $s_y$, which controls the shifting of the polygon over the grid in Y axis, in a range $[0,d_n]$,
	\item $\theta$, which controls the rotation of the polygon over the grid, in a range $[0,90^\circ]$.
\end{itemize}	

It should be noted that the control variables and their ranges are selected so as to provide all the available configurations of node positioning inside the polygon. Combining these three variables in the selected ranges and allowing values for them in a continuous space, ensures the availability of every possible configuration. For the implementation of the optimization procedure, an augmented grid (figure \ref{fig:Poly2Grid_aug_box}) that does not restrict the rotation and shifting for the given polygon, in the given control variables ranges, is constructed.

\subsubsection{Optimization Index}
\label{subsec:optimInd}	
The optimization index $J$ intends to represent a quantity directly matched with the coverage potential of a polygon ROI, given a fixed scanning density and is formulated as follows:

\begin{equation}
\label{eq:optIndex}
J = a \cdot J_1 +b \cdot J_2 - c \cdot J_3
\end{equation}
$J$ is consisted of three separate, normalized $J_i$ terms, so that:
$$0 \leq J_i \leq 1 \in \Re, \; i \in \{1,2,3\},$$
each aiming to control a specific aspect of the optimal node placement.

$J_1$ term expresses the fundamental objective of the overall optimization procedure, which is to fit the maximum count of nodes possible inside the given ROI, and is defined as follows:

\begin{equation}
\label{eq:nodesInTerm}
J_1 = \frac{s}{\frac{A_p}{d_n ^2}},
\end{equation} 
where $s$ stands for the count of nodes placed inside the polygon, that will be used for the MST construction and $A_p$ stands for the area of the polygon. Denominator represents the theoretical maximum count of nodes that could fit inside the given polygon and is used to normalize $J_1$. Thus,
$$s \leq \frac{A_p}{d_n ^2}$$
Both $A_p$ and $d_n$ are constant for a specific mission, so, maximizing $J_1$ term, also maximizes the count of nodes that will be placed inside the polygon.

In addition, $J_2$ term intends to provide a better placement of the nodes inside the polygon, so as to provide increased coverage for the marginal areas of the ROI, and is defined as follows:

\begin{equation}
\label{eq:minBBAreaTerm}
J_2 = \frac{A_p}{A_{bb}},
\end{equation} 
where $A_{bb}$ stands for the area of the augmented bounding box.  In the example of figure \ref{fig:Poly2Grid_aug_box}, $A_p$ represents the cyan-colored area of the polygon and $A_{bb}$ represents the area of the green, rectangle bounding box. Since $A_p$ is constant for a specific ROI, by maximizing the overall $J_2$ term, $A_{bb}$ is minimized. In practice, the addition of this term mostly controls the rotation angle $\theta$, in order to be optimal for the given polygon. Apparently,
$$A_p<A_{bb},$$
thus $J_2$ term is normalized as well.

Finally, $J_3$ term aims to improve the placement of the nodes inside the given ROI, attempting to equally align the paths from the ROI's borders and provide increased coverage for the marginal areas as well. $J_3$ is defined as follows:

\begin{equation}
\label{eq:eqMargTerm}
\begin{split}
J_3 = \frac{\left| | x_{max_{bb}} - x_{max_p}| - | x_{min_p} - x_{min_{bb}} | \right|}{2 \cdot \left| x_{max_{bb}} - x_{min_{bb}} \right|} + \\
+ \frac{\left| | y_{max_{bb}} - y_{max_p}| - | y_{min_p} - y_{min_{bb}} | \right|}{2 \cdot \left| y_{max_{bb}} - y_{min_{bb}} \right|},
\end{split}
\end{equation} 
where \{$x_{max_{bb}}$, $x_{min_{bb}}$, $y_{max_{bb}}$, $y_{min_{bb}}$\} are the vertices of the augmented bounding box (green box in figure \ref{fig:Poly2Grid_aug_box}) and \{$x_{max_p}$, $x_{min_p}$, $y_{max_p}$, $y_{min_p}$\} are the vertices of the standard bounding box (red box in figure \ref{fig:Poly2Grid_aug_box}). This term quantifies the absolute difference of margins between the polygon and the augmented bounding box, for each axis. $J_3$ is normalized as well, with both fractions contributing equally to the overall term. The minimization of $J_3$ aligns the standard bounding box in the center of the augmented bounding box and in practice performs micro-adjustments mostly in the $s_x$ and $s_y$ control variables, in order to provide the best possible nodes' placement for the coverage paths.

\begin{figure}[!b]
	\centering
	\subfigure[J1 term]{%
		\label{fig:J1_term}%
		\includegraphics[height=1.24in]{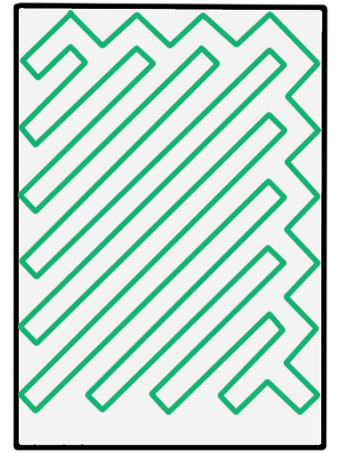}}%
	\qquad
	\subfigure[J1+J2 terms]{%
		\label{fig:J1-J2_term}%
		\includegraphics[height=1.24in]{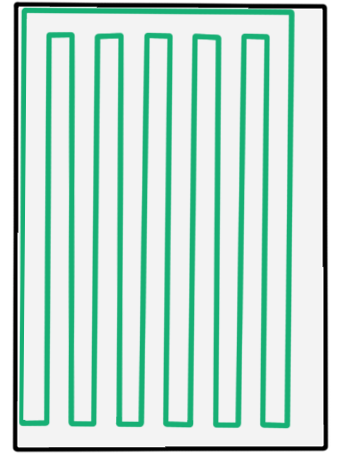}}%
	\qquad
	\subfigure[J1+J2+J3 terms]{%
		\label{fig:J1-J2-J3_term}%
		\includegraphics[height=1.24in]{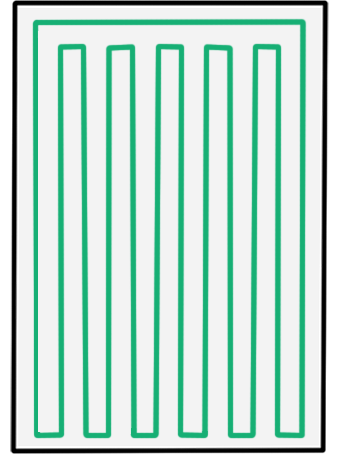}}%
	\caption{Contribution of each optimization term}
	\label{fig:Ji_terms}
\end{figure}

%

Finding a tuple of $\{s_x, \; s_y, \; \theta\}$ that maximizes $J$ leads to an optimized nodes' setup for the given ROI and $d_s$. $J_1$ and $J_2$ have a positive contribution and act as a reward to the overall optimization index, while $J_3$ acts as a penalty. The percentage of influence for every term, is regulated by the constants a, b, c. The overall optimization index $J$ is normalized as well, with the constants allowed to take values in the following ranges:
$$0 \leq J < 1 \in \Re ,$$
$$\text{for} \quad 0 \leq a,b,c \leq 1 \in \Re, \quad a+b=1$$ 
The normalization of J helps to make it independent of polygon's size and shape, i.e. no extra effort is needed to adjust the constants a,b,c on each different polygon topology.

Figure \ref{fig:Ji_terms} intends to visually explain the contribution of each of the $J_i$ terms $(i \in [1,2,3])$ to the overall optimization procedure, using an example of a fixed ROI and $d_s$ and showing the effect of the sequential addition of each term. All of these three cases manage to include the same number of nodes in the ROI (54 nodes), thus they are optimal with respect to the $J_1$ term's objective. The addition of $J_2$ term (figure \ref{fig:J1-J2_term}), facilitates the calculation of the optimal for this grid rotation angle $\theta$, affecting also significantly the number of turns in the generated path. Finally, $J_3$ term's role (figure \ref{fig:J1-J2-J3_term}) is to center the generated path in the ROI, providing a better coverage for the marginal areas. Overall, it should be noted that the simpler the shape of a ROI is, the greater the contribution of $J_2$ and $J_3$ terms get. While in complex-shaped, concave ROIs $J_1$ term carries the greatest load of this optimization procedure's objective, in simpler cases, such as the one presented in figure \ref{fig:Ji_terms}, the role of the other two optimization terms is of critical importance for the achieved coverage and the overall qualitative features of the paths.

\subsection{Task Allocation \& Path Calculation}
\label{subsec:PathPlanning}
Having an optimized representation of the ROI on a grid, DARP algorithm \cite{kapoutsis2017darp} takes over to divide the complete region to sub-regions, exclusive for each UAV, and provide independent paths for all of them utilizing STC \cite{gabriely2001spanning}. The generated paths have energy efficient characteristics, as they respect UAVs' initial positions, avoiding redundant movements that do not contribute to the coverage process and reduce the number of unnecessary turns.

\subsubsection{DARP - Area Division}
\label{subsec:DARP}

DARP algorithm solves the mCPP problem by dividing it into multiple single-robot CPP problems. It is an iterative process implemented to construct an assignment matrix, labeling all cells to correspond to a unique UAV. More information regarding DARP's implementation details, along with a complexity analysis, can be found in \cite{kapoutsis2017darp}.

After the area division, every UAV undertakes a specific part of the grid to cover. Having exclusive sub-regions ensures the generation of non-intersecting paths, making them safe for multiple UAVs to operate at same time, avoiding potential collisions among them.

\begin{figure*}[!t]
	\centering
	\subfigure[Progress 0\% - initial state]{%
		\label{fig:darp_init}%
		\includegraphics[height=1.68in]{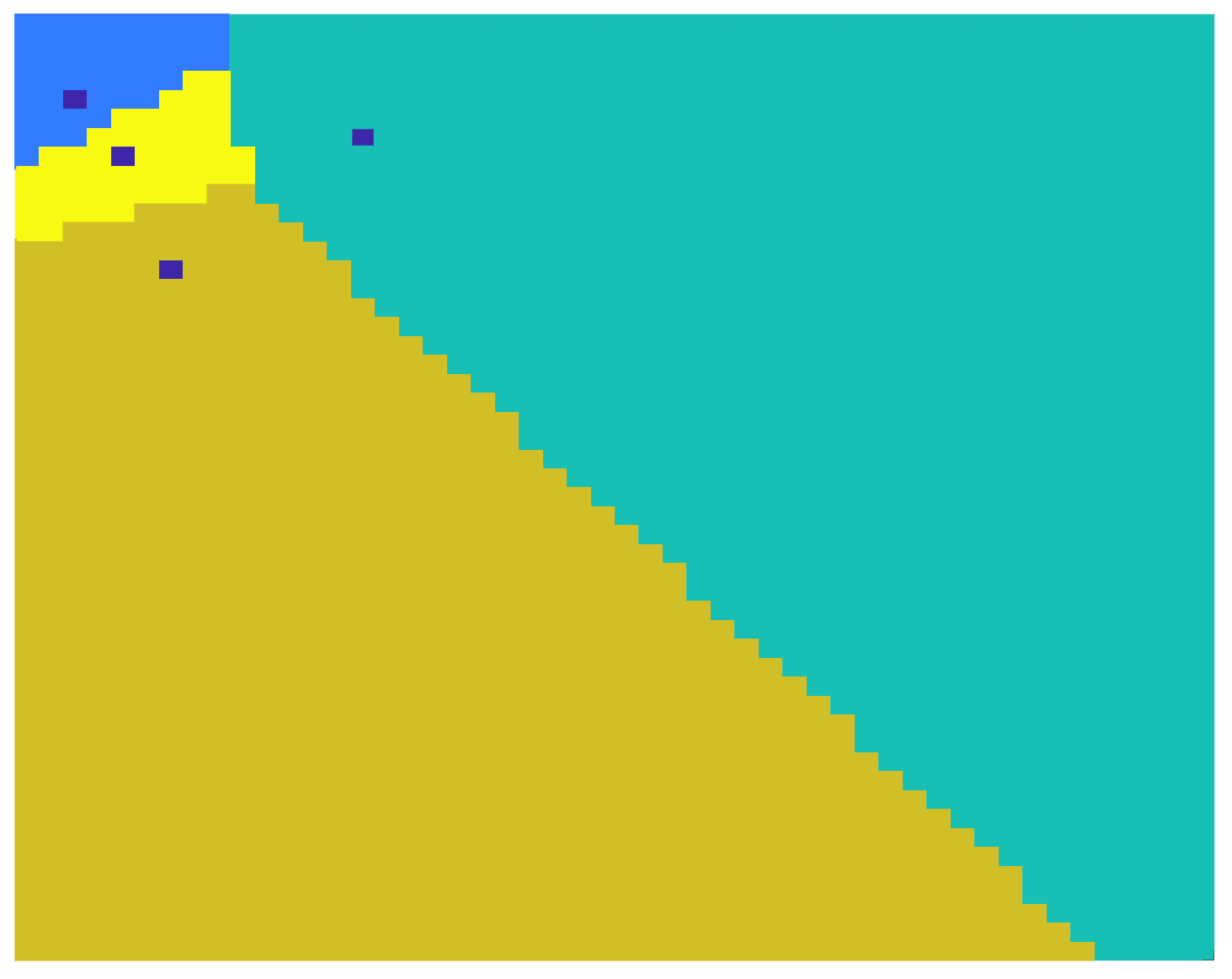}}%
	\qquad
	\subfigure[Progress 20\%]{%
		\label{fig:darp_150}%
		\includegraphics[height=1.68in]{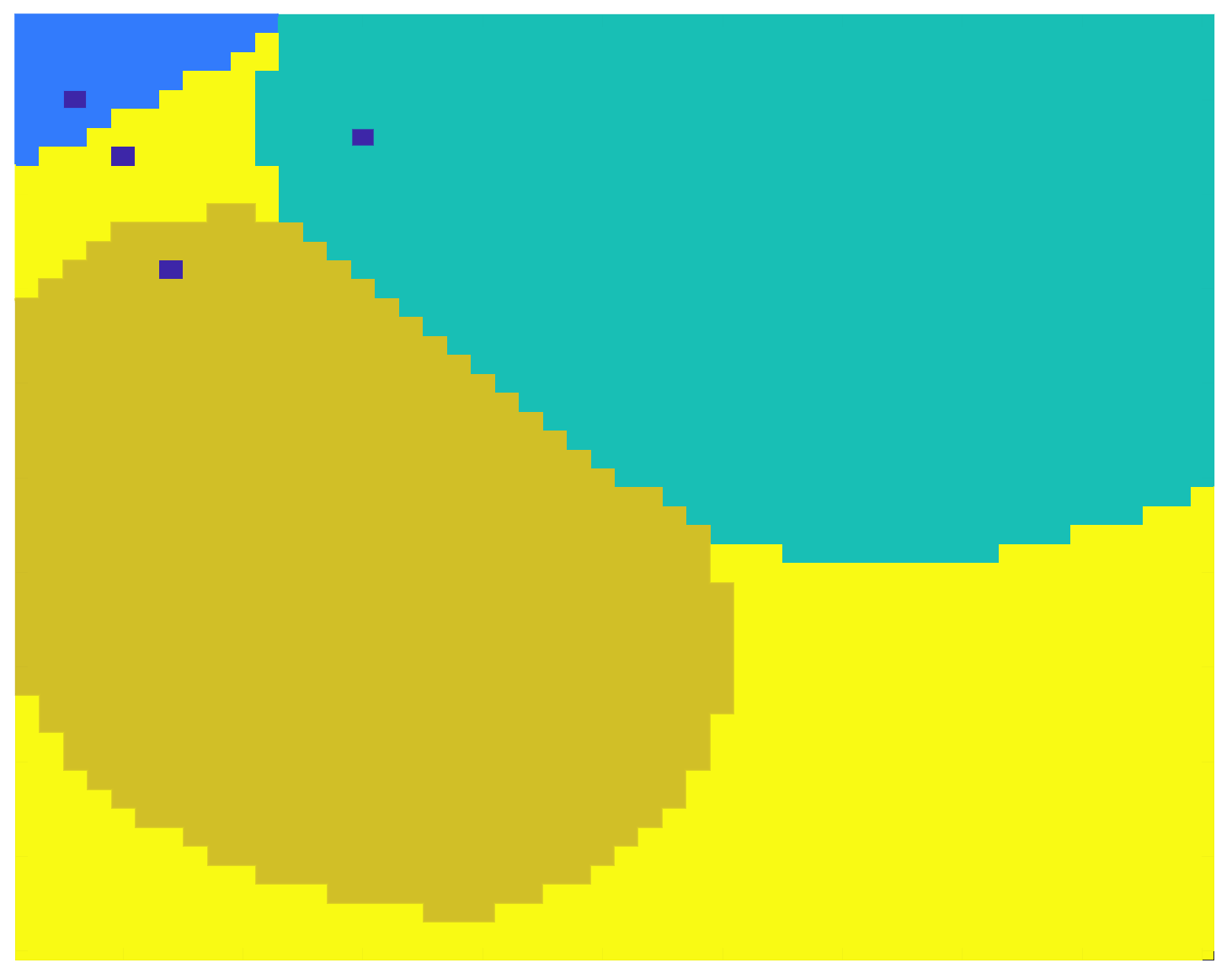}}%
	\qquad
	\subfigure[Progress 40\%]{%
		\label{fig:darp_300}%
		\includegraphics[height=1.68in]{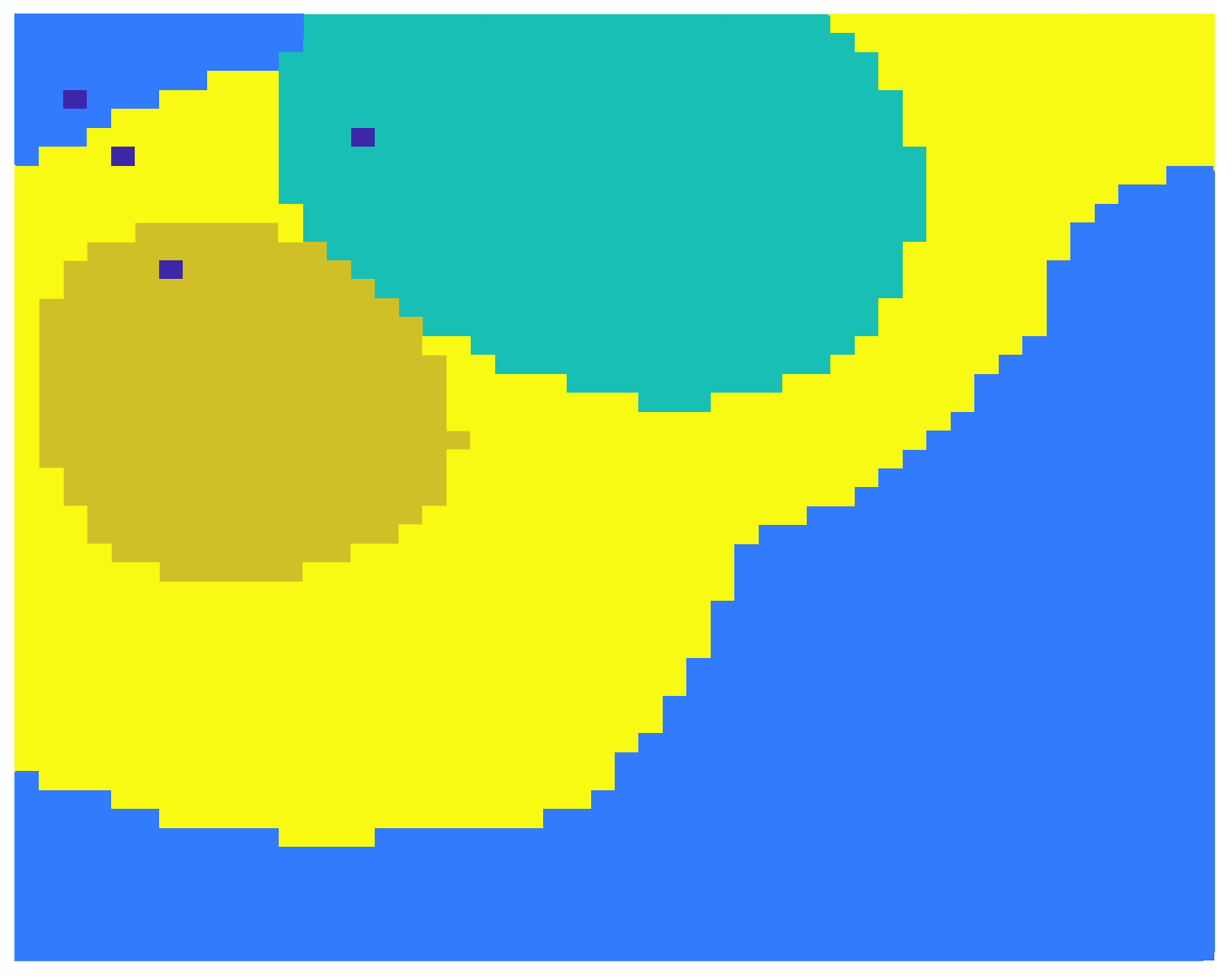}}%
	\qquad
	\subfigure[Progress 60\%]{%
		\label{fig:darp_450}%
		\includegraphics[height=1.68in]{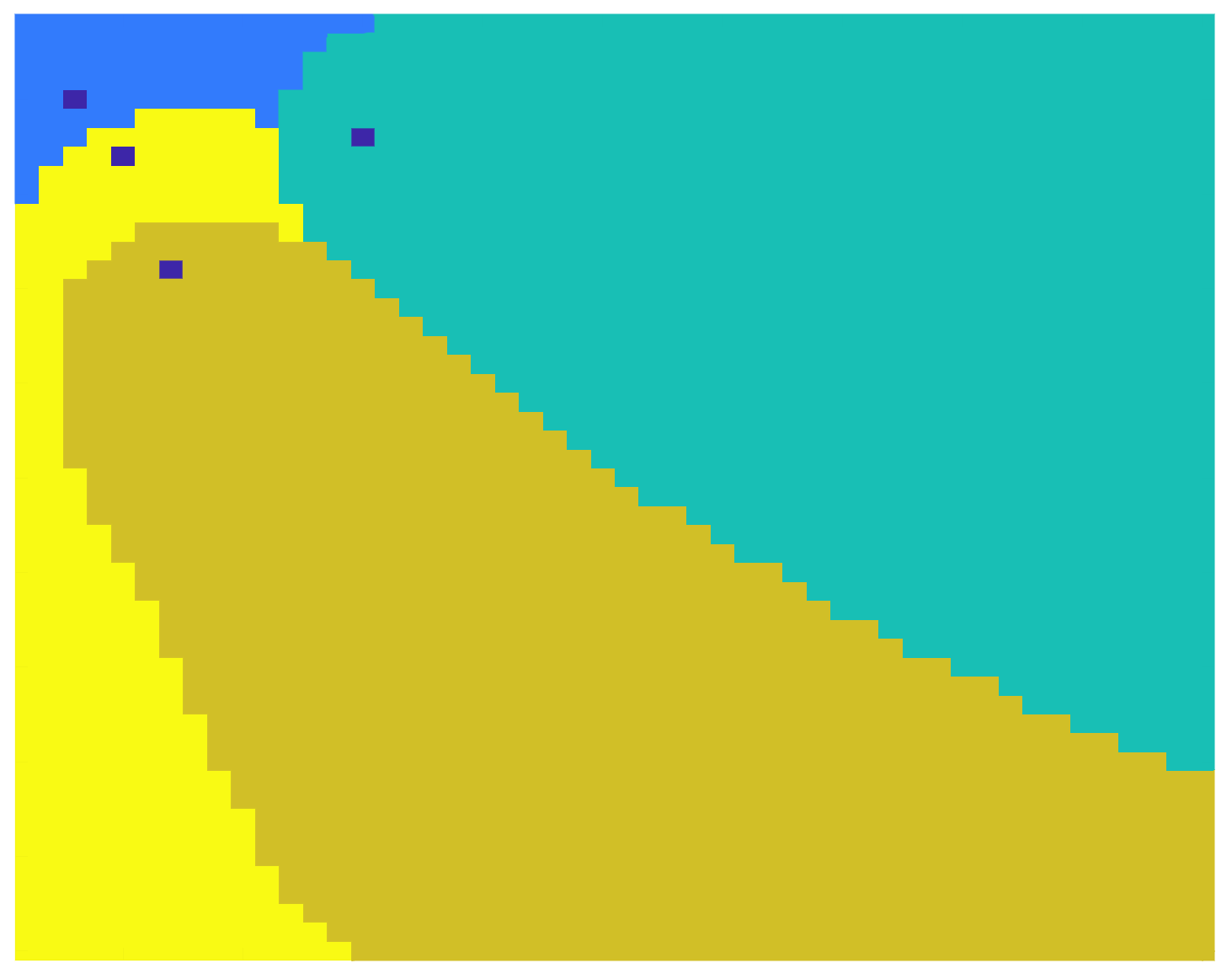}}%
	\qquad
	\subfigure[Progress 80\%]{%
		\label{fig:darp_600}%
		\includegraphics[height=1.68in]{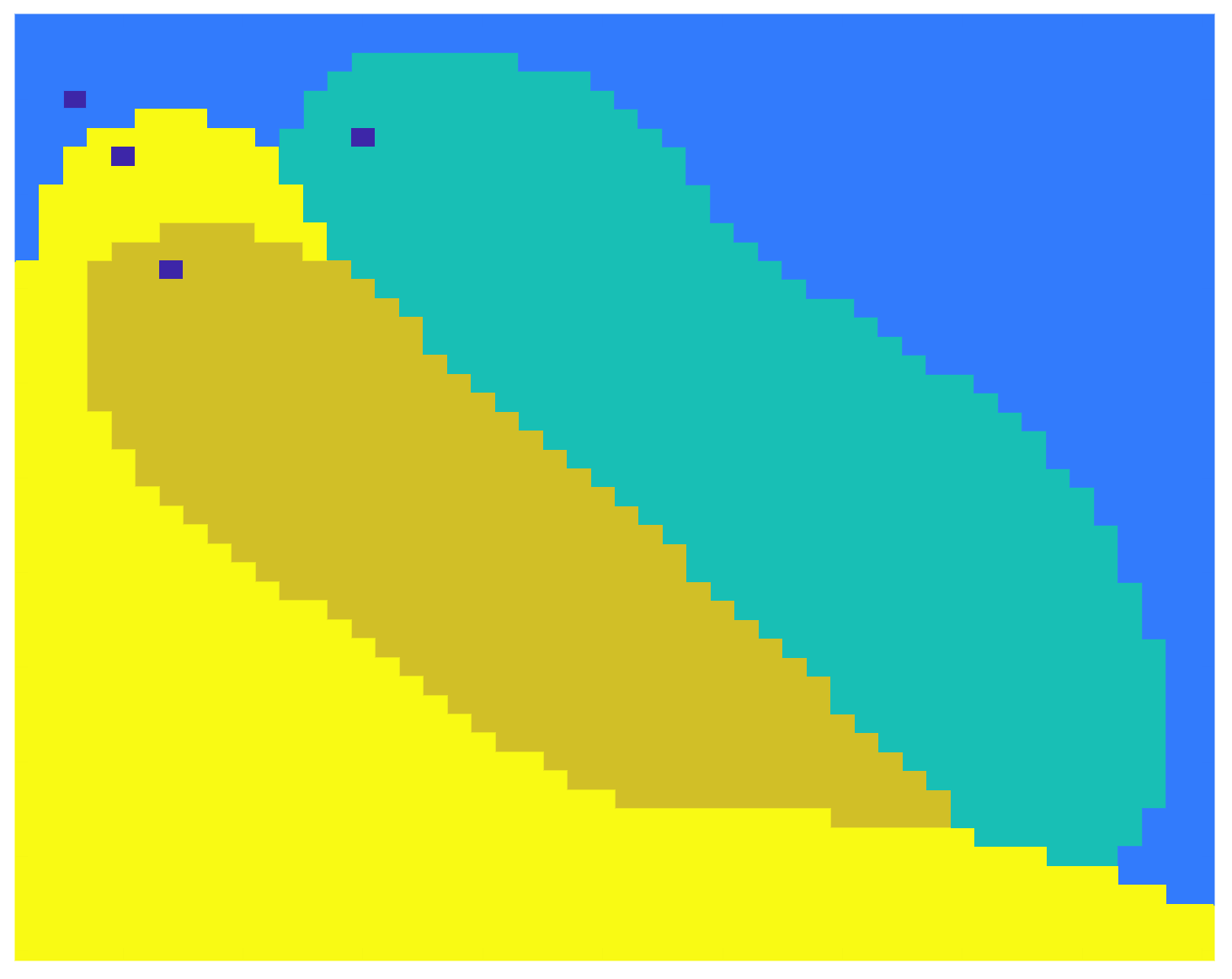}}%
	\qquad
	\subfigure[Progress 100\% - converged state]{%
		\label{fig:darp_final}%
		\includegraphics[height=1.68in]{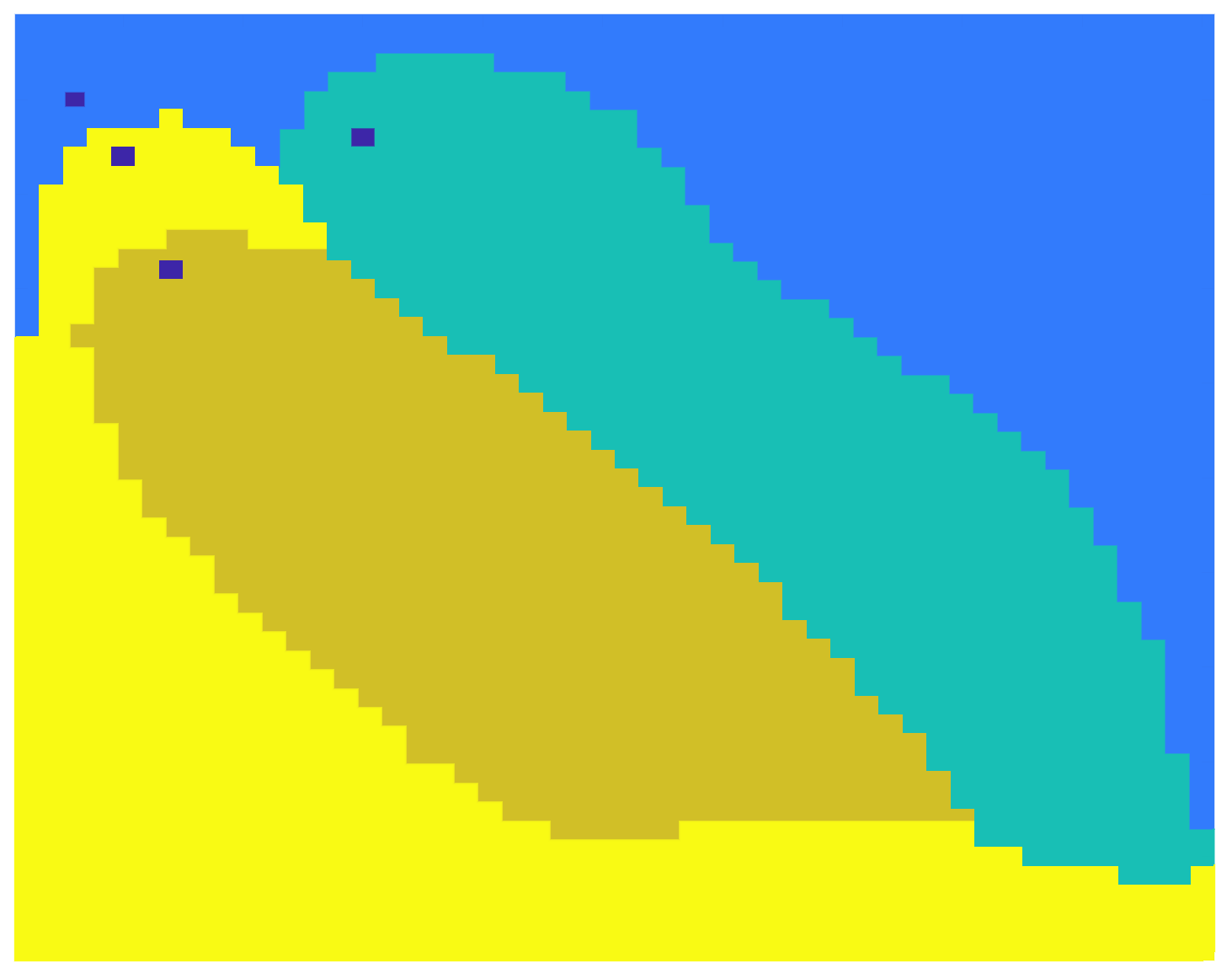}}%
	\caption{Area allocation during different time-steps of DARP execution}
	{The ROI is a plain orthogonal area without obstacles and the initial positions of the UAVs are marked with dark blue dots}
	\label{fig:darp}
\end{figure*}

Figure \ref{fig:darp} shows the allocation of an area to four robots, for the initial positions shown, over the execution time of DARP. \ref{fig:darp_init} shows the initial Voronoi allocation, while \ref{fig:darp_final} shows the final solution that the algorithm has converged to.

\begin{figure}[!h]
	\centering
	\includegraphics[width=.7\linewidth]{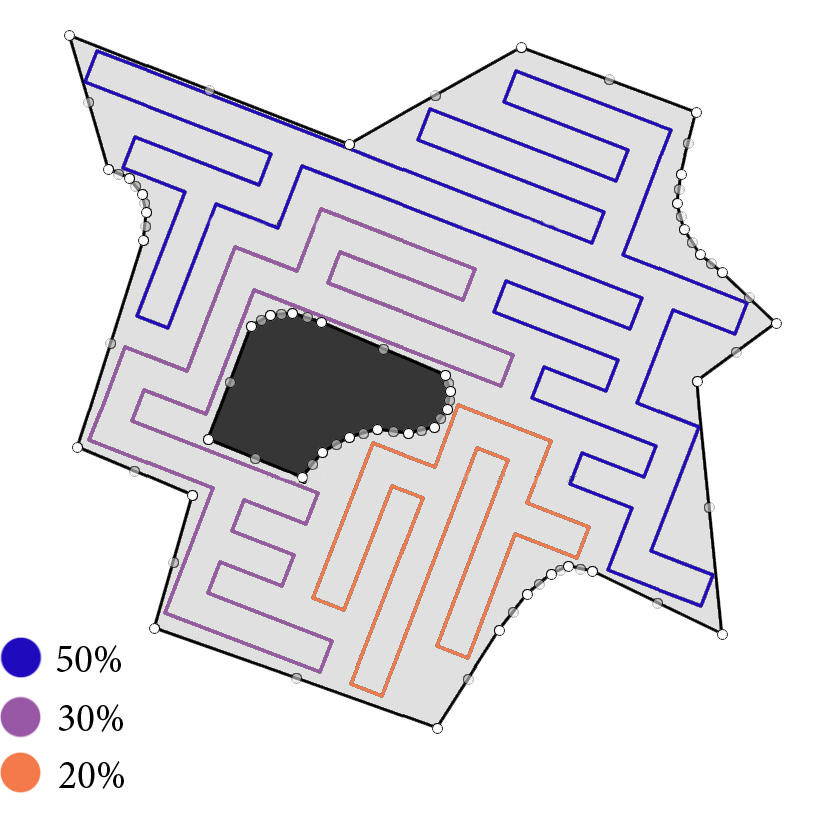}
	\caption{Proportional area allocation}
	\label{fig:portions}
\end{figure}

The original version of DARP \cite{kapoutsis2017darp} provides a fair area division, meaning that all UAVs will be assigned with exactly the same percentage of the area to cover. However, in the context of this work, DARP has been extended and is now able to perform a proportional area division as well. The percentages in this case are user-defined and facilitate the simultaneous operation of heterogeneous UAVs, with different energy and operational capabilities in the same mission. To achieve this, the equation 14 of \cite{kapoutsis2017darp} for the overall cost function, presented below:

$$
J = \frac{1}{2}\sum_{i=1}^{v_n}(k_i - f)^2
$$
where $k_i$ and f denote the current portion for the i-th UAV and the global ``fair share'': $f = \frac{l}{v_n}$ (\#\textit{unoccupied cells} devided by the \#\textit{UAVs}) respectively, is altered as follows:

\begin{equation}
\label{eq:portionAllocation}
\begin{split}
J = \frac{1}{2}\sum_{i=1}^{v_n}(k_i - p_i)^2,\\
\sum_{i=1}^{v_n}p_i = 1
\end{split}
\end{equation}
where $p_i$ denotes the target portion for the i-th UAV. Figure \ref{fig:portions} shows an example of paths' generation for a proportional area allocation based on user-defined percentages.

\subsubsection{Reduced Turns MSTs \& Path Generation}
\label{subsec:MSTpaths}

Once every robot gets assigned with its exclusive part of the overall area, a single-robot STC problem is solved for every sub-region. For the path generation the standard approach used in the STC algorithm \cite{gabriely2001spanning} is followed. Specifically, a MST is constructed for every sub-region and a path that circumnavigates it is generated afterwards.

\begin{figure}[!h]%
	\centering
	\subfigure[Upper]{%
		\label{fig:upper}%
		\includegraphics[width=0.18\linewidth]{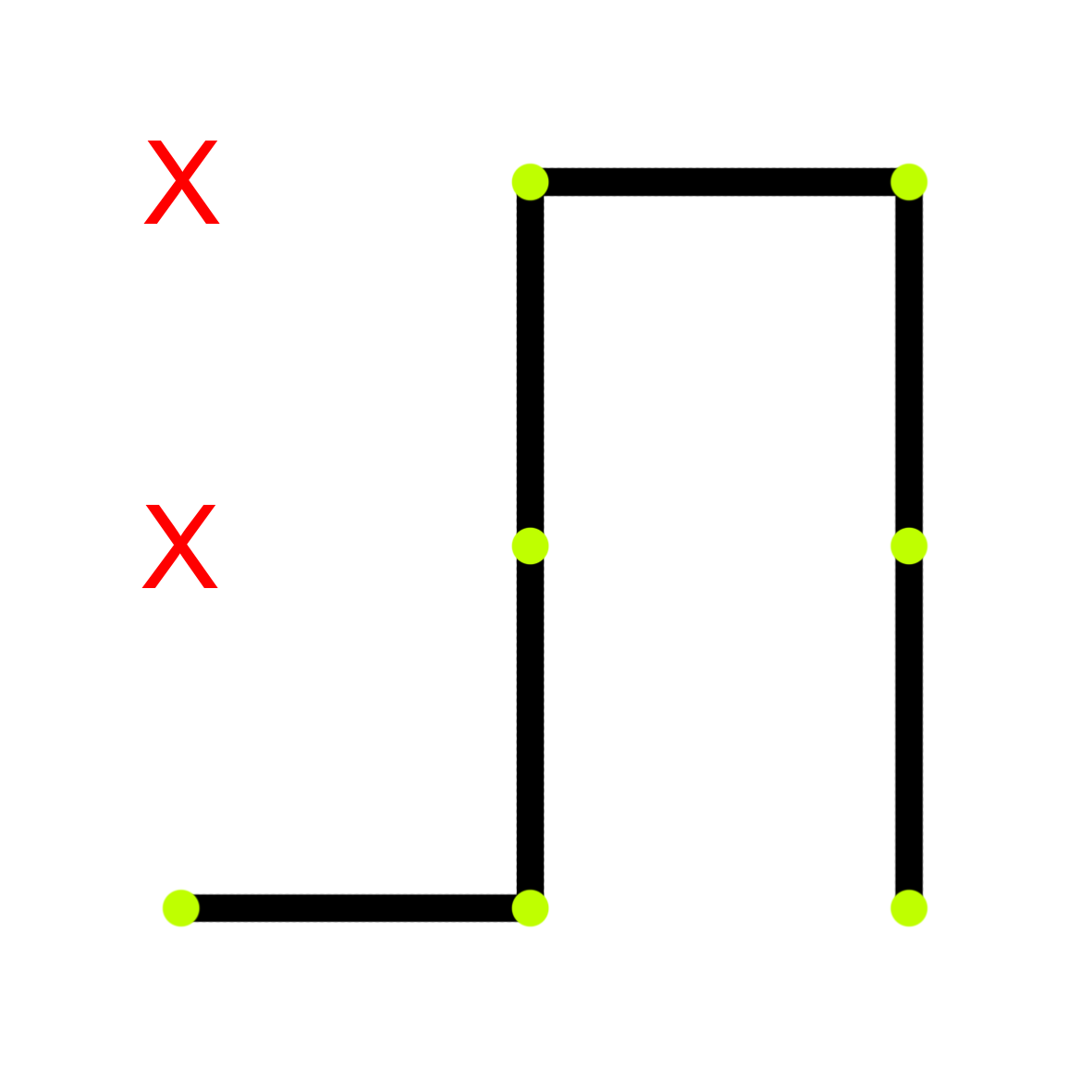}}%
	\qquad
	\subfigure[Lower]{%
		\label{fig:lower}%
		\includegraphics[width=0.18\linewidth]{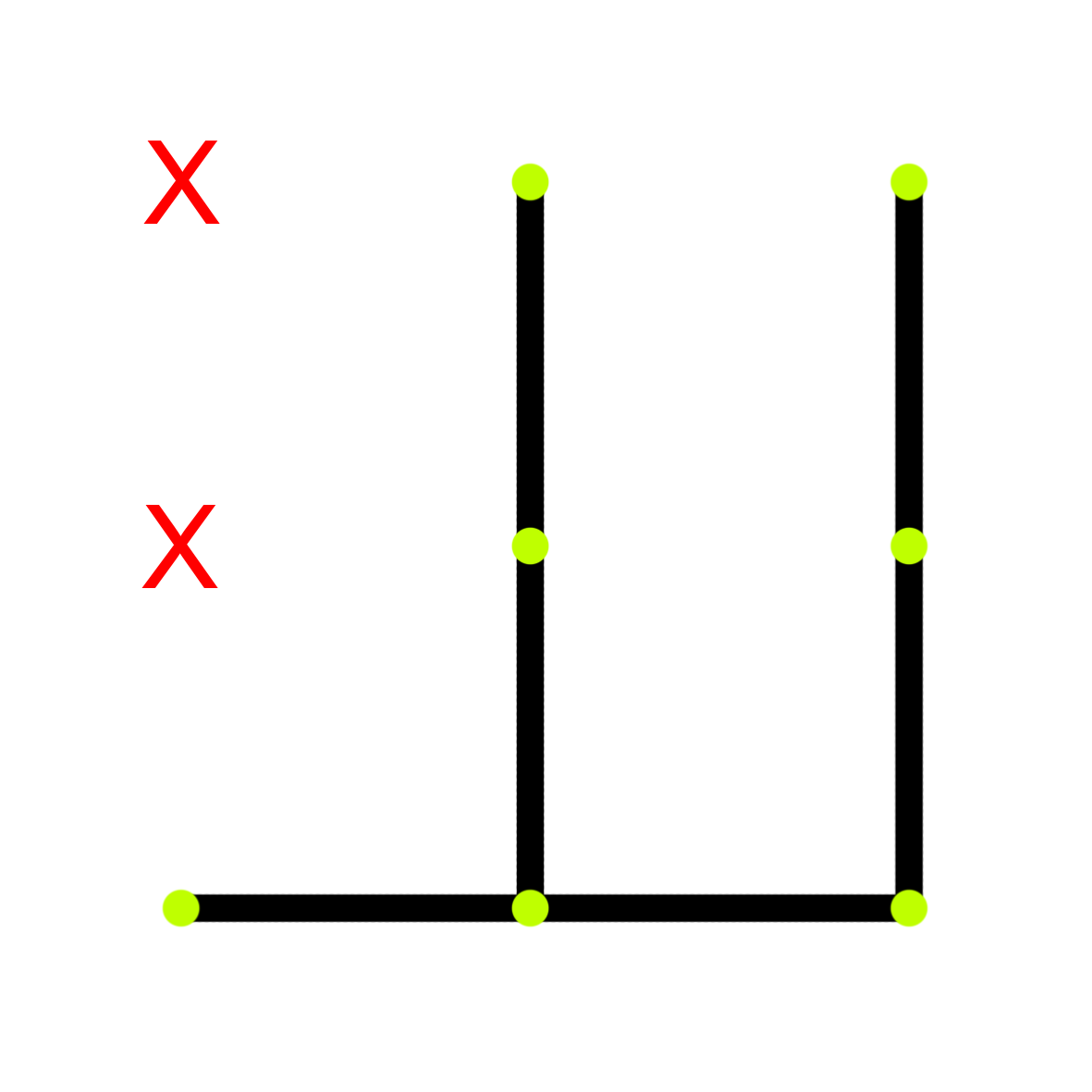}}%
	\qquad
	\subfigure[Most right]{%
		\label{fig:right}%
		\includegraphics[width=0.18\linewidth]{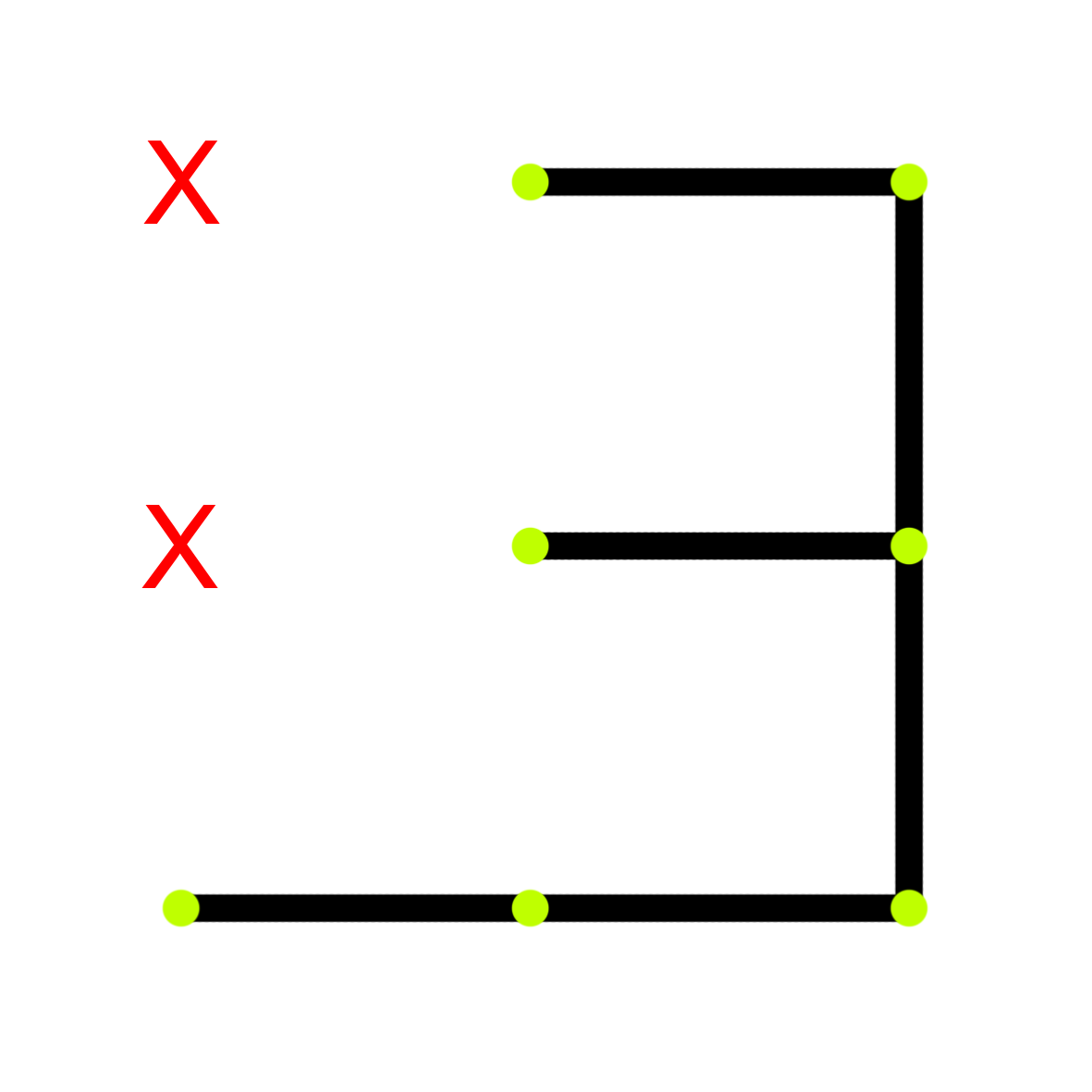}}%
	\qquad
	\subfigure[Most left]{%
		\label{fig:left}%
		\includegraphics[width=0.18\linewidth]{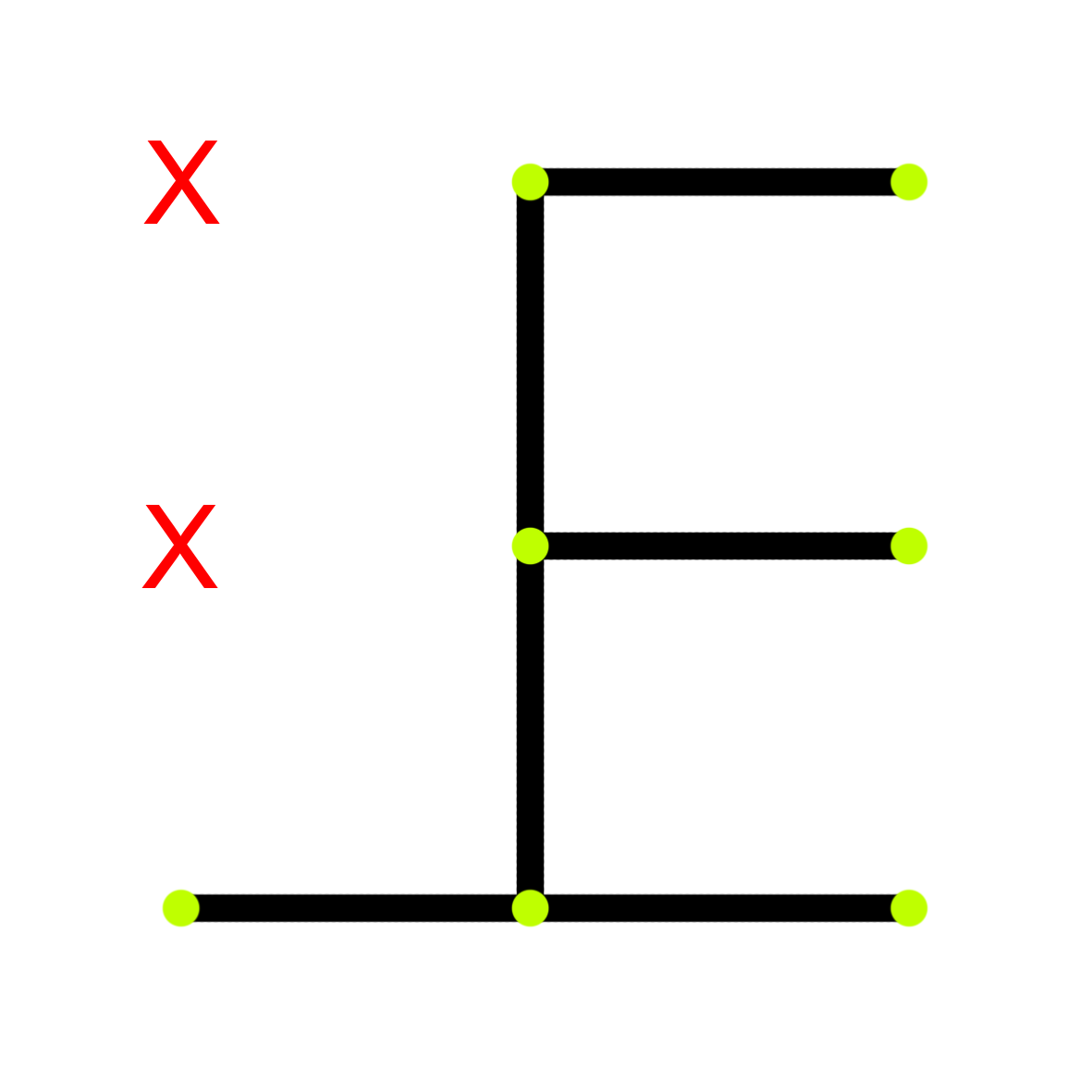}}%
	\caption{MSTs connections tested for turn reduction}
	\label{fig:MSTsTurnReduction}
\end{figure}

Due to the spanning-tree nature of the algorithm, it is likely that paths will end up having a lot of unnecessary turns, if no action is taken to avoid them. The number of turns is one of the major factors that affect the flight time and the consumed energy. In order to deal with this problem, the idea is to (i) test a set of $n$ different combinations of weights, to control the connections of nodes in the generated MSTs, for every sub-region, and out of them to (ii) select the setup that generates the MST, which produces the minimum-turns path. In the context of this work, for every sub-region four different combinations are tested, which correspond to the ones that create MSTs connected in the upper, lower, most right and most left levels available. Figure \ref{fig:MSTsTurnReduction} shows an example with the four MSTs on a 3x3 grid for a better understanding.

\begin{figure}[!h]%
	\centering
	\subfigure[ROI coverage with 84 turns]{%
		\label{fig:non-turn-reduced-path}%
		\includegraphics[width=\linewidth]{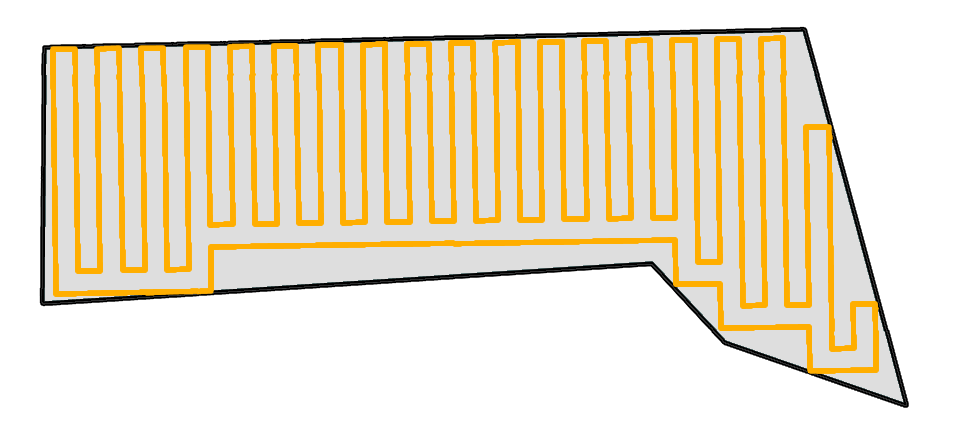}}%
	\qquad
	\subfigure[ROI coverage with 42 turns]{%
		\label{fig:turn-reduced-path}%
		\includegraphics[width=\linewidth]{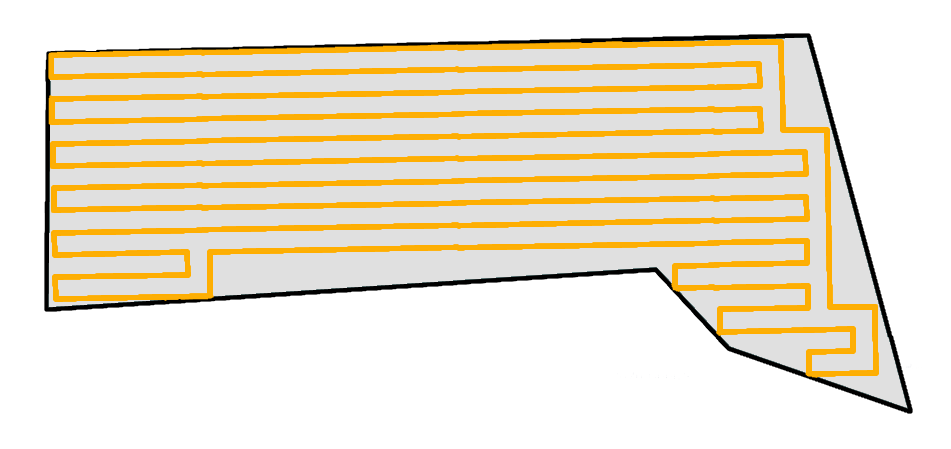}}%
	\caption{Turn reduction using the best MST orientation}
	\label{fig:turnReduction}
\end{figure}

Figure \ref{fig:turnReduction} presents an indicative example of the effect that the turn reduction mechanism has on the overall performance of the system. Without taking care of the path orientation the robot will have to follow a path with 84 turns, as shown in figure \ref{fig:non-turn-reduced-path}, whereas figure \ref{fig:turn-reduced-path} shows that for the same operational environment there is a path that completely covers the ROI with only 42 turns.

\section{Simulated Evaluation}
\label{sec:simulation}

This section presents a simulated evaluation of the proposed methodology, that is organized in two separate subsections. Specifically,  subsection \ref{subsec:CPP-eval} presents an evaluation of the introduced optimization procedure, by analyzing the contribution of each term to the overall quality of the generated paths. In addition to that, a comparison of the proposed method with one of the most powerful, in terms of complete coverage and support of very complex-shaped ROIs, state-of-the-art alternative CPP methods \cite{bahnemann2019revisiting} is performed. 
Building on top of that, subsection \ref{subsec:mCPP-study} includes a study in order to assess the multi-robot feature of the proposed methodology, comparing the results acquired for two fixed ROIs of different size, when increasing the number of UAVs used for coverage. For reproducibility reasons, all of the ROIs used for the simulated evaluations, along with the extensive list of results produced, are publicly accessible on-line\footnote{\url{https://github.com/savvas-ap/cpp-simulated-evaluations}}.

\subsection{Single-Robot Paths Evaluation}
\label{subsec:CPP-eval}

For the evaluations performed in subsections \ref{subsubsec:ablation} and \ref{subsubsec:comparison}, a set of 20 randomly generated polygons is used. This set includes convex and concave polygons, with and without obstacles and a wide range of areas. Both the proposed method and \cite{bahnemann2019revisiting} were executed with a fixed scanning density of 40 meters for all ROIs. The simulations were carried out with an also fixed altitude of 40 meters. Moreover, there was made the assumption that the UAV was equipped with an RGB sensor with a HFOV of $73.4^\circ$\footnote{This HFOV was selected as a typical specification for commercial UAVs, based on the sensor that one of the most popular commercial UAVs is equiped with (DJI phantom 4 pro: \url{https://www.dji.com/phantom-4-pro/info\#specs})}, in order to perceive the environment. The selected mission parameters, for this sensor, lead to collection of images with a percentage of overlap of 50\%, with their neighbor images.

\begin{figure}[!t]
	\centering
	\includegraphics[width=\linewidth]{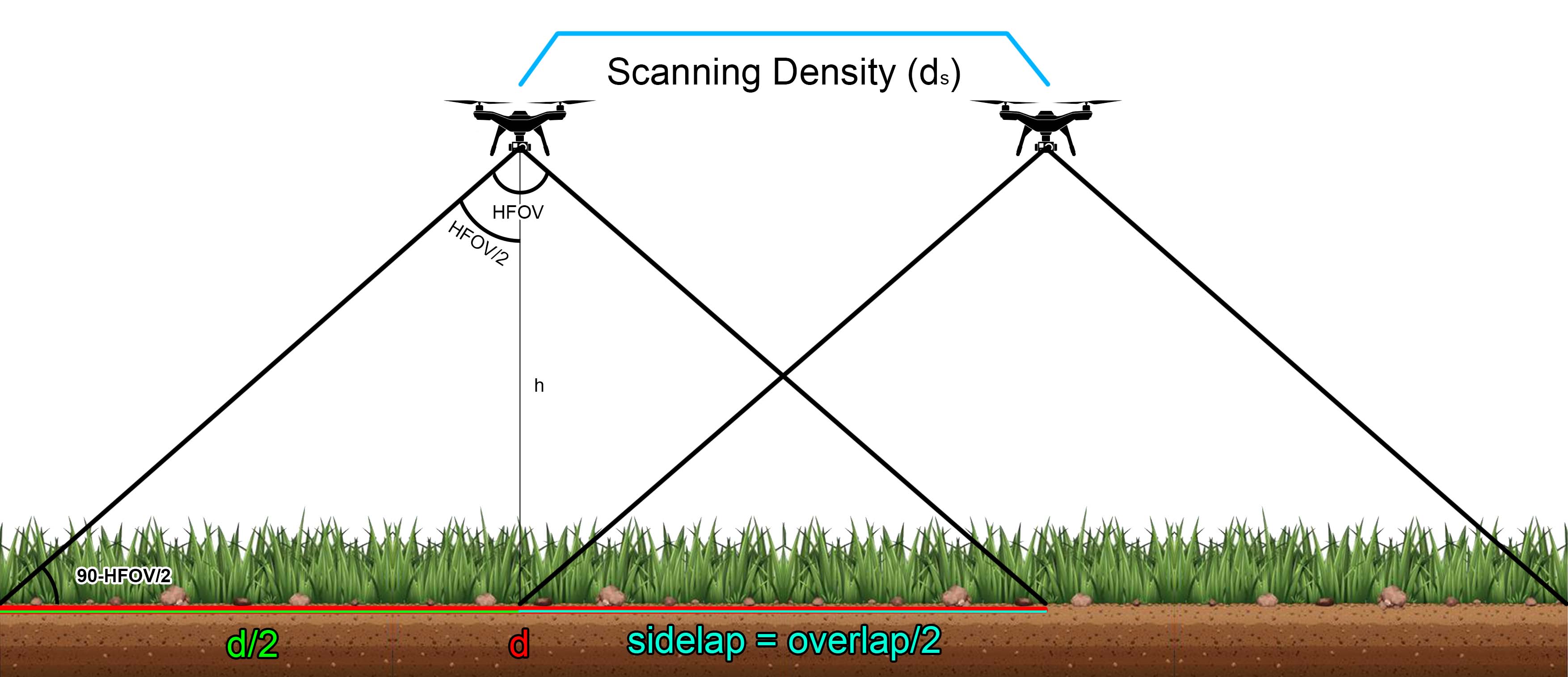}
	\caption{Land covered and overlap in the captured images}
	\label{fig:hovOverlap}
\end{figure}

In order to better understand the meaning of overlap, let's imagine two UAVs flying on the same altitude (h), the one next to the other, as shown in figure \ref{fig:hovOverlap}. Considering that the two UAVs are equipped with sensors with the same HFOV, the length of the ground covered by each, is:

\begin{equation}
\begin{split}
d = 2 \cdot \frac{h \cdot sin(HFOV/2)}{sin(90-HFOV/2)} \Rightarrow\\
d = 2 \cdot h \cdot tan(HFOV/2)
\end{split}
\label{eq:dHOV}
\end{equation}

Now, lets imagine that these two UAVs fly in an altitude and a distance between them, so that the ground covered by each of them, has a part that is also covered by the other one. In this case, it is considered that there is an overlap of information between the collected images. In the context of this paper, the term sidelap refers to the overlapping information between an image with its neighbor, in one of the two sides, while the term overlap refers to the overall overlapping information in both sides of the collected image, so that:
\begin{equation}
overlap = 2 \cdot sidelap
\label{eq:sidelap}
\end{equation}
In addition, there should be a separation of the terms \textit{overlap} and \textit{percentage of overlap $(p_o)$}. The term \textit{overlap} refers to distance, as explained above, while the term \textit{$p_o$} refers to the percentage of overall overlapping information at both sides of an image. The terms $d_s$, $h$, $p_o$ and $HFOV$ can be related with the following formula:

\begin{equation}
\begin{split}
d_s = d - sidelap = d - \frac{overlap}{2} = d - \frac{p_o \cdot d}{2} \Rightarrow\\
d_s = (2 - p_o) \cdot h \cdot tan(HFOV/2)
\end{split}
\label{eq:overlap}
\end{equation}

This way, the selected $p_o$ of 50\%, for the simulations, corresponds to a 50\% of  unique information in every collected image and 25\%, at each side, that is included in one of the two neighbor images. It is important to note that $p_o$ should be selected based on the targeted application (e.g., generation of orthomosaics, 3D maps, elevation models etc.).

For evaluation purposes, every ROI is considered to be consisted of discrete coverage cells, with an area of $0.25 - 4 m^2$, depending on the area of the overall ROI. After the simulation of a generated plan, for each coverage cell is counted the number of scans that was performed during the mission. Based on the number of scans performed, each cell gets assigned in one of the following categories:\\

\begin{itemize}
	\item 0: Not covered
	\item 1: Covered - unique coverage
	\item $\geq$ 2: Covered - overlapping coverage
\end{itemize}

Table \ref{table:metrics} describes all metrics that were used for the performance evaluation of the algorithms, on each different ROI. Out of these metrics, the \textit{PoOC}, the number of \textit{Turns} and the overall path \textit{Length} are considered to mainly describe the qualitative features of a generated path. In addition to them, there were created a \textit{heatmap of coverage} visualizing the number of visitations at each cell of a ROI and a \textit{normalized histogram of coverage}, showing the distribution of coverage cells according to the number of visitations performed. To better conceptualize the formation of these histograms, consider that for each coverage cell of a ROI, the number of scans performed at it, by multiple passages of the UAV, is counted. After the path simulation, a column is created for every count of performed scans (0, 1, 2, ...), with a height proportional to the number of cells that were visited that many times. To make the scale for the y axis independent of the size of the ROI, the histograms got normalized (the sum of all columns and the maximum possible value for the y axis is always 1). The results out of all ROIs are combined for the overall evaluation of the methods. Specifically, for the metrics presented in table \ref{table:metrics} are computed averages and, in addition, an aggregate histogram of coverage for every path planning method is created, by the combination of all coverage cells, out of all regions.

\begin{table}[!h]
	\def\arraystretch{1.2}
	\centering
	\resizebox{\linewidth}{!}{\begin{tabular}{ c | c  c } 
		\hline
		\textbf{Short} & \textbf{Metric} & \textbf{Unit} \\ 
		\hline
		PoC & Percentage of Coverage (Scans $\geq 1 $) & \% \\
		
		PoOC & Percentage of Overlapping Coverage (Scans $\geq 2 $) & \% \\
		
		Turns & Number of Turns/Waypoints & - \\
		
		N-Turns & Normalized Number of Turns & $Turns \over {1000 m^2}$ \\
		
		Length & Path Length & km \\
		
		N-Length & Normalized Path Length & $m \over {1000 m^2}$ \\
		\hline
	\end{tabular}}
	\caption{Evaluation metrics for subsection \ref{subsec:CPP-eval}}
	\label{table:metrics}
\end{table}

%


\subsubsection{Optimization Terms Importance Analysis}
\label{subsubsec:ablation}
In order to assess the contribution of each optimization term to the quantity of coverage and the overall quality of paths, an ablation study was performed. The goal was to realize how much and in which way, the addition of each term affects the ROI coverage task, by studying their effects on the evaluation metrics.

\begin{figure*}%
	\centering
	\subfigure[Non-Optimized STC]{%
		\label{fig:STC}%
		\includegraphics[height=1.58in]{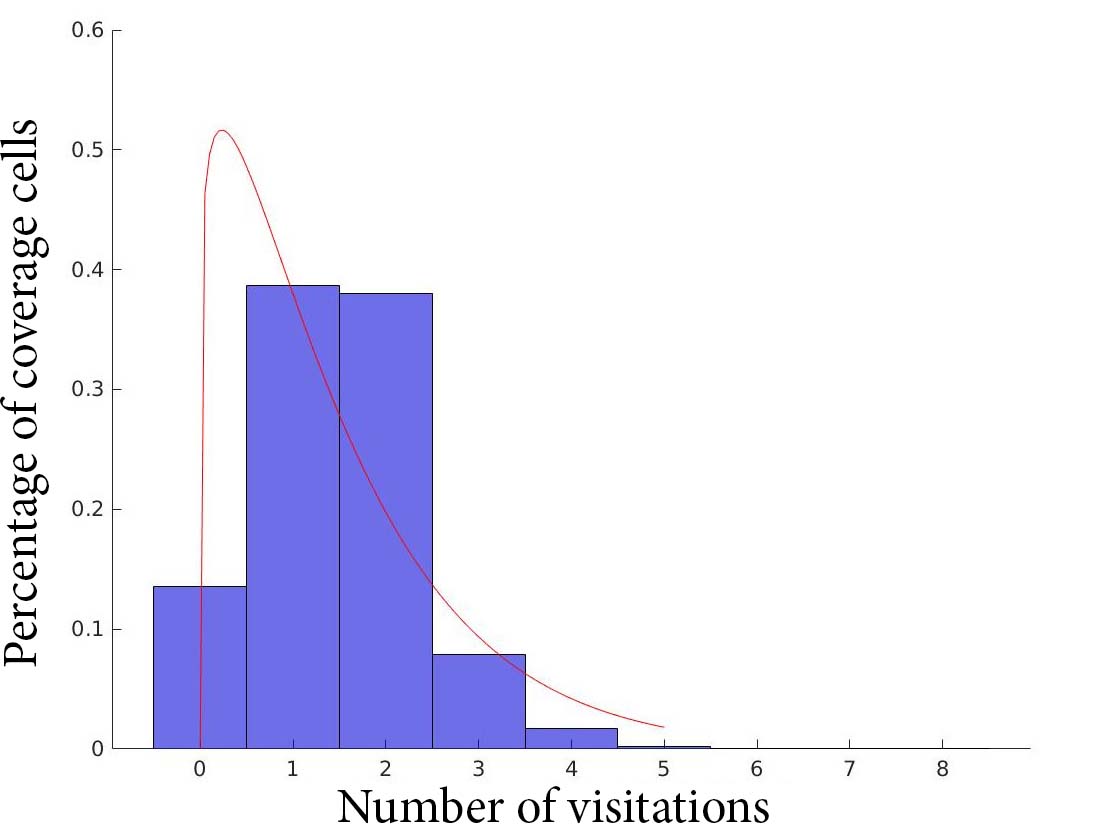}}%
	\qquad
	\subfigure[J1 Term Optimization]{%
		\label{fig:J1ovrl}%
		\includegraphics[height=1.58in]{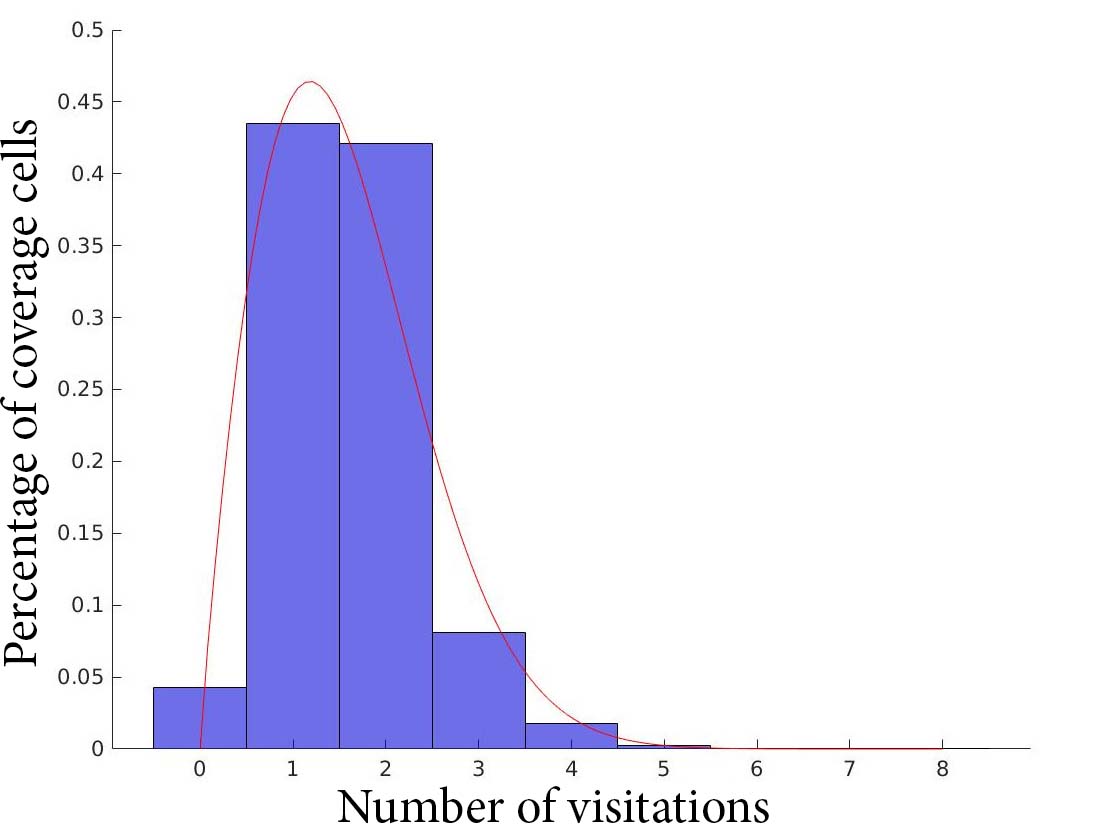}}%
	\qquad
	\subfigure[J1+J2 Terms Optimization]{%
		\label{fig:J1J2ovrl}%
		\includegraphics[height=1.58in]{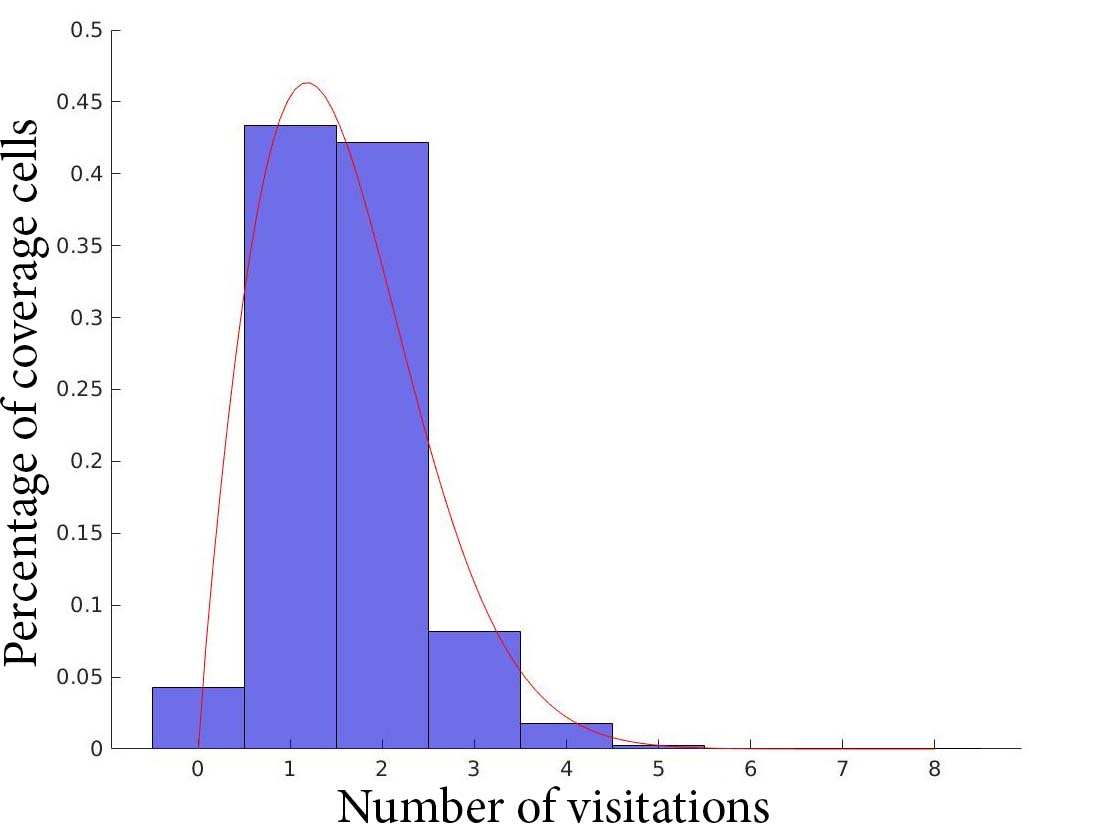}}%
	\qquad
	\subfigure[J1+J2+J3 Terms Optimization]{%
		\label{fig:J1J2J3ovrl}%
		\includegraphics[height=1.58in]{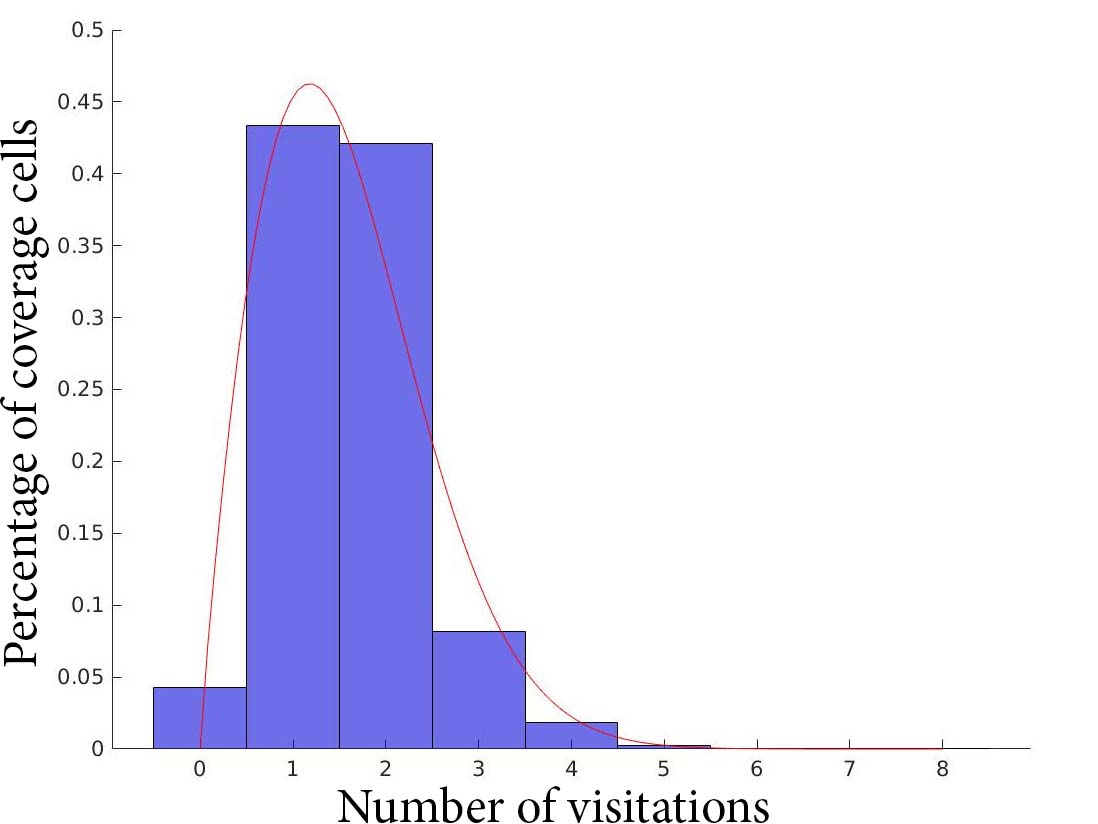}}%
	\qquad
	\subfigure[\cite{bahnemann2019revisiting} Length Reduction]{%
		\label{fig:ETHZLRovrl}%
		\includegraphics[height=1.58in]{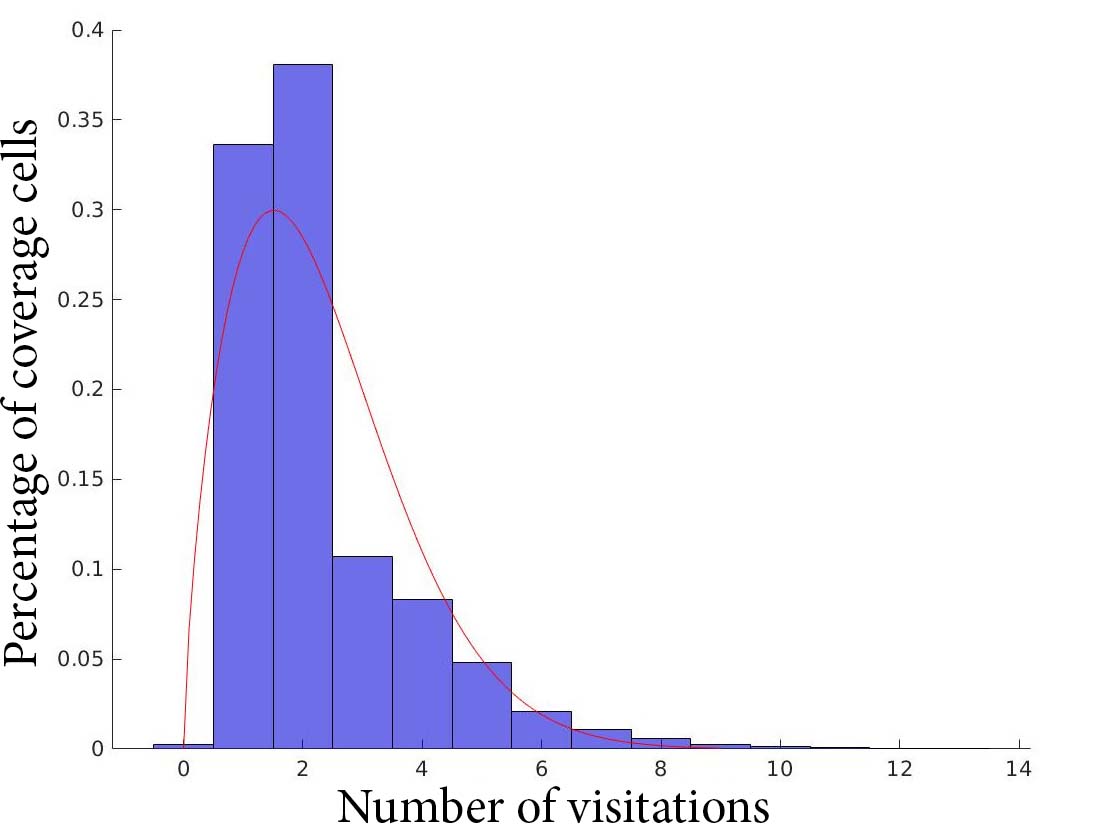}}%
	\qquad
	\subfigure[\cite{bahnemann2019revisiting} Turns Reduction]{%
		\label{fig:ETHZWPRovrl}%
		\includegraphics[height=1.58in]{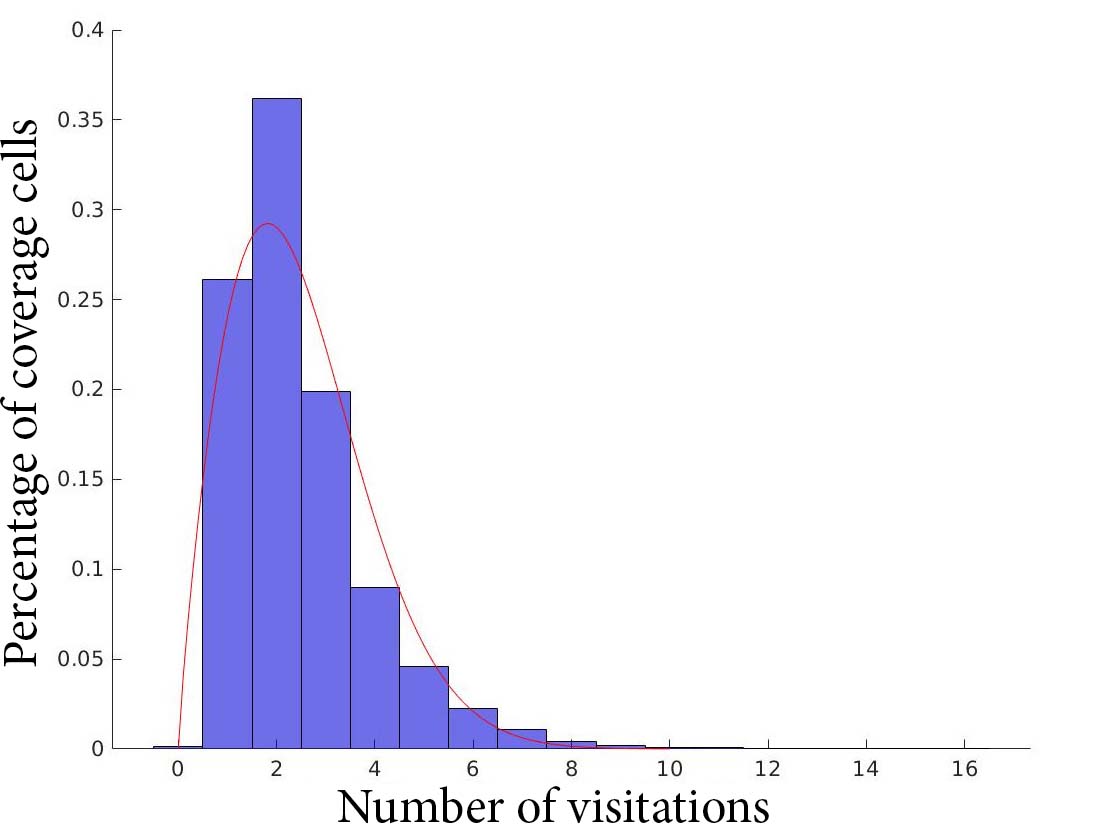}}%
	\caption{Aggregate normalized histograms of coverage}
	\label{fig:AvgEvMetr}
\end{figure*}

The first 4 columns of table \ref{table:AvgEvMetr} and the figures \ref{fig:STC}-\ref{fig:J1J2J3ovrl}, present the overall results of this study, as an average out of all ROIs. As the results make clear, the introduction of each term of the optimization procedure, helps to significantly increase the PoC and improve the quality of the generated paths. At this point, it is important to notice that the turn reduction procedure described in section \ref{subsec:MSTpaths}, is applied in all the approaches of STC presented below, independently of the node placement optimization. Comparing the results of the non-optimized STC with those of the proposed approach (J1+J2+J3 terms optimization), the average PoC is increased by 8.64\%, the average PoOC gets, in all cases, a value close to the user-defined $p_o$ (50\%), the average number of turns is reduced in comparison with the non-optimized STC, while the average length of paths does not get significantly increased, taking into account the benefit in terms of coverage.

\begin{figure*}[!t]
	\centering
	\subfigure[Non-Optimized STC]{%
		\label{fig:NOSTC}%
		\includegraphics[height=1.1in]{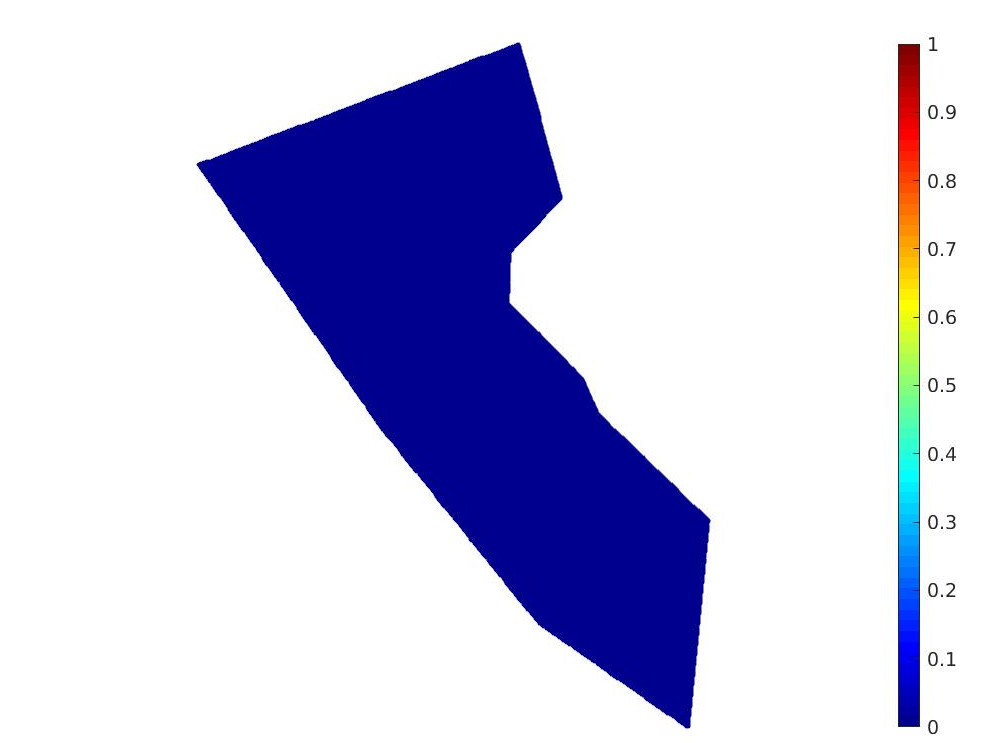}}%
	\qquad		
	\subfigure[J1 Term Optimization]{%
		\label{fig:J1}%
		\includegraphics[height=1.1in]{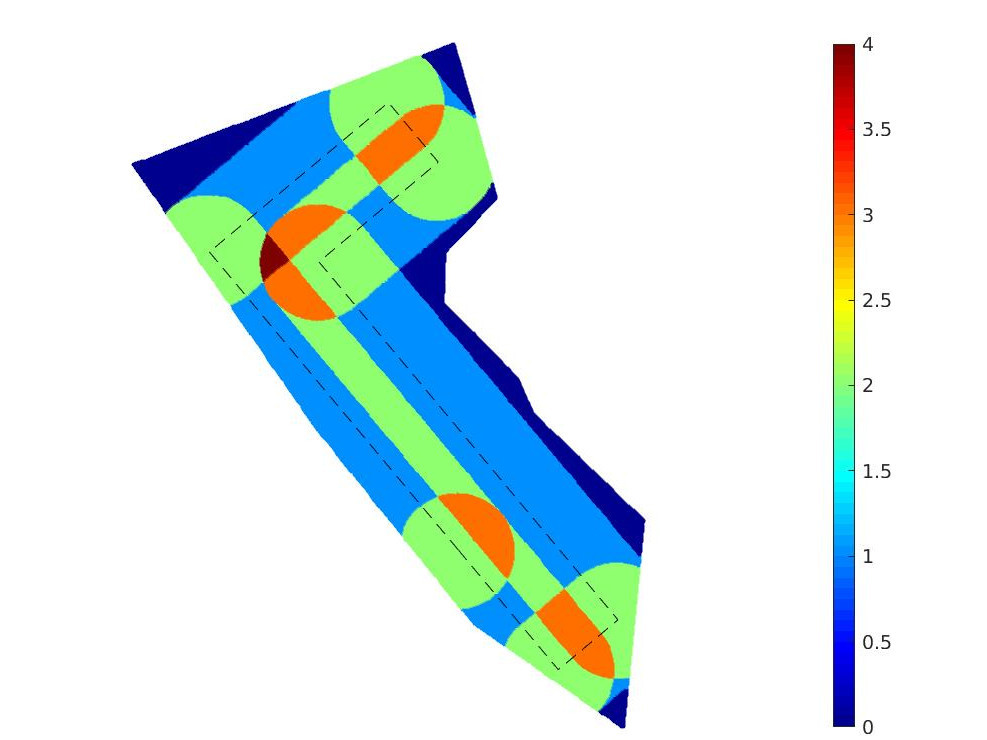}}%
	\qquad
	\subfigure[J1+J2 Terms Optimization]{%
		\label{fig:J1J2}%
		\includegraphics[height=1.1in]{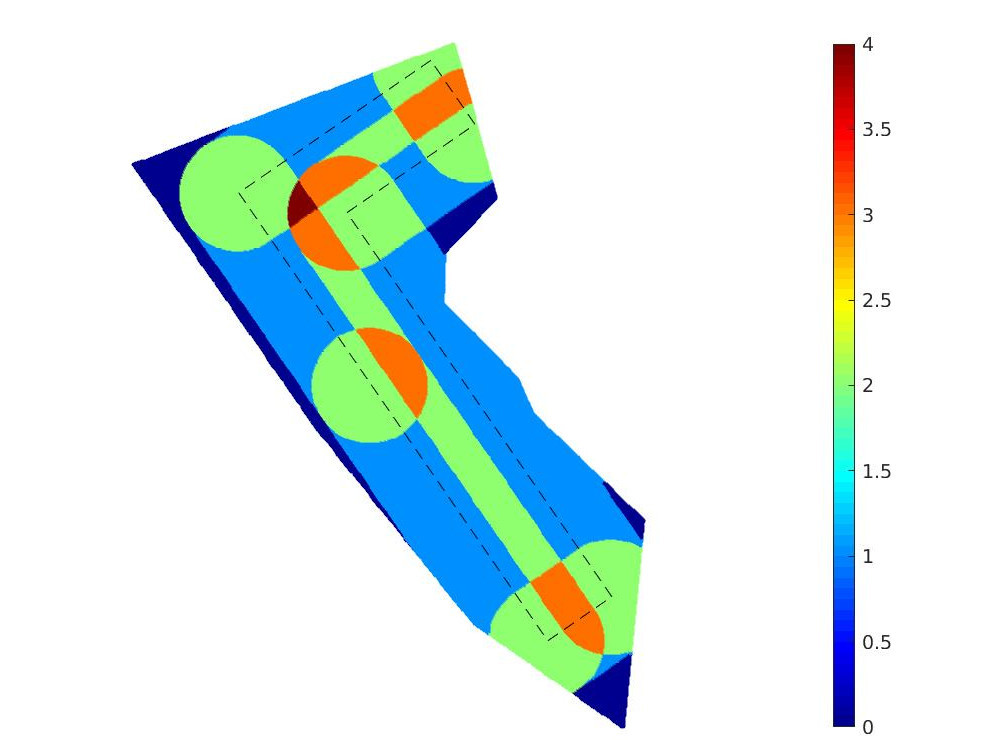}}%
	\qquad
	\subfigure[J1+J2+J3 Terms Optimization]{%
		\label{fig:J1J2J3}%
		\includegraphics[height=1.1in]{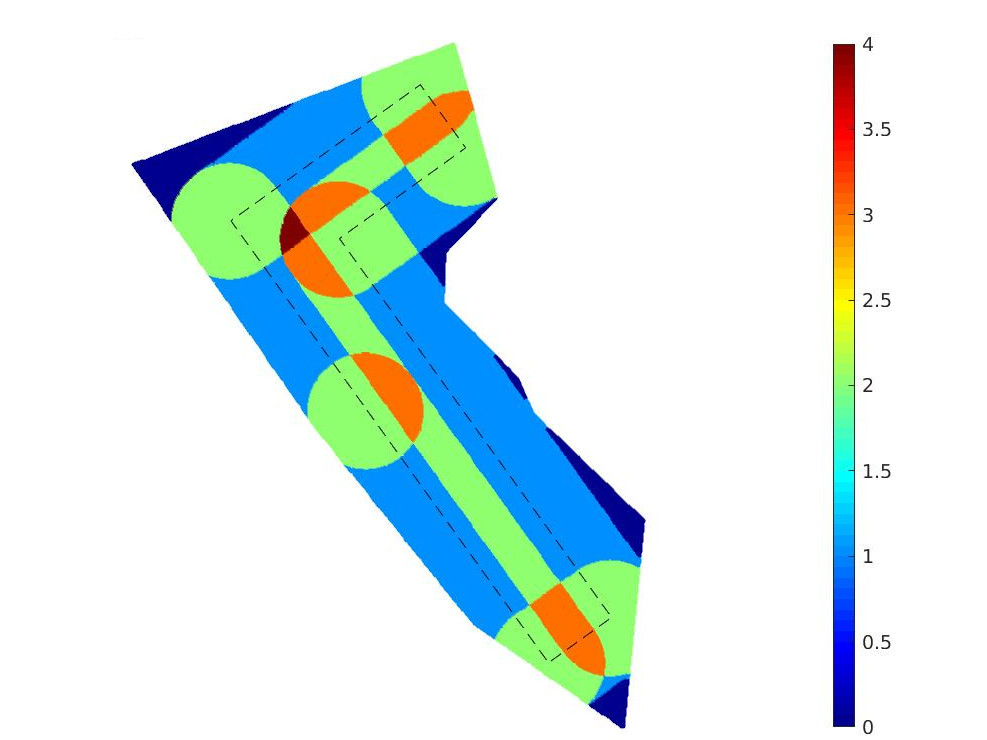}}%
	\qquad
	\subfigure[Non-Optimized STC]{%
		\label{fig:NOSTCHOC}%
		\includegraphics[height=1.13in]{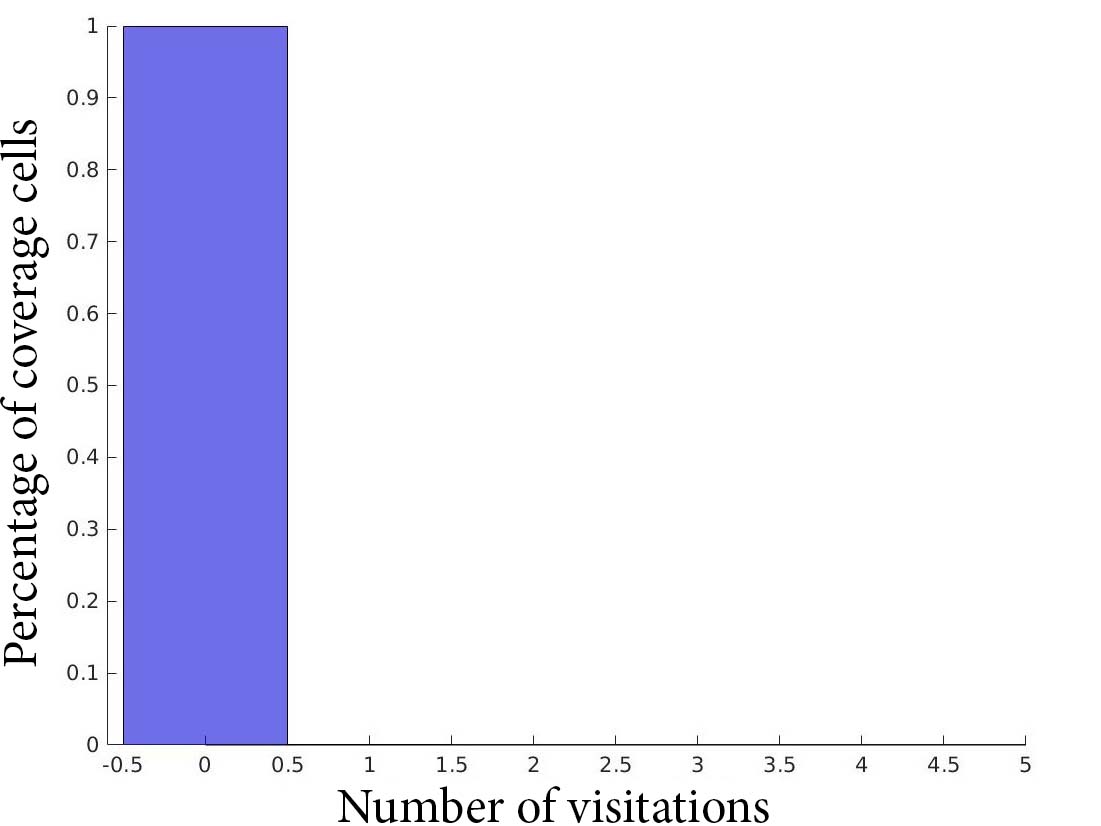}}%
	\qquad
	\subfigure[J1 Term Optimization]{%
		\label{fig:J1HOC}%
		\includegraphics[height=1.13in]{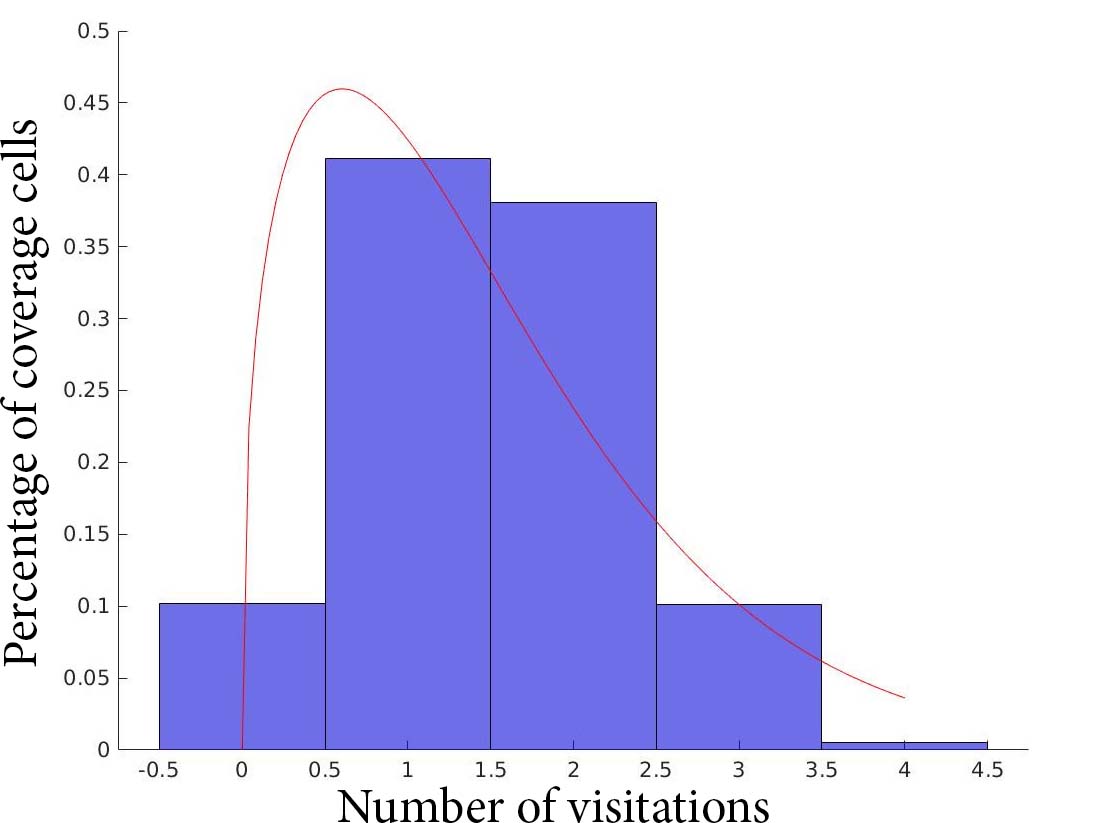}}%
	\qquad
	\subfigure[J1+J2 Terms Optimization]{%
		\label{fig:J1J2HOC}%
		\includegraphics[height=1.13in]{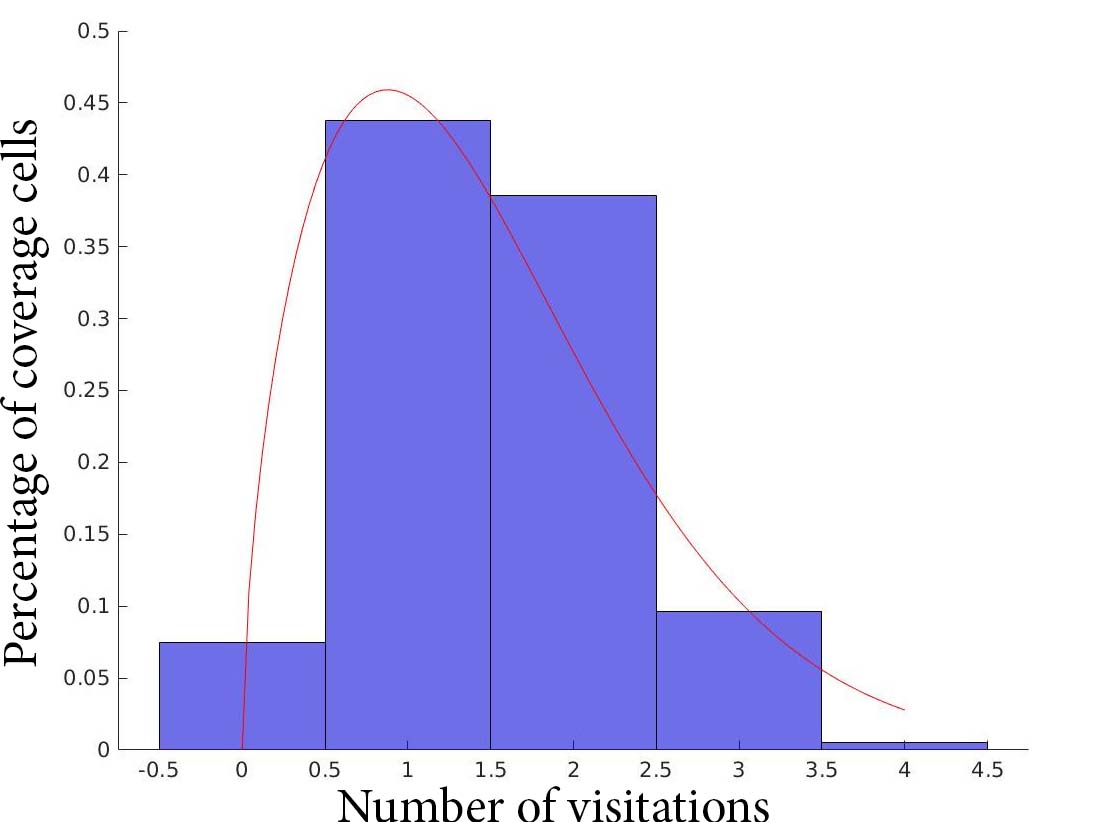}}%
	\qquad
	\subfigure[J1+J2+J3 Terms Optimization]{%
		\label{fig:J1J2J3HOC}%
		\includegraphics[height=1.13in]{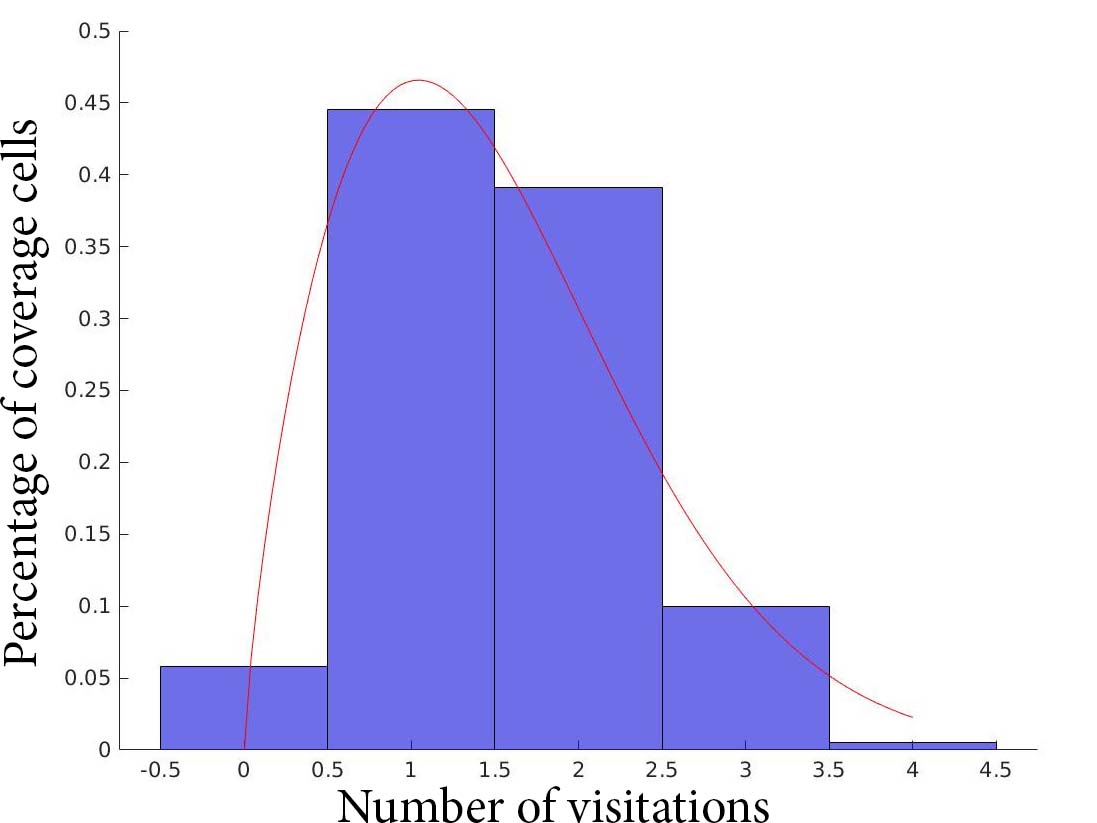}}%
	\caption{Optimization terms importance analysis - Testbet \#1}
	{\ref{fig:NOSTC}-\ref{fig:J1J2J3}: Heatmaps of coverage, \ref{fig:NOSTCHOC}-\ref{fig:J1J2J3HOC}: Histograms of coverage}
	\label{fig:ablationExmpl}
\end{figure*}

Figure \ref{fig:ablationExmpl} and table \ref{table:ablationExmpl} present the results for a specific ROI - testbed \#1. In testbed \#1, the implementation of the original STC algorithm \cite{gabriely2001spanning} could not run for the selected scanning density, without the proposed optimization scheme. However, the introduction of the proposed optimization procedure managed to find a grid tailored to the ROI's topology, so as to calculate paths, with the defined specifications, that provide decent coverage for the ROI. The first term, J1, is critical for the maximization of the length of paths that will fit in the ROI. At the same time, the other two terms, J2 and J3, are responsible for the optimal placement of paths inside the ROI. As a result, for this mission, that without optimization was considered non-executable, the proposed approach achieves a PoC larger than 94\% and PoOC pretty close to the user-selected $p_o$ (50\%).

\begin{table*}
	\def\arraystretch{1.2}
	\centering
	\resizebox{\textwidth}{!}{\begin{tabular}{ c | c c c | c | c c } 
		\hline
		\textbf{Average} & \textbf{Non-Optimized STC} & \textbf{J1 Term} & \textbf{J1+J2 Terms} & \textbf{J1+J2+J3 Terms} & \textbf{\cite{bahnemann2019revisiting} Length Reduction} & \textbf{\cite{bahnemann2019revisiting} Turns Reduction} \\ 
		\hline
		PoC & 87.73\% & 96.19\% & 96.02\% & \textbf{96.37\%} & 99.86\% & 99.91\% \\
		
		PoOC & 47.93\% & 51.96\% & 51.76\% & \textbf{51.74\%} & 64.85\% & 71.49\% \\
		
		Turns & 85.28 & 81.83 & 80.61 & \textbf{82.06} & 210.39 & 95.72 \\
		
		N-Turns & 0.09 & 0.09 & 0.09 & \textbf{0.09} & 0.21 & 0.11 \\
		
		Length & 22.24 & 24.12 & 24.12 & \textbf{24.04} & 33.95 & 43.15 \\
		
		N-Length & 20.81 & 23.95 & 23.93 & \textbf{23.90} & 33.77 & 40.94 \\
		\hline
	\end{tabular}}
	\caption{Average evaluation metrics}
	\label{table:AvgEvMetr}
\end{table*}

\begin{table*}
	\def\arraystretch{1.2}
	\centering
	\begin{tabular}{ c | c  c  c  c} 
		\hline
		\textbf{} & \textbf{Non-Optimized STC} & \textbf{J1 Term} & \textbf{J1+J2 Terms} & \textbf{J1+J2+J3 Terms} \\ 
		\hline
		PoC & - & 89.84\% & 92.54\% & \textbf{94.19\%} \\
		
		PoOC & - & 48.72\% & 48.78\% & \textbf{49.68\%} \\
		
		Turns & - & 8 & 8 & \textbf{8} \\
		
		N-Turns & - & 0.21 & 0.21 & \textbf{0.21} \\
		
		Length & - & 0.80 & 0.80 & \textbf{0.80} \\
		
		N-Length & - & 21.44 & 21.44 & \textbf{21.44} \\
		\hline
	\end{tabular}
	\caption{Optimization parameters importance analysis - Testbet \#1}
	\label{table:ablationExmpl}
\end{table*}

\begin{table*}
	\def\arraystretch{1.2}
	\centering
	\begin{tabular}{ c | c  c  c  } 
		\hline
		\textbf{} & \textbf{J1+J2+J3 Terms} & \textbf{\cite{bahnemann2019revisiting} Length Reduction} & \textbf{\cite{bahnemann2019revisiting} Turns Reduction} \\ 
		\hline
		PoC & \textbf{95.18\%} & 99.97\% & 99.05\% \\
		
		PoOC & \textbf{56.97\%} & 75.41\% & 81.15\% \\
		
		Turns & \textbf{84} & 266 & 109 \\
		
		N-Turns & \textbf{0.15} & 0.48 & 0.20 \\
		
		Length & \textbf{13.12} & 21.68 & 28.99 \\
		
		N-Length & \textbf{23.90} & 39.49 & 52.79 \\
		\hline
	\end{tabular}
	\caption{Comparison with state-of-the-art - Testbed \#2}
	\label{table:compExmpl}
\end{table*}

\begin{figure*}[!h]
	\centering
	\subfigure[J1+J2+J3 Terms Optimization]{%
		\label{fig:optExComp}%
		\includegraphics[height=1.58in]{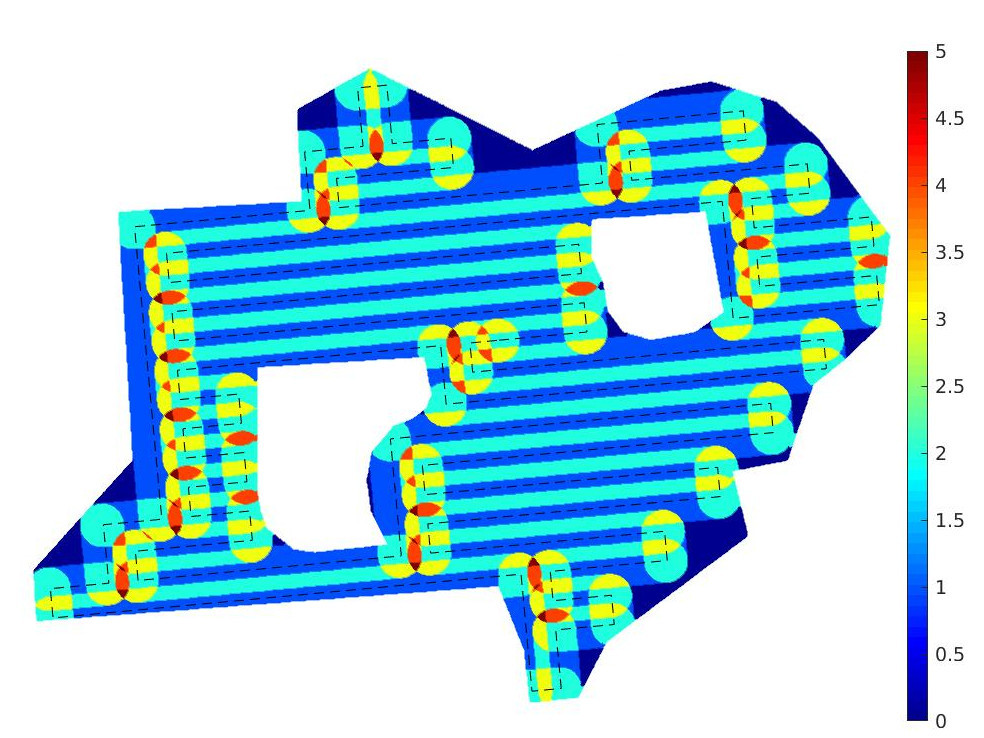}}%
	\qquad
	\subfigure[\cite{bahnemann2019revisiting} Length Reduction]{%
		\label{fig:ETHLREx}%
		\includegraphics[height=1.58in]{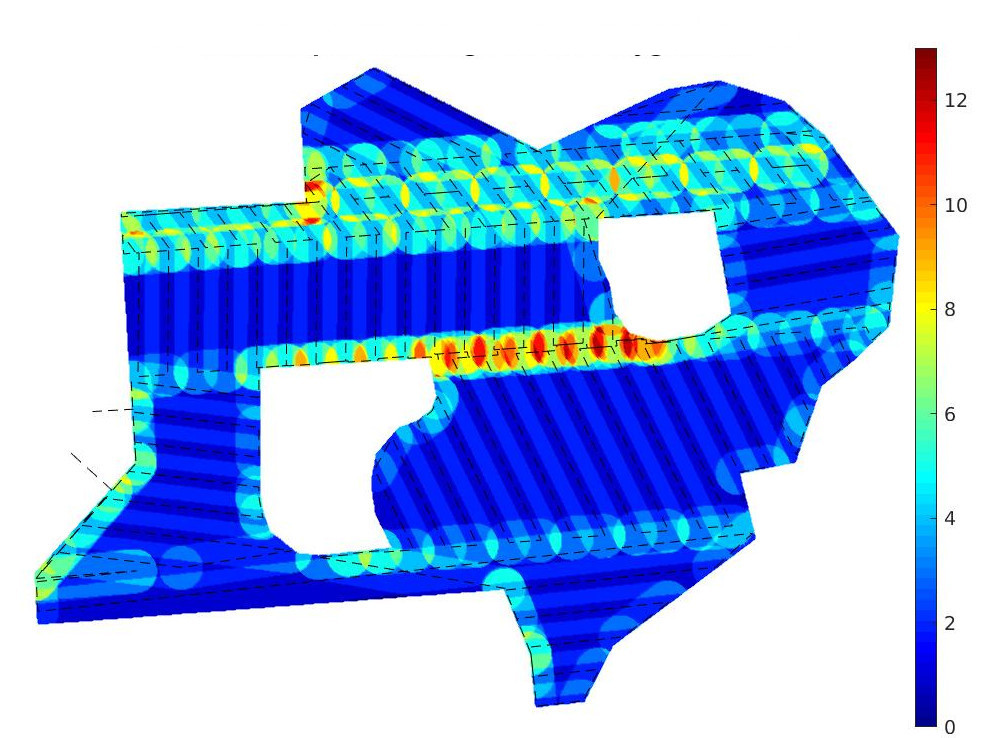}}%
	\qquad
	\subfigure[\cite{bahnemann2019revisiting} Turns Reduction]{%
		\label{fig:ETHWPREx}%
		\includegraphics[height=1.58in]{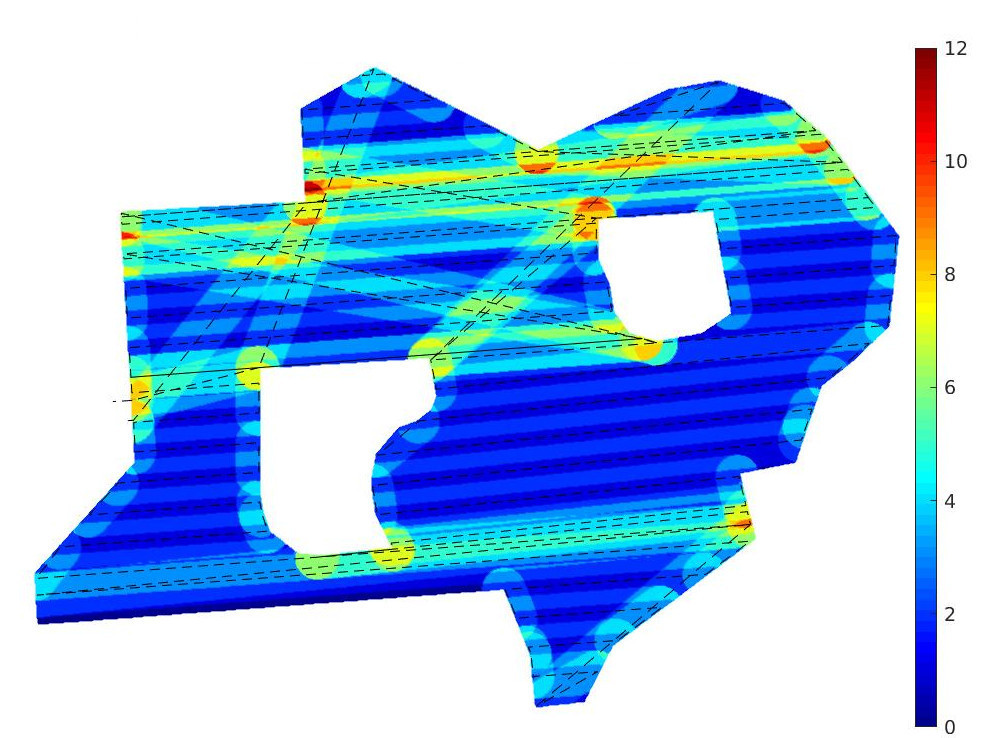}}%
	\qquad
	\subfigure[J1+J2+J3 Terms Optimization]{%
		\label{fig:optExCompHOC}%
		\includegraphics[height=1.58in]{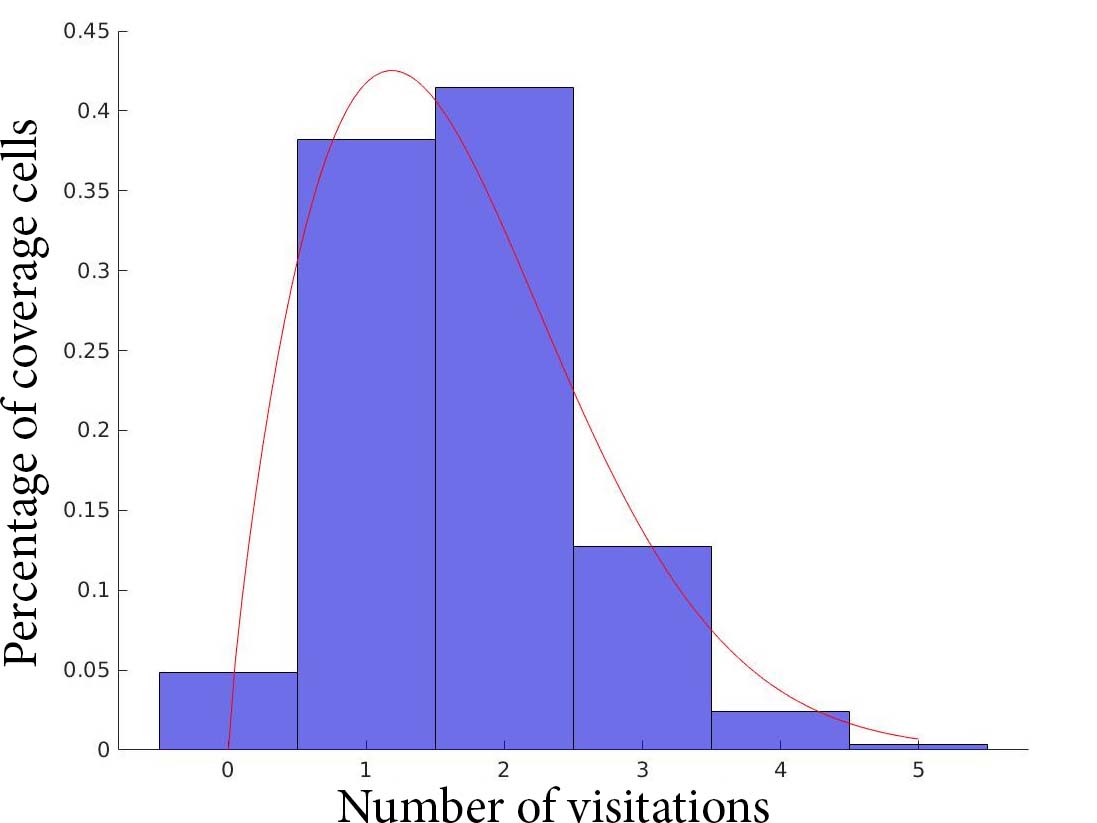}}%
	\qquad
	\subfigure[\cite{bahnemann2019revisiting} Length Reduction]{%
		\label{fig:ETHLRExHOC}%
		\includegraphics[height=1.58in]{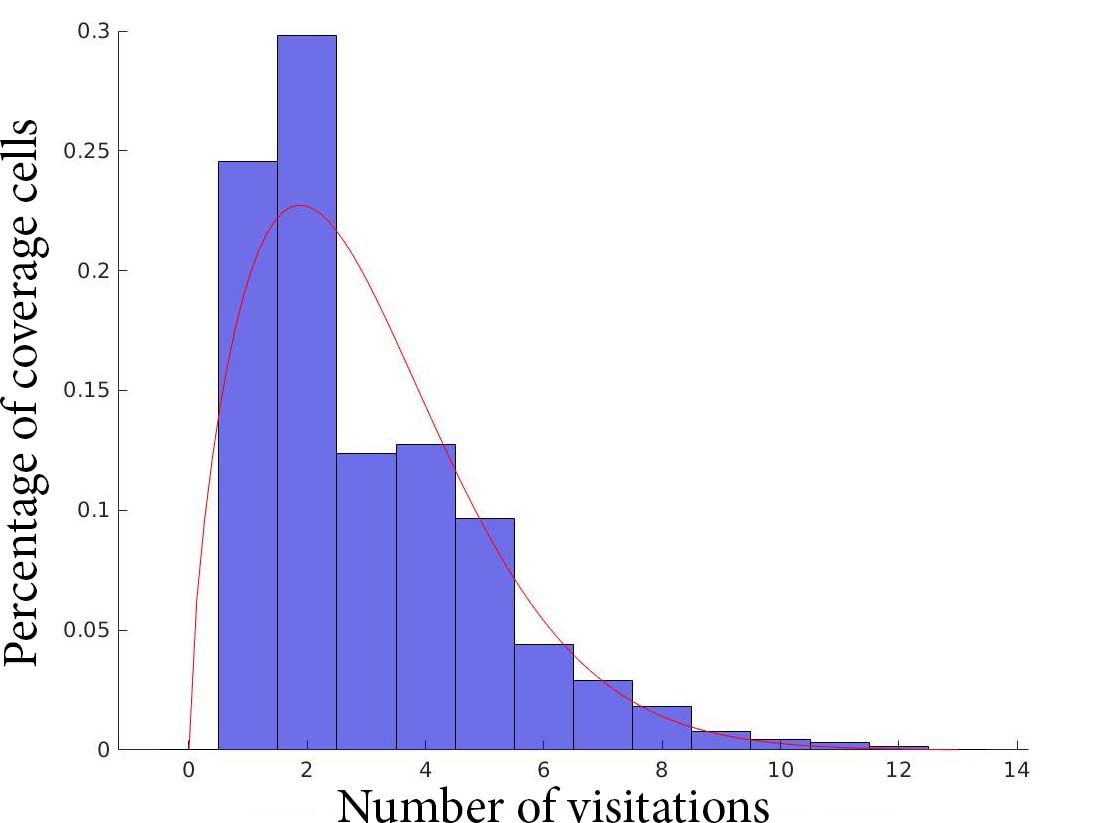}}%
	\qquad
	\subfigure[\cite{bahnemann2019revisiting} Turns Reduction]{%
		\label{fig:ETHWPRExHOC}%
		\includegraphics[height=1.58in]{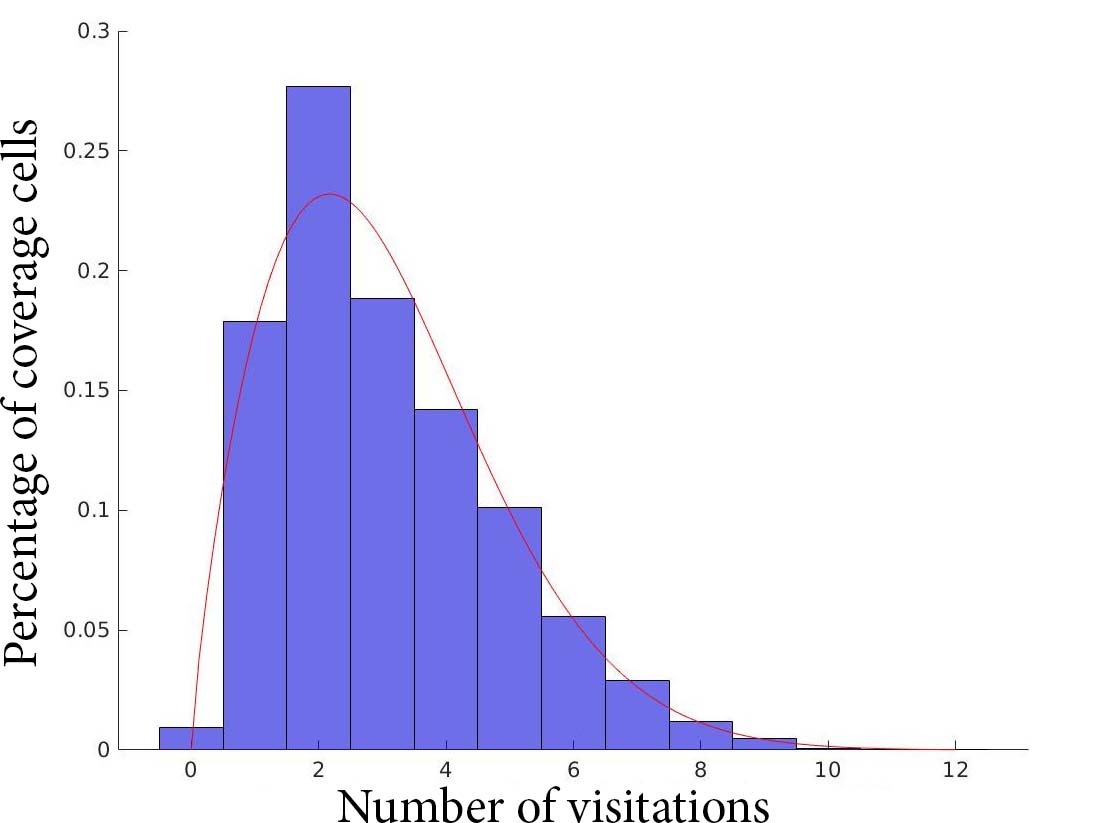}}%
	\caption{Comparison with state-of-the-art - Testbed \#2}
	{\ref{fig:optExComp}-\ref{fig:ETHWPREx}: Heatmaps of coverage, \ref{fig:optExCompHOC}-\ref{fig:ETHWPRExHOC}: Histograms of coverage}
	\label{fig:compExmpl}
\end{figure*}

\subsubsection{Comparison with State-of-the-Art}
\label{subsubsec:comparison}

\ref{subsubsec:ablation} shows that the proposed optimization procedure significantly improves the effectiveness of the method in comparison with the non-optimized one, in terms of coverage. The goal of this subsection is to investigate how the proposed approach stands next to one of the most powerful, in terms of complete coverage, state-of-the-art methods. For this reason, coverage plans following the procedure described in \cite{bahnemann2019revisiting} are also calculated for all ROIs, with two different objectives (cost functions), one intending to minimize the length of paths and one intending to minimize the number of turns.

The overall results of this comparison are presented in figures \ref{fig:J1J2J3ovrl}-\ref{fig:ETHZWPRovrl} and the three last coloumns of table \ref{table:AvgEvMetr}. The method proposed in \cite{bahnemann2019revisiting} achieved almost complete coverage for all ROIs, with both cost functions, with an average close to $100\%$. On the other hand, the proposed method achieves a smaller, but still descent PoC ($96.37\%$ - i.e. $3.54\%$ smaller than the turn reduction mode and $3.49\%$ smaller than the length reduction mode of \cite{bahnemann2019revisiting}). However, both the number of turns and the length of produced paths, by the proposed method, are significantly smaller in comparison with both, dedicated to these purposes versions of \cite{bahnemann2019revisiting}. To be more precise, the proposed approach achieved a number of average turns 2.56  and 1.16 times smaller than the length reduction and turns reduction modes of \cite{bahnemann2019revisiting}. In addition, it also achieved an average path's length 1.41 and 1.79 times smaller than the two, aforementioned versions of \cite{bahnemann2019revisiting}, respectively. This way, the proposed approach combines the advantages of both these dedicated versions, in a more efficient way. Moreover, the PoOC is a lot closer to the user-selected $p_o$, meaning that there are not performed redundant scans that waste operational resources. Finally, figures \ref{fig:J1J2J3ovrl}-\ref{fig:ETHZWPRovrl} show that, with the proposed approach, most of the coverage cells are scanned 1 to 2 times and the maximum number of scans for the same cell is 8, while with \cite{bahnemann2019revisiting} most of the coverage cells are scanned 1 to 4 times and the maximum number of scans for the same cell is 14 (figure \ref{fig:ETHZLRovrl}) and 16 (figure \ref{fig:ETHZWPRovrl}) respectively.

Figure \ref{fig:compExmpl} and table \ref{table:compExmpl} present the results for a specific ROI - testbed \#2. \cite{bahnemann2019revisiting} manages to achieve an average PoC over $99\%$, with both cost functions. The average PoC achieved by the proposed method is smaller, but still descent and appropriate for real-life use (over $95\%$). However, once again, the paths generated by the proposed method are a lot more energy-aware. To be more specific, the number of turns is 3.17 and 1.29 times smaller, and the overall path length is 1.65 and 2.21 times smaller, than the length reduction and the turns reduction modes of \cite{bahnemann2019revisiting}, respectively. In addition, the count of redundant scans performed with the proposed methodology is significantly smaller as well, managing to keep a percentage of overlapping coverage close to the user-defined $p_o$ (50\%).

While the difference in the average PoC between the proposed method and \cite{bahnemann2019revisiting} is not negligible, the PoC achieved by the proposed methodology is more than high-enough to face the majority of challenges faced in real-life operations (e.g., \cite{bochkarev2016minimizing}, \cite{stolfi2020cooperative}, \cite{paradzik2016multi}, \cite{kapoutsis2015real}). In contrast, it is also worth noting for \cite{bahnemann2019revisiting} that in order to achieve almost complete coverage, it forces the vehicles to revisit multiple times specific parts of the ROIs, as shown in figures \ref{fig:ETHLREx} and \ref{fig:ETHWPREx}. In addition, both the energy-efficiency features and the multi-UAV capability, that can significantly reduce the operational time, provided by the proposed method, are strong incentives to prefer it in real-life operations.

%
%
%
%
%

\begin{table*}[!h]
	\def\arraystretch{1.2}
	\centering
	\resizebox{.8\linewidth}{!}{\begin{tabular}{ c | c  c } 
			\hline
			\textbf{Short} & \textbf{Metric} & \textbf{Unit} \\ 
			\hline
			PoC & Percentage of Coverage (Scans $\geq 1 $) & \% \\
			
			PoOC & Percentage of Overlapping Coverage (Scans $\geq 2 $) & \% \\
			
			\#Bat/UAV & Number of batteries needed for each UAV to complete the mission & - \\
			
			Flight Time & Theoretical time demanded to complete the mission & min \\
			
			Deployment Time & Time demanded to deploy the gear & min \\
			
			Change Battery Delay & Time needed to leave a spot, change battery and continue the mission & min \\
			
			Total time & Flight Time, including gear deployment and batteries changes & min \\
			
			Flight Cost & Estimated cost of the flight & Euro \\
			
			\hline
	\end{tabular}}
	\caption{Evaluation metrics for subsection \ref{subsec:mCPP-study}}
	\label{table:metrics2}
\end{table*}

\subsection{Multi-Robot Marginal Utility Study}
\label{subsec:mCPP-study}

In order to study the effects of using multiple-UAVs in coverage missions with the proposed algorithm, a series of simulations was performed. The performed study includes two ROIs with strategically different areas, for a varying number of UAVs, starting from 1 and up to 15. The simulations were carried out with fixed scanning density of 47 meters, altitude of 60 meters and the same assumption regarding the RGB sensor of all UAVs was made ($73.4^\circ$ HFOV). As a result, the percentage of overlap with the adjacent images was 90\%. Finally, the speed for all the UAVs was set to be constant at 3m/s.

Table \ref{table:metrics2} shows the metrics that were calculated for the multi-robot study. PoC and PoOC are calculated just as described in subsection \ref{subsec:CPP-eval}. The rest of them are explained below.

In order to calculate the \textit{\#Bat/UAV} and the \textit{Flight Time (FT)}, \textit{Flight Duration} is calculated first, as follows:
\begin{equation}
FlightDuration = \frac{Length}{Speed} + Turns \cdot \frac{c_1 \cdot Speed}{c_2 + |Speed|}
\label{eq:fligthDuration}
\end{equation}
The first term corresponds to the time that would be needed in case that the mission was executed with a constant speed, while the second term corresponds to a delay added to this time, for each of the turns in the path. $c_1$, $c_2$ have constant values and they should be chosen accordingly, in order to fine-tune this sigmoid function based on each UAV's type and flight behavior.

The \textit{\#Bat/UAV} refers to the number of the batteries needed for each UAV in order to complete its path, considering that a single battery can provide 25 minutes of flight duration.
\begin{equation}
\#Bat/UAV = \lceil\frac{FlightDuration}{25}\rceil,
\label{eq:batNum}
\end{equation}
where and $\lceil r \rceil$ denotes the ceiling of $r$ (smallest integer that is not smaller than r).

\textit{Flight Time} refers to the maximum of the \textit{Flight Duration}s', for all UAVs participating in a mission. This way:
\begin{equation}
Flight Time = argmax[FlightDuration(v_i)],
\label{eq:FlightTime}
\end{equation}
where $i \in [1,v_n]$.

\textit{Deployment Time (DT)} refers to the indicative time needed to deploy the gear for an experiment (e.g., GCS, UAVs, etc.), from a single human operator. The \textit{Deployment Time} is considered to be 5 minutes, increased by 3 minutes for each of the UAVs participating, and is calculated as follows:
\begin{equation}
Deployment Time = 5 + 3 \cdot v_n
\label{eq:deploymentTime}
\end{equation}

The \textit{Change Battery Delay (CBD)} refers to the time that would be needed if, when a path for a UAV could not be executed with a single battery, the UAV would stop the execution of the mission, fly back to the home position to change battery and return back to the point it stopped the mission to continue. For this delay it is considered that each UAV would need to cross twice an indicative distance, proportional to the area of the ROI, for each battery change that should take place ($\#Bat/UAV > 1$) and that the battery change would also add 3 minutes of delay. Specifically:
\begin{equation}
CBD = (\#Bat/UAV -1) \cdot (2 \cdot \frac{\sqrt{area_{ROI}}} {3 \cdot Speed \cdot 60} + 3 \cdot v_n   )
\label{eq:batDelay}
\end{equation}

The \textit{Total Time} comes out from the sum of the \textit{Flight Time}, \textit{Deployment Time} and \textit{Change Battery Delay}.
\begin{equation}
Total Time = FT + DT + CBD
\label{eq:timeTotal}
\end{equation}

Finally, the \textit{Flight Cost (FC)} refers to the estimated cost of a flight. Given the KW price of the national power supplier of Greece, the battery capacity and the flight duration of a common commercial UAV\footnote{\url{https://www.dji.com/phantom-4-pro}}, the \textit{Flight Cost per Minute (FCM)} is considered to be:
\begin{equation}
FCM = 0.017228 \; Euro/min
\label{eq:FCM}
\end{equation}
For the calculation of the \textit{Flight Cost}, \textit{FCM} is multiplied by the \textit{Total Time}, excluding the \textit{Deployment Time} and the time needed to change batteries (however not the time needed to cross the distance in order to change batteries), for each UAV. Specifically:
\begin{equation}
\begin{split}
Flight Cost = (Total Time - DT - \\ (\#Bat/UAV-1) \cdot 3) \cdot v_n \cdot FCM
\end{split}
\label{eq:flightCost}
\end{equation}

It should be noted that the \textit{Flight Cost} calculation does not intend to provide a thorough, or precise economic analysis. The intention of calculating this quantity is to provide an indicative measurable clue, about the productivity gain compared to the increasing operational cost of deploying multiple UAVs. However, one should always have in mind that this operational cost, most of the times is minimal compared to the investment cost that is demanded for multiple UAVs. For a proper economic analysis to take place, many other factors should be specified, with the most important of them being the field of application. For example, depending on the specific needs of a type of mission, the investment cost for a single UAV, along with the appropriate sensors, could significantly vary (approximately 500-30.000 euros). In addition, there are cases where the significant time reduction will also reduce the demanded human-hours to perform a task (e.g., scan huge agricultural fields), having a great impact on the human-cost and as a result on the investment's payback time as well. On the contrary, there are cases where the performance gains could not justify the investment cost, such as mapping missions deployed by hobbyists, or even professionals who rarely perform coverage tasks. Finally, there are also cases where the performance gains cannot be directly contradicted to a mission's cost, such as the search and rescue missions. In this specific application, the increasing number of UAVs does not only reduces the demanded time to scan a ROI and probably detect objects of interest (e.g., injured humans), but also increases the probability of detecting moving targets (e.g., wandering humans in risk) and saving human lives. Overall, this part of the multi-robot study intends to provide a clue about the value of deploying multiple UAVs in coverage missions, however, it cannot be considered in any case as a proper economic analysis.

\begin{figure}[!b]%
	\centering
	\includegraphics[width=\linewidth]{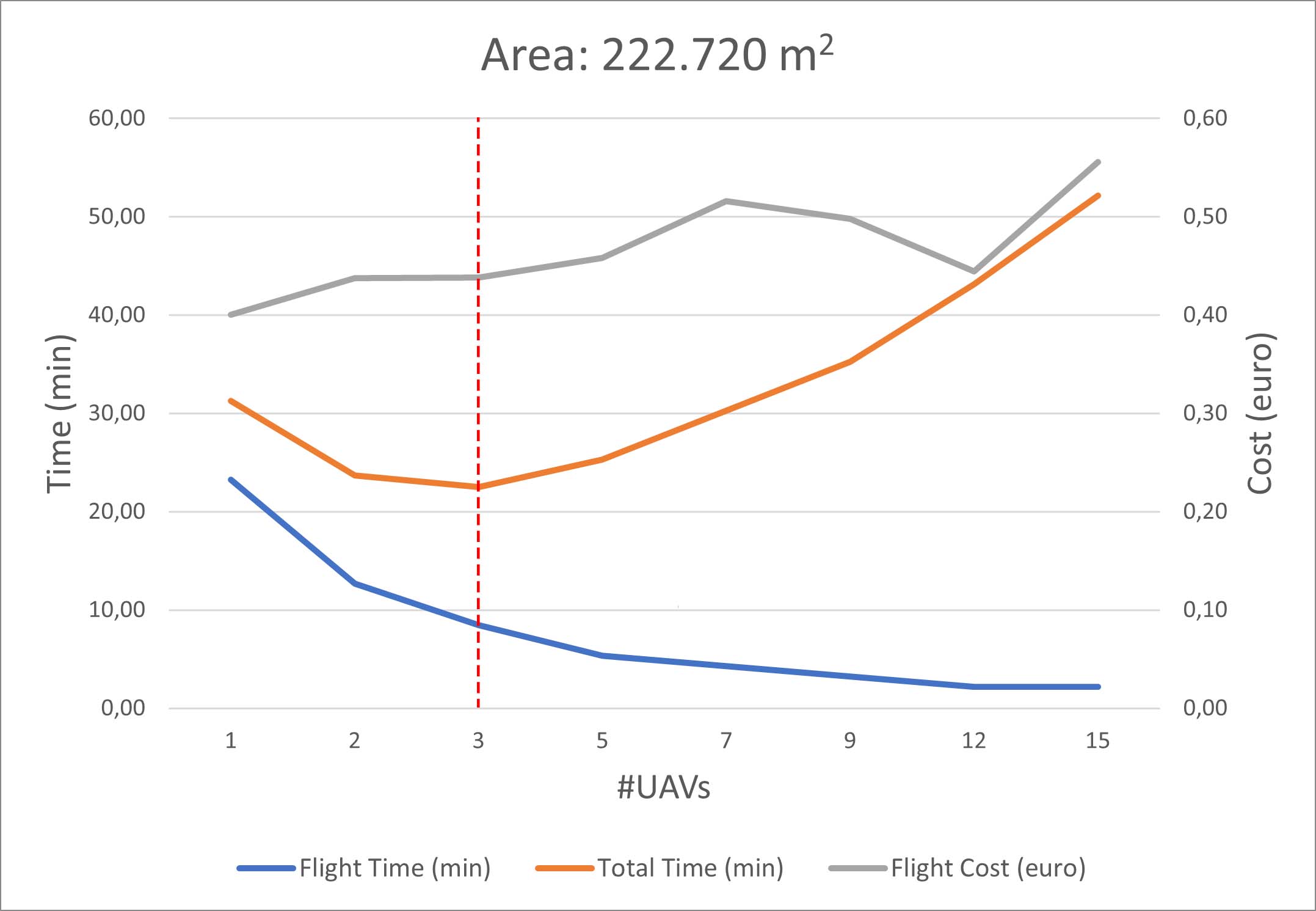}
	\caption{Mission times and flights' cost - Testbed \#3} 
	\label{fig:mCPP_study-3}
\end{figure}

\begin{figure*}[!h]
	\centering
	\subfigure[Coverage with 1 UAV]{%
		\label{fig:small1}%
		\includegraphics[width = .3\linewidth]{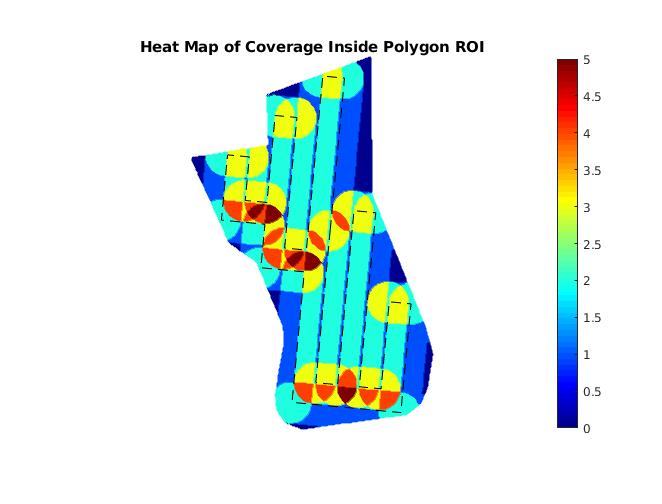}}%
	\qquad
	\subfigure[Coverage with 3 UAVs]{%
		\label{fig:small3}%
		\includegraphics[width = .3\linewidth]{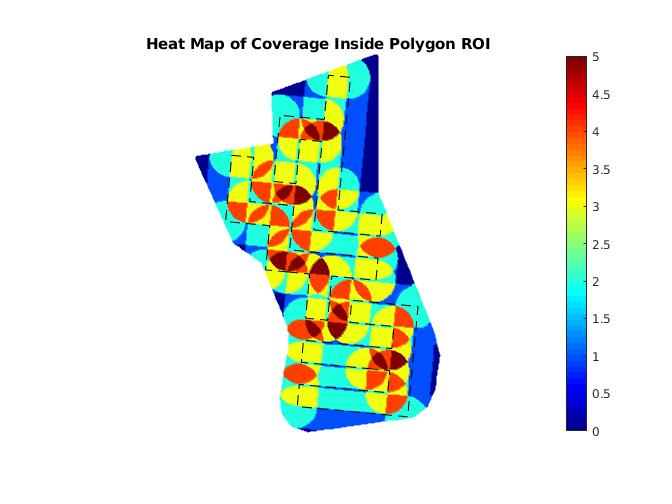}}%
	\qquad
	\subfigure[Coverage with 15 UAVs]{%
		\label{fig:small15}%
		\includegraphics[width = .3\linewidth]{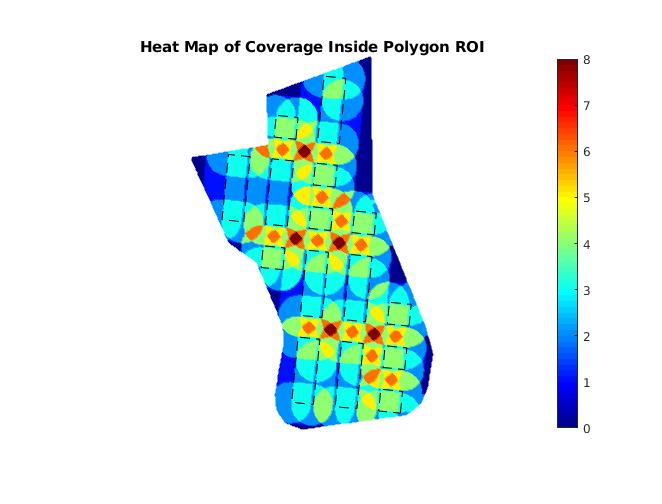}}%
	\caption{Heatmaps of coverage - Testbed \#3}
	\label{fig:mCPP-smallROI}
\end{figure*}

Table \ref{table:mCPP-SmallROI} and figures \ref{fig:mCPP_study-3} and \ref{fig:mCPP-smallROI} include the results of this multi-robot study for testbed \#3. The area of this ROI is 222.720 $m^2$, which given the selected mission parameters, is considered to be of a small-average size for coverage missions. From the acquired results, it is clear that increasing the number of UAVs does not significantly affect the achieved PoC. However, it has a slight impact on the PoOC, which has an increasing tendency, when increasing the number of vehicles. In addition, increasing the UAV's number has also a slight impact to the maximum times of revisiting certain points of the ROI, as shown in figure \ref{fig:mCPP-smallROI}. The generated mission for a single UAV, in this ROI, was executable with a single battery (Flight Time$<$25 min). As expected, the Flight Time constantly decreases when deploying additional vehicles for the coverage mission, however, the Total Time, which also includes the Deployment Time, has a minimum value when performing the mission with 3 UAVs. The addition of the second and the third UAV cause an increase on the Flight Cost by 0.04 Euro, in both cases, compared to the single-UAV deployment. Overall, the Flight Cost for this ROI has a global minimum value when the mission is performed with a single UAV and an increasing tendency when adding multiple vehicles (with a local minimum for 12 UAVs). The efficiency gains do not justify the deployment of more than 3 UAVs for this mission, while the Total Time decrease by 8.77 min, could easily justify the Flight Cost increase by 0.04 Euro, for this purpose.

\begin{figure}[!b]%
	\centering
	\includegraphics[width=\linewidth]{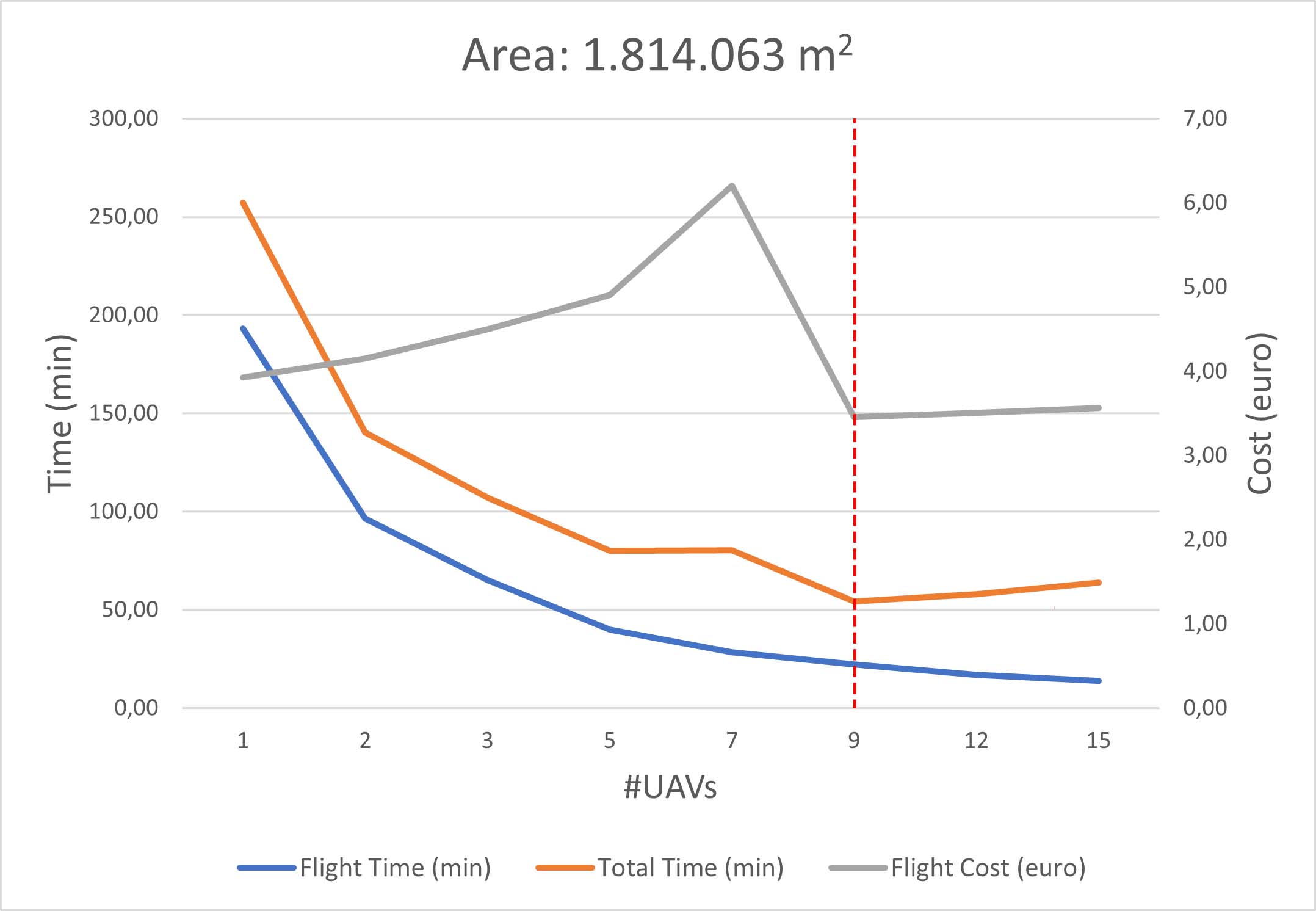}
	\caption{Mission times and flights' cost - Testbed \#4} 
	\label{fig:mCPP_study-4}
\end{figure}

\begin{table*}[!h]
	\def\arraystretch{1.2}
	\centering
	\resizebox{\linewidth}{!}{\begin{tabular}{ c | c  c | c | c  c  c | c | c } 
			\hline
			\textbf{\#UAVs} & \begin{tabular}{@{}c@{}} \textbf{PoC} \\ \textbf{(\%)} \end{tabular} & \begin{tabular}{@{}c@{}} \textbf{PoOC} \\ \textbf{(\%)} \end{tabular} & \textbf{\#Bat/UAV} & \begin{tabular}{@{}c@{}} \textbf{Flight Time} \\ \textbf{(min)} \end{tabular} & \begin{tabular}{@{}c@{}} \textbf{Deployment Time} \\ \textbf{(min)} \end{tabular} & \begin{tabular}{@{}c@{}} \textbf{Change Battery Delay} \\ \textbf{(min)} \end{tabular} & \begin{tabular}{@{}c@{}} \textbf{Total Time} \\ \textbf{(min)} \end{tabular} & \begin{tabular}{@{}c@{}} \textbf{Flight Cost} \\ \textbf{(Euro)} \end{tabular} \\ 
			\hline
			1 & 95.28 & 74.01 & 1 & 23.25 & 8.00 & 0.00 & 31.25 & 0.40 \\
			2 & 95.31 & 77.01 & 1 & 12.70 & 11.00 & 0.00 & 23.70 & 0.44 \\
			\rowcolor{rowGreen}
			3 & 95.22 & 81.43 & 1 & 8.48 & 14.00 & 0.00 & 22.48 & 0.44 \\
			5 & 95.76 & 78.39 & 1 & 5.32 & 20.00 & 0.00 & 25.32 & 0.46 \\
			7 & 95.28 & 84.96 & 1 & 4.28 & 26.00 & 0.00 & 30.28 & 0.52 \\
			9 & 95.30 & 84.68 & 1 & 3.21 & 32.00 & 0.00 & 35.21 & 0.50 \\
			12 & 95.62 & 86.99 & 1 & 2.15 & 41.00 & 0.00 & 43.15 & 0.44 \\
			15 & 95.30 & 90.52 & 1 & 2.15 & 50.00 & 0.00 & 52.15 & 0.56 \\
			\hline
	\end{tabular}}
	\caption{Multi-robot evaluation results - Testbed \#3}
	\label{table:mCPP-SmallROI}
\end{table*}

\begin{table*}[!h]
	\def\arraystretch{1.2}
	\centering
	\resizebox{\linewidth}{!}{\begin{tabular}{ c | c  c | c | c  c  c | c | c } 
			\hline
			\textbf{\#UAVs} & \begin{tabular}{@{}c@{}} \textbf{PoC} \\ \textbf{(\%)} \end{tabular} & \begin{tabular}{@{}c@{}} \textbf{PoOC} \\ \textbf{(\%)} \end{tabular} & \textbf{\#Bat/UAV} & \begin{tabular}{@{}c@{}} \textbf{Flight Time} \\ \textbf{(min)} \end{tabular} & \begin{tabular}{@{}c@{}} \textbf{Deployment Time} \\ \textbf{(min)} \end{tabular} & \begin{tabular}{@{}c@{}} \textbf{Change Battery Delay} \\ \textbf{(min)} \end{tabular} & \begin{tabular}{@{}c@{}} \textbf{Total Time} \\ \textbf{(min)} \end{tabular} & \begin{tabular}{@{}c@{}} \textbf{Flight Cost} \\ \textbf{(Euro)} \end{tabular} \\
			\hline
			1 & 97.66 & 83.95 & \color{red} 8 & 193.13 & 8.00 & 55.92 & 257.05 & 3.93 \\
			2 & 97.68 & 83.57 & \color{red} 4 & 96.50 & 11.00 & 32.97 & 140.47 & 4.15 \\
			3 & 97.53 & 85.17 & \color{red} 3 & 65.07 & 14.00 & 27.98 & 107.05 & 4.50 \\
			5 & 97.67 & 85.64 & \color{red} 2 & 39.98 & 20.00 & 19.99 & 79.97 & 4.91 \\
			7 & 97.50 & 86.44 & \color{red} 2 & 28.46 & 26.00 & 25.99 & 80.45 & 6.20 \\
			\rowcolor{rowGreen}
			9 & 97.86 & 88.31 & 1 & 22.28 & 32.00 & 0.00 & 54.28 & 3.45 \\
			12 & 97.68 & 88.46 & 1 & 16.97 & 41.00 & 0.00 & 57.97 & 3.51 \\
			15 & 97.52 & 86.64 & 1 & 13.78 & 50.00 & 0.00 & 63.78 & 3.56 \\
			\hline
	\end{tabular}}
	\caption{Multi-robot evaluation results - Testbed \#4}
	\label{table:mCPP-LargeROI}
\end{table*}

\begin{figure*}[!h]
	\centering
	\subfigure[Coverage with 1 UAV]{%
		\label{fig:large1}%
		\includegraphics[width = .3\linewidth]{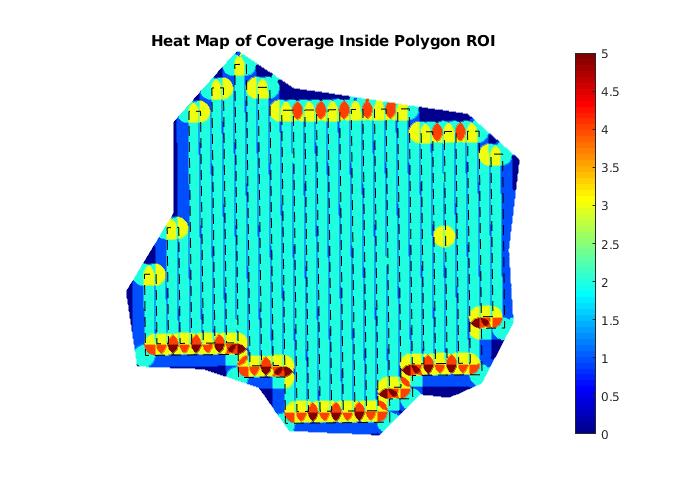}}%
	\qquad
	\subfigure[Coverage with 9 UAVs]{%
		\label{fig:large3}%
		\includegraphics[width = .3\linewidth]{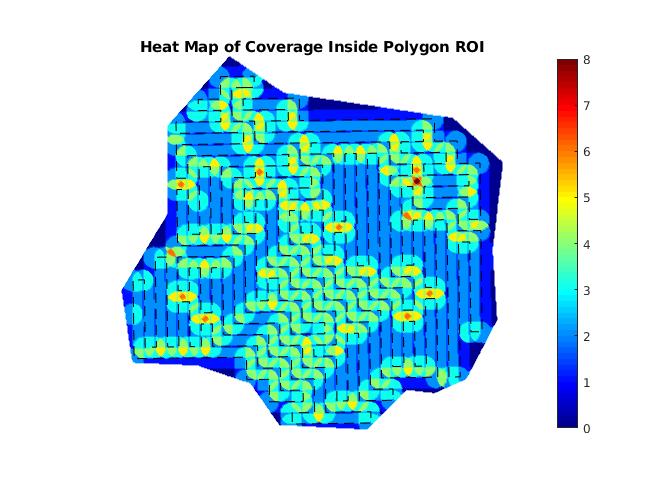}}%
	\qquad
	\subfigure[Coverage with 15 UAVs]{%
		\label{fig:large15}%
		\includegraphics[width = .3\linewidth]{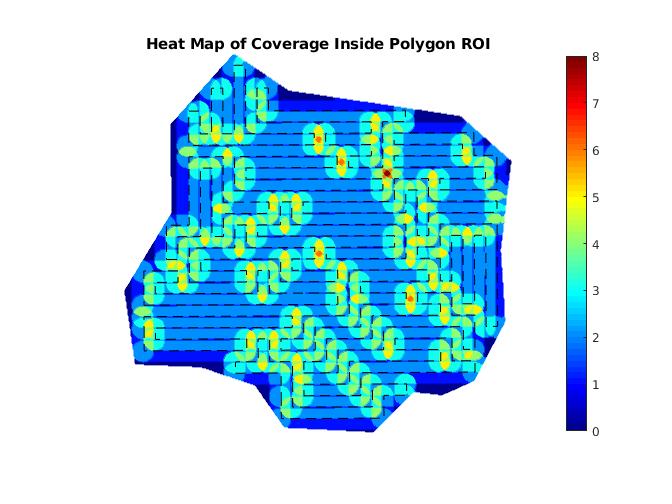}}%
	\caption{Heatmaps of coverage - Testbed \#4}
	\label{fig:mCPP-largeROI}
\end{figure*}

The results of this study for testbed \#4 are presented in table \ref{table:mCPP-LargeROI} and figures \ref{fig:mCPP_study-4} and \ref{fig:mCPP-largeROI}. In this case, the area of the ROI is 1.814.063 $m^2$, which is considered to be of a large size for coverage missions, given the specific mission parameters. Regarding the PoC, as also noticed from the results for testbed \#3, there are no significant discrepancies with respect to the number of UAVs. A slightly increasing tendency in the PoOC was observed, however, smaller than the one observed for testbed \#3. The same increase in the maximum times of revisiting certain points of the ROI was observed here as well, as shown in figure \ref{fig:mCPP-largeROI}. It is worth noting that in order to perform this mission with a single UAV, 8 batteries are demanded (Flight Time$>>$25 min). Overall, the Flight Time decreases constantly, by increasing the number of UAVs. Regarding the Total Time of the mission, it has a global minimum when the mission is performed with 9 UAVs, which is the minimum number of UAVs demanded to perform the mission with a single battery per UAV. The Total Time reduction for the mission performed with 9 UAVs, compared to the one performed with a single UAV, is an impressive amount of 202.77 min. The even more impressive fact, is that the estimated Flight Cost has a steep increasing tendency till the point that the mission is not executable with a single battery per UAV (7 UAVs), and a global minimum value right after it, for the mission performed with 9 UAVs. From that point on, both the Total Time and the Flight Cost slightly increase, however the estimated values suggest that performing the mission with more UAVs, is still far more efficient than performing it with less than 9. Overall, for such a big ROI it is clear that it is not efficient, by any means, to deploy a mission with less vehicles than the number demanded to perform it with a single battery per vehicle. In addition, it is worth noting that with most of the available tools that do not support multi-robot operations, this mission could not be executed at all, as they do not also support to pause and continue the mission from a certain point.

As a general conclusion of this multi-robot study, the deployment of multiple unmanned vehicles in coverage tasks can introduce huge efficiency benefits under specific conditions. When the objective of such missions is to scan very large areas, or in cases where the missions' duration is critical, such as in the search and rescue domain, the multi-robot feature can not only introduce tremendous benefits, regarding both the missions' duration and operational cost, but can also unlock a whole new range of possible applications. However, as in most cases, here as well, there is a marginal utility point where the addition of any further vehicle does not only stops offering benefits, but on the contrary increases cost and decreases the mission's efficiency. A good starting point when considering the number of UAVs that should participate in a mission, is the count of vehicles that makes possible the execution of a mission with a single battery per vehicle, something that is advisable if the application justifies this extra investment cost.

\begin{table}[!t]
	\def\arraystretch{1.2}
	\centering
	\resizebox{\linewidth}{!}{\begin{tabular}{ c | c  c } 
			\hline
			\textbf{\#UAVs} & \begin{tabular}{@{}c@{}} \textbf{Optimization Time} \\ \textbf{(sec)} \end{tabular} & \begin{tabular}{@{}c@{}} \textbf{Overall Execution Time} \\ \textbf{(sec)} \end{tabular}\\ 
			\hline
			1 & 2.87 & 2.88 \\
			2 & 2.69 & 2.69 \\
			3 & 2.74 & 2.75 \\
			5 & 2.73 & 2.73 \\
			7 & 2.78 & 2.78 \\
			9 & 2.77 & 2.77 \\
			12 & 2.95 & 2.96 \\
			15 & 2.75 & 2.75 \\
			\hline
			Average & 2.79 & 2.79 \\
			\hline
	\end{tabular}}
	\caption{Execution times - Testbed \#3}
	\label{table:logTime-SmallROI}
\end{table}

\subsubsection{Algorithm's Execution Time}

For all of the executions of the proposed, optimized mCPP algorithm in the context of this multi-robot study, the needed computation time was logged and included in tables \ref{table:logTime-SmallROI} and \ref{table:logTime-LargeROI}. As an overall comment, for both testbeds the optimization procedure consumes the greatest amount of the execution time. In addition, the time demanded to run the optimization procedure for all cases of each testbed is approximately the same, as this task is dependent only on the generated grid's dimensions, and not on the number of vehicles. In general, the optimization procedure inherits the same complexity with simulated annealing algorithm \cite{blum2021learning}. In addition, more specific information about the run time of the path planning and the area allocation procedures, as well as a thorough computational and memory complexity analysis of DARP algorithm can be found in the original paper \cite{kapoutsis2017darp}.

\begin{table}[!t]
	\def\arraystretch{1.2}
	\centering
	\resizebox{\linewidth}{!}{\begin{tabular}{ c | c  c } 
			\hline
			\textbf{\#UAVs} & \begin{tabular}{@{}c@{}} \textbf{Optimization Time} \\ \textbf{(sec)} \end{tabular} & \begin{tabular}{@{}c@{}} \textbf{Overall Execution Time} \\ \textbf{(sec)} \end{tabular}\\
			\hline
			1 & 16.47 & 16.49 \\
			2 & 16.70 & 16.72 \\
			3 & 16.17 & 16.19 \\
			5 & 16.30 & 16.34 \\
			7 & 16.35 & 16.42 \\
			9 & 16.02 & 16.08 \\
			12 & 15.91 & 16.34 \\
			15 & 16.30 & 16.70 \\
			\hline
			Average & 16.28 & 16.41 \\
			\hline
	\end{tabular}}
	\caption{Execution times - Testbed \#4}
	\label{table:logTime-LargeROI}
\end{table}

\begin{figure*}[!t]%
	\centering
	\includegraphics[width=.9\linewidth]{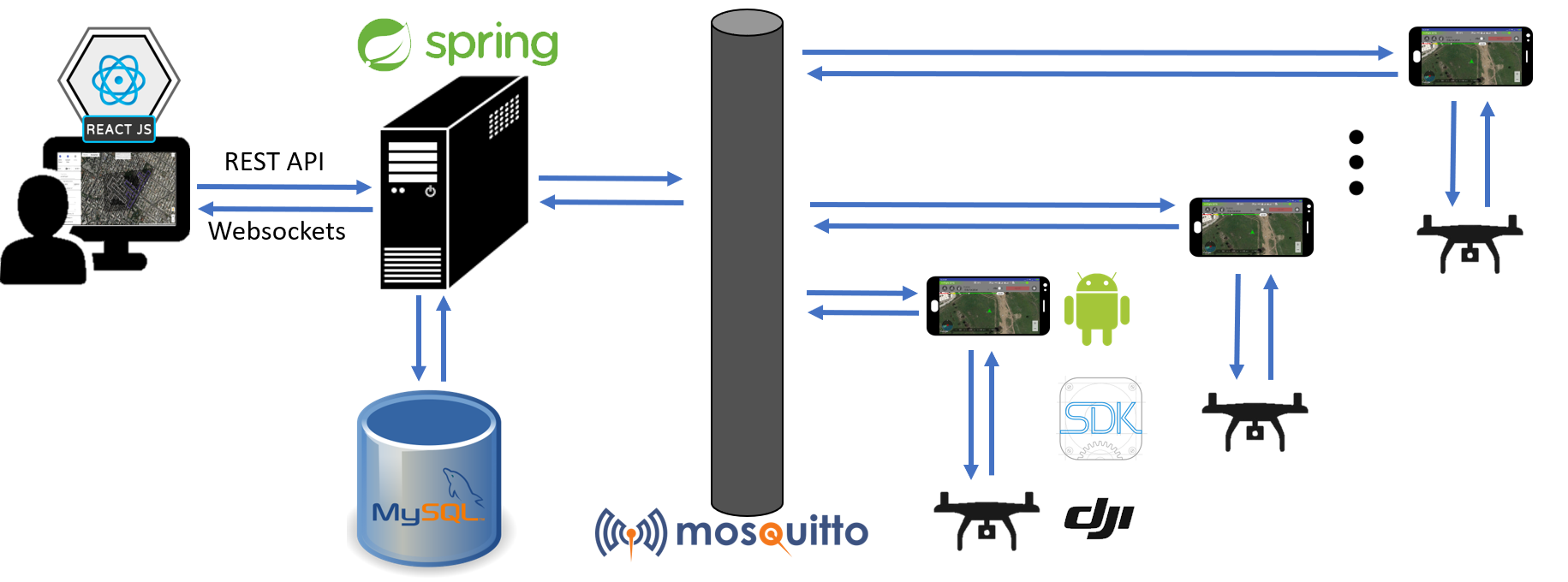}
	\caption{Software platform architecture and technologies used} 
	\label{fig:systemArchitecture}
\end{figure*}

\section{Software Platform Architecture}
\label{sec:platform}

In order to provide an end-to-end solution able to perform coverage missions on-field, along with the path planning methodology presented and evaluated above, has been developed a custom software platform including everything needed to deploy real-life missions. Figure \ref{fig:systemArchitecture} presents the architecture of the overall developed system, along with the technologies used.

Specifically, in the context of this work has been developed from scratch a web-based software application, including (i) a \textit{user-friendly GUI}, (ii) the proposed \textit{mCPP methodology} and (iii) a \textit{database} to store, manage and retrieve missions. In addition, a messaging mechanism and a custom-developed android application for the UAVs' controllers have been developed as well, in order to ensure the real-time communication with the UAVs. 

As this section mostly describes work that required software development and regards specific technologies, protocols and APIs (Application Programming Interfaces), in order to avoid an unclear, highly abbreviated description, it was decided to present the work at a glance and focus mainly on parts that could be useful for the platform's end-user.

\begin{figure*}[!t]%
	\centering
	\includegraphics[width=.9\linewidth]{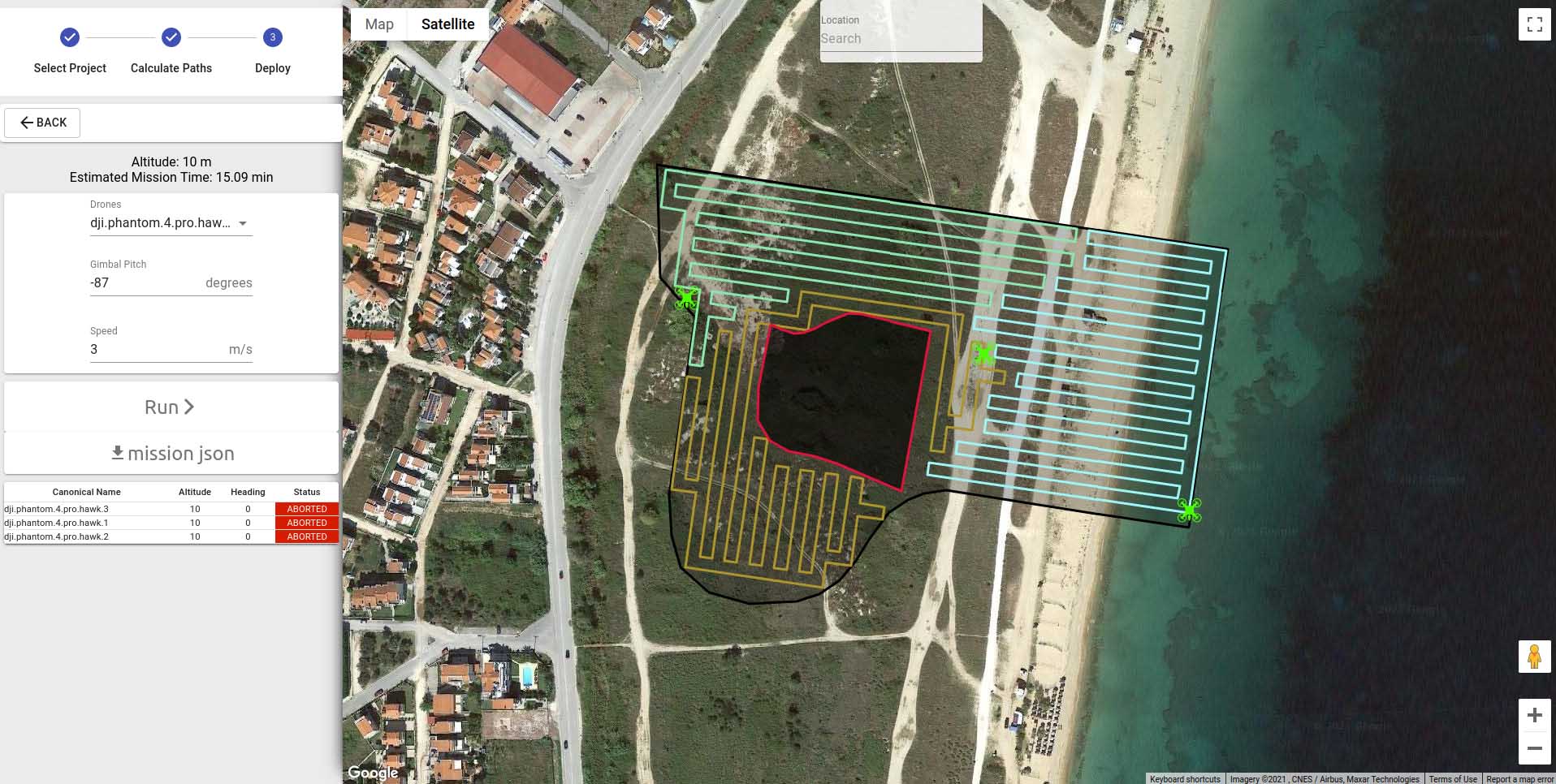}
	\caption{Mission execution as presented from the Graphical User Inteface}
	\label{fig:GUI}
\end{figure*}

\subsection{Backend/Frontend Interraction}
\label{subsec:UIVis1}
For the platform development has been employed a client-server architecture, a common paradigm for the web, meaning that the server delivers and manages most of the multi-UAV mission data and related computations that are going to be consumed by the client. In the client side (web browser), a modern and intuitive user interface powered by Material-UI framework and ReactJS (JavaScript library) has been implemented, that is supported by most browsers. In order to ensure real-time communication for the UAVs' positions and statuses between the client and the server, WebSockets protocol has been used, which is a popular technology used when servers need to push a lot of data, or frequently update the browser. In addition, the server exposes a RESTful (REpresentational State Transfer) API for computational requests and CRUD operations (Create, Read/retrieve, Update and Delete) for the database. The server side (backend) has been developed utilizing the Spring Framework for the Java Platform. Finally, the interaction between backend and the MySQL (Structured Query Language) database occurs with the Hibernate framework, an object-related mapping (ORM) library for the Java language.


\subsection{User Interface - Interaction and Visualization}
\label{subsec:UIVis}

The GUI developed for the ground control station (GCS), is the main interaction layer of the platform with the user. Through it, the user is able to define, process, store, manage, execute and supervise missions. In the right side of the GUI is placed a map, in order to locate and define the ROI, check the generated paths and supervise the progress of missions during the execution. Additionally, in the left side of GUI there is a toolbar, organized in three separate tabs. From the first tab of the toolbar, the user can access the database to retrieve stored missions or select to create new ones. As a next step, in the second tab the user can define the parameters for a new mission, or edit the ones of an already stored. At this step the user is able to define a ROI (and optionally some obstacles inside it) and the rest of the mission's parameters (number of UAVs, altitude, percentage of overlap, percentage of the ROI that should be assigned to each UAV and strictly-in-poly/better-coverage mode) in order to calculate coverage paths. Given the selected altitude and overlap, the user can also inspect the mission's scanning density $d_s$ and the ground sampling distance (GSD). GSD is calculated using equation \ref{eq:dHOV} and the resolution of the sensor that will be used\footnote{GSD is calculated from the take off position, considering that the whole ROI has an almost flat topology. In order to have a constant GSD  for ROIs with altitude variations, the flight altitude should be adjusted based on the actual distance from the ground, at each point of the region (see also\cite{gomez2020mission}).}, a value that is retrieved from the database:


\begin{figure}[!b]%
	\centering
	\includegraphics[width=\linewidth]{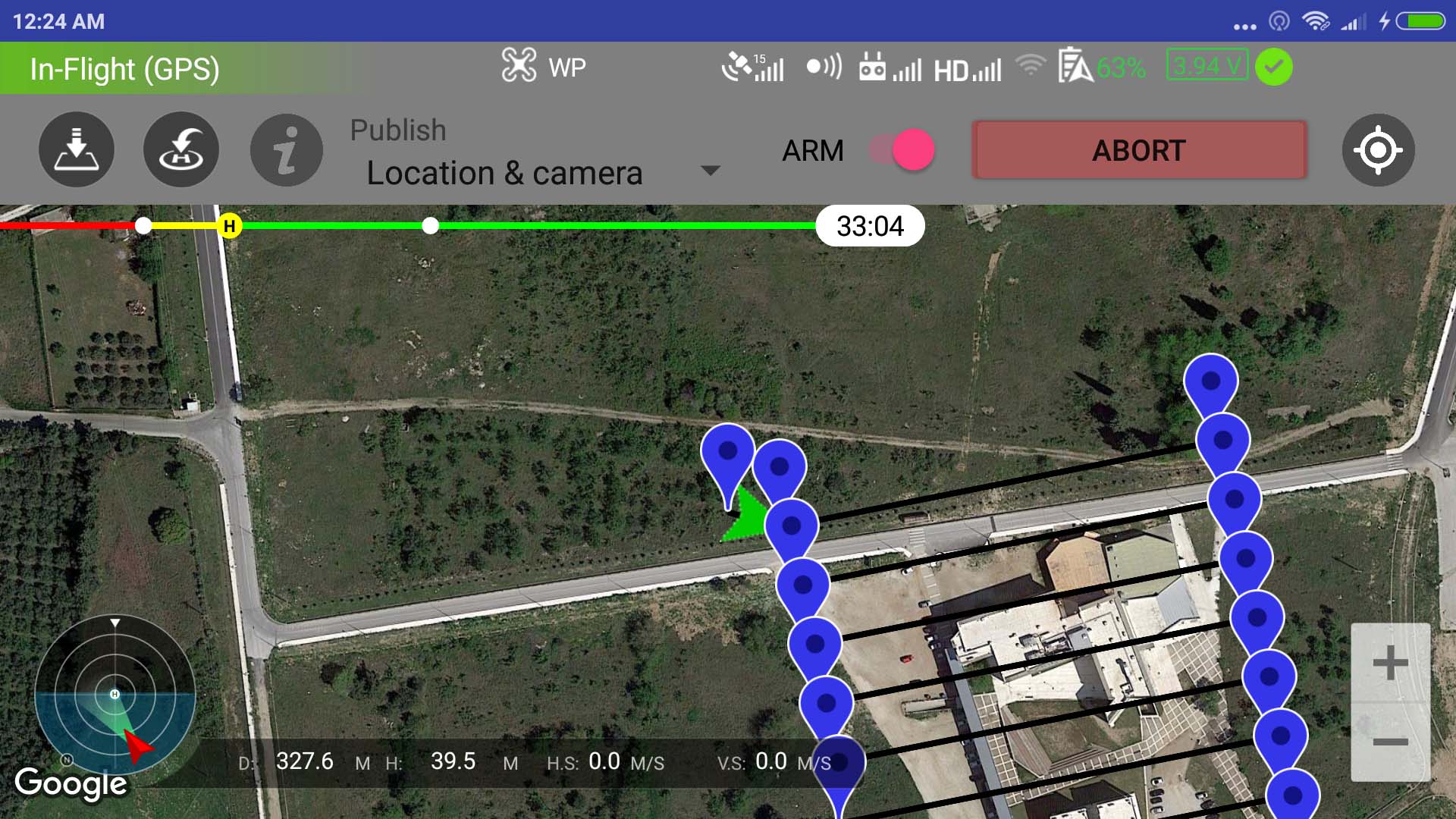}
	\caption{Custom-developed android application}
	\label{fig:androidApp}
\end{figure}

\begin{equation}
\begin{split}
GSD = \frac{d}{hRes} = \frac{2 \cdot h \cdot tan(HFOV/2)}{hRes},
\end{split}
\label{eq:GSD}
\end{equation}
where hRes is the resolution of the horizontal dimension of the sensor.
The final step to deploy a mission is included in the third tab, where the user can define the operational parameters (camera gimbal pitch and speed), select the UAVs that will participate (stored in the database) and supervise the progress of the mission.  Figure \ref{fig:GUI} presents the GUI visualization during the real-time operation for a mission with 3 UAVs.

\subsection{Communication with UAVs - Android Application}
\label{subsec:CLwithUAVs}
In the controller of each UAV is connected a smartphone, running a mobile application that works as a user interface and interaction layer of the UAV with their operator. In the context of this work, a custom android application has been developed with the use of the DJI API  (figure \ref{fig:androidApp}), that takes the role of an intermediate communication layer of the overall platform with each UAV. This application is responsible to gather information (telemetry, photographs, statuses and mission logs) and transmit them to the backend of the platform. In addition, it visualizes mission's information and operational parameters, that could be useful for the operator.

The overall system has been setup with the objective to provide a high-level of autonomy for the missions. Based on this logic, the interaction of the platform's user with the whole system can be done entirely from the GCS, with no need to directly control the UAVs with the controllers or the smartphones' applications. However, at any time, based on the user's judgment, it is possible to take control and directly operate any of the UAVs, if needed.

\section{Real Life Experiments}
\label{sec:experiments}

In this section are presented some indicative examples of real-life applications, where the proposed platform can be used to obtain data, along with results that are usually desired in such tasks. This way, the functionalities and efficiency of the platform are evaluated through the quality of the acquired data and the produced results, i.e. through the reason why such platforms are developed and used. To acquire the desired photogrammetry results, the gathered data were processed with the use of the web platform of OpenDroneMap, WebODM\footnote{\url{https://www.opendronemap.org/webodm/}}. WebODM is a powerful, open source solution for processing UAV imagery and acquiring results in a wide range of applications. The objective of this section is to present the applicability and the robust operation of the proposed platform, in real-world coverage missions.

\subsection{Experimental Setup}
\label{subsec:setup}
\begin{figure}[!h]%
	\centering
	\includegraphics[width=\linewidth]{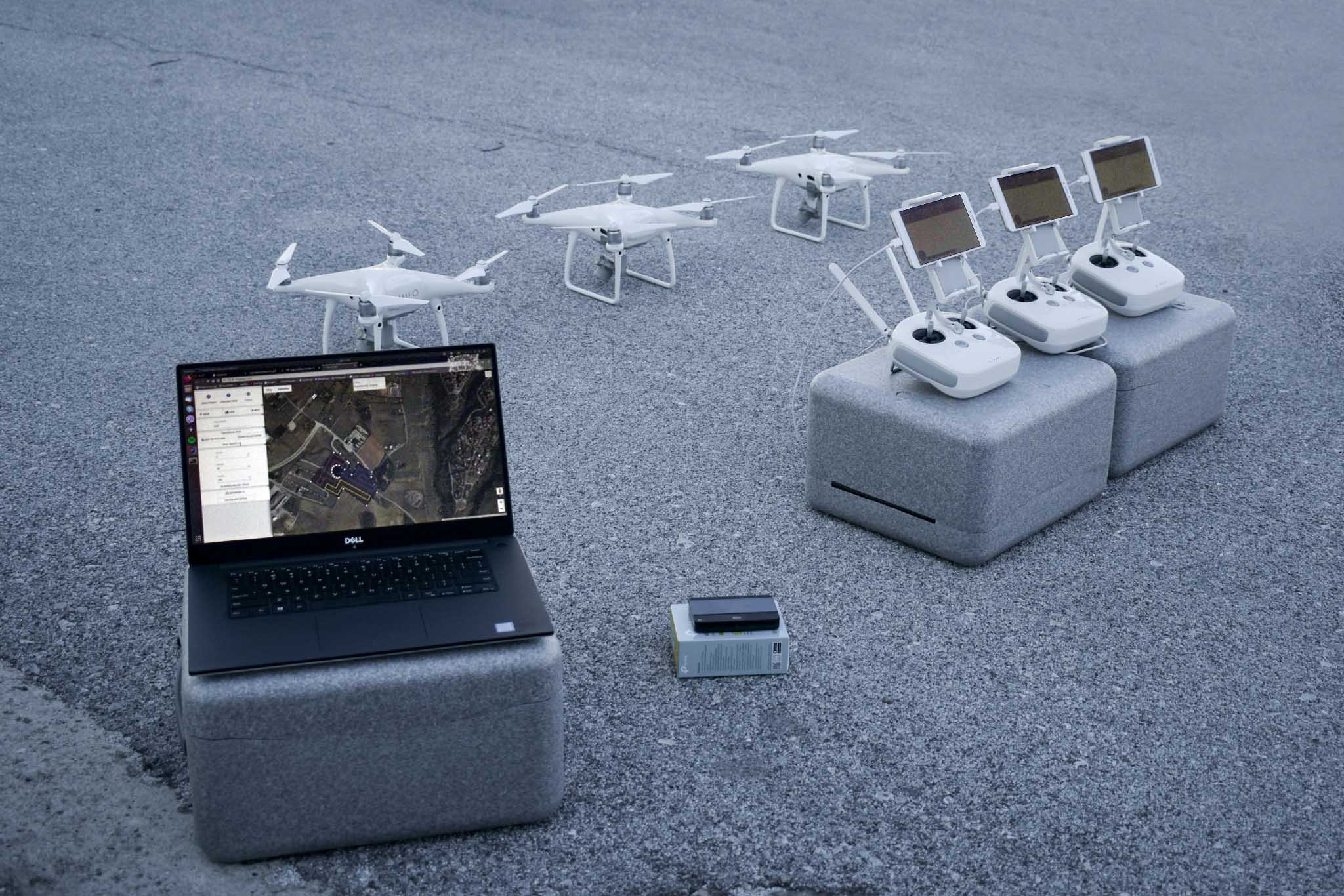}
	\caption{Experimental hardware setup}
	\label{fig:expSetup}
\end{figure}

For the real-life experiments were used:
\begin{itemize}
	\item a common commercial laptop\footnote{Dell XPS 9570}, as a portable GCS,
	\item a swarm of common commercial UAVs\footnote{\url{https://www.dji.com/phantom-4-pro}},
	\item a set of android smartphone devices\footnote{Xiaomi Mi Max 2}, to run the custom-developed application, described in section \ref{sec:platform} and
	\item a portable 4G router\footnote{tp-link M7350}, as an intermediate communication layer between the smartphones and the portable GCS.
\end{itemize}
The hardware setup used, is shown in figure \ref{fig:expSetup}.

\subsection{Precision Agriculture Application}
\label{subsec:precAgrExp}

As reported in \cite{hunt2018good} remote sensing from unmanned aircraft systems is a very important new technology that introduced many benefits for the farmers in the field of precision agriculture, especially crop nutrient management. According to \cite{karatzinis2020towards}, some of the advances that the use of UAVs introduced in agriculture are (i) the affordable crop mapping, (ii) the early identification of problems through vegetation indices, that prevents yield losses and (iii) the better planning of crop management operations.

In order to prove the efficiency of the developed platform in the collection of data for such tasks, an experiment was performed in a field with olive trees. The main objectives of this experiment were to (i) get an updated image of the field on map, in order to facilitate the planning of various works, (ii) acquire some kind of plant health information, that could give indications on the year's harvest and possible actions that should be taken to improve it and (iii) acquire some kind of bio-mass estimation that is in general useful to monitor plant and weeds growth, as well as organize a wide range of agricultural tasks.

Depending on the type of operations and the accuracy of data required, UAVs used in precision agriculture are usually equipped with RGB, multi-spectral or hyper-spectral cameras. In this experiment participated two UAVs, equipped with RGB cameras. The mission altitude was selected to be 18 meters and the percentage of overlap $(p_o)$ 90\%, as advised from the WebODM documentation for a decent 3D reconstruction. The selected altitude and $p_o$ resulted in a $d_s$ of about 14 meters and the GSD in the collected images gets a value of 0.49 cm/px. In addition, the operational speed was selected to be 2 m/s, which is low enough to not spoil the quality of the collected images, and the gimbal pitch was selected to be 45 degrees, an angle that usually contributes in the quality of an area's 3D reconstruction. Overall, for this ROI were collected 525 images, that were later used for the reconstruction of the results presented below.


\begin{figure*}[!h]
	\centering
	\subfigure[Orthomosaic overlaid by the ROI and generated paths]{%
		\label{fig:presAgrPathOrtho}%
		\includegraphics[height=2.28in]{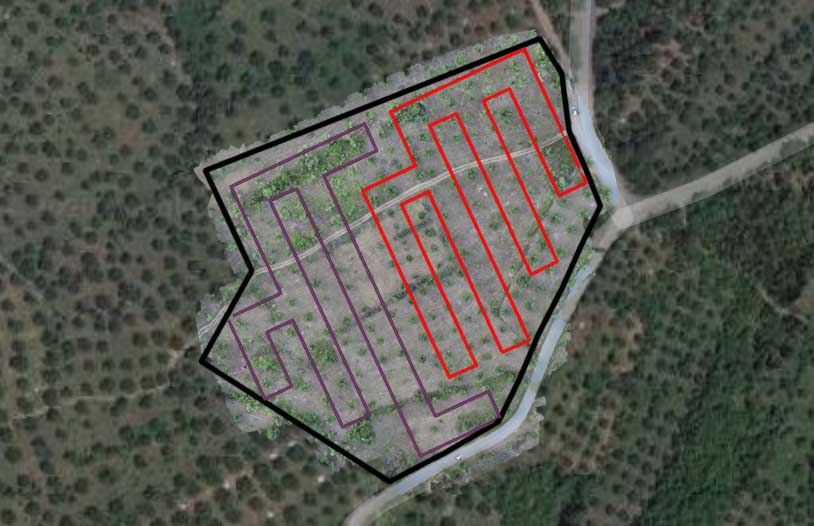}}%
	\qquad
	\subfigure[Plant health]{%
		\label{fig:olivesHealth}%
		\includegraphics[height=2.28in]{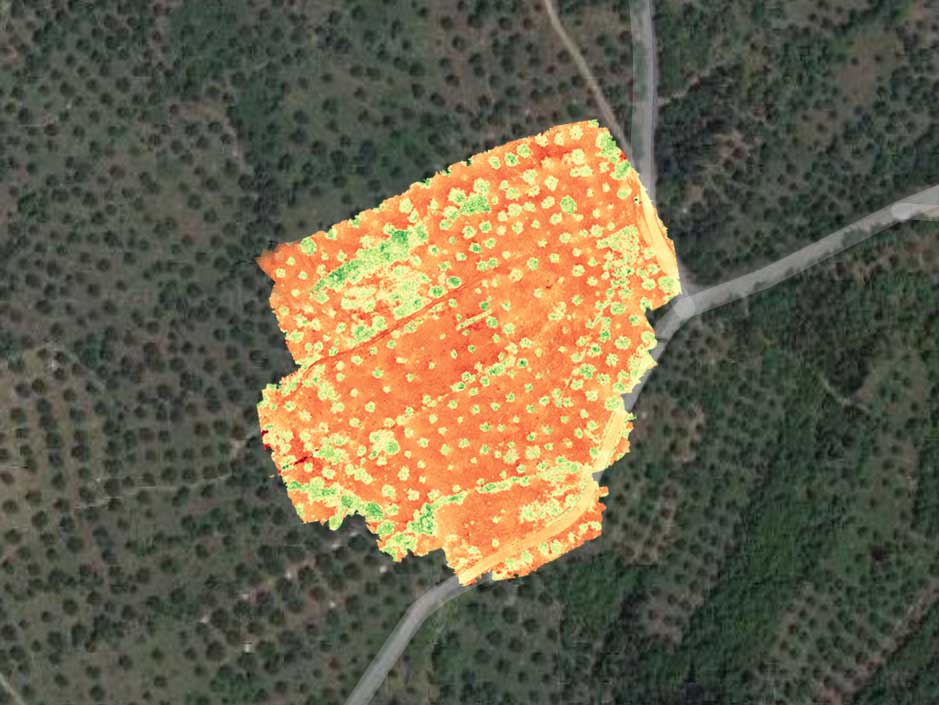}}%
	\qquad
	\subfigure[Point cloud]{%
		\label{fig:olivesPointcloud}%
		\includegraphics[height=2.83in]{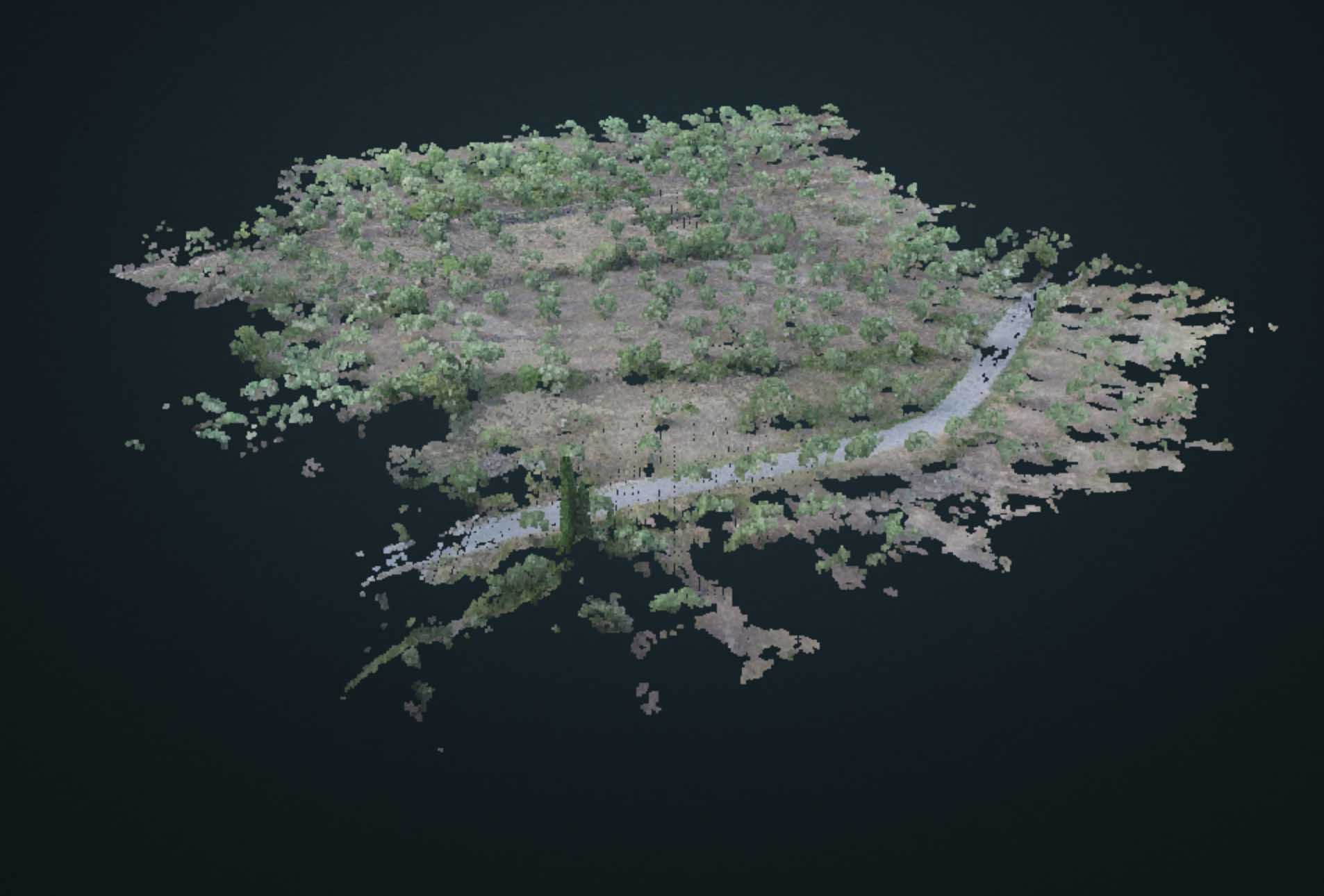}}%
	\qquad
	\subfigure[Elevation model]{%
		\label{fig:olivesElevation}%
		\includegraphics[height=2.83in]{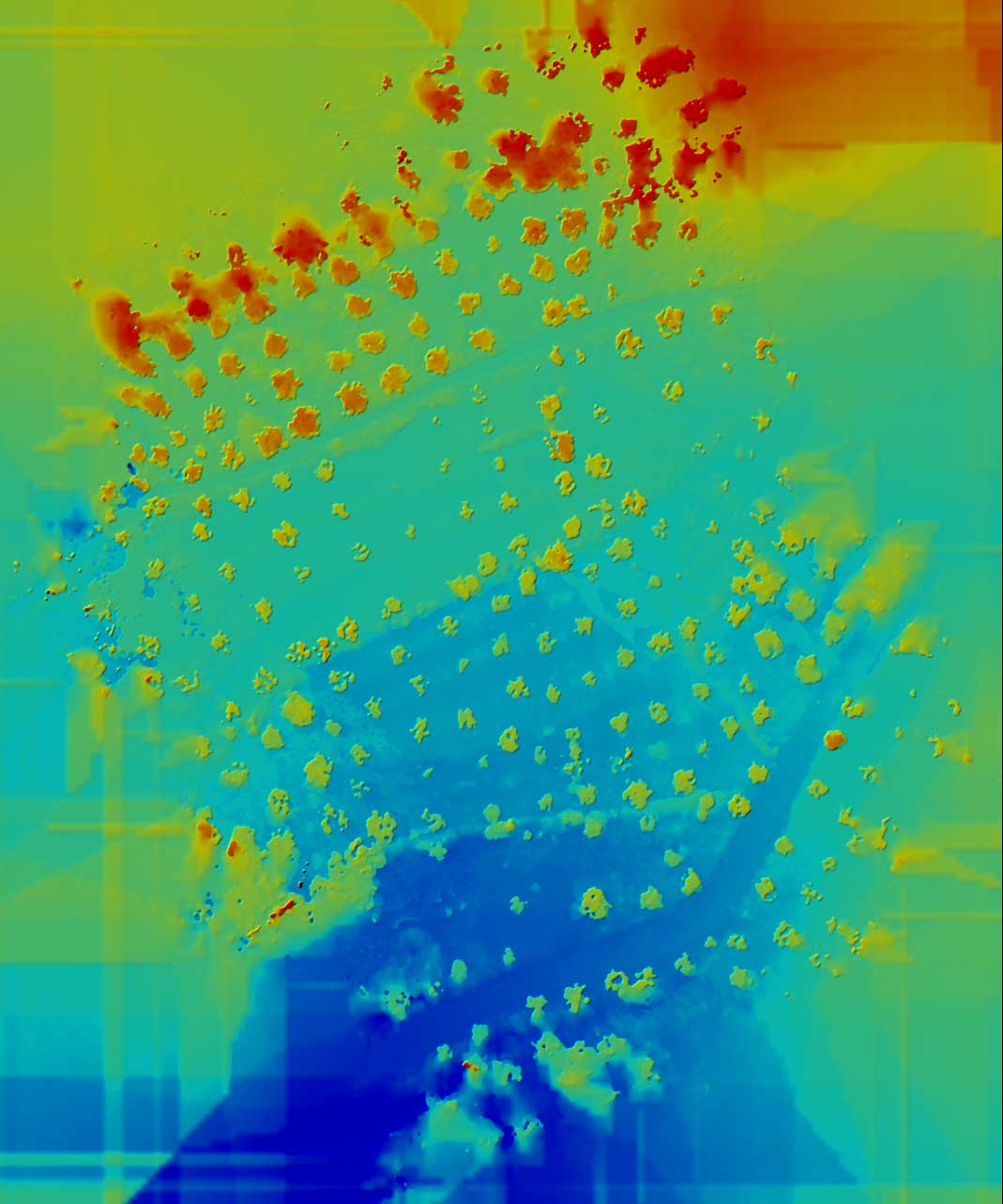}}%
	\caption{Photogrammetry results for precision agriculture}
	\label{fig:olives-results}
\end{figure*}

Figure \ref{fig:olives-results} illustrates the key outcomes of this experiment, which are usually desired in this type of precision agriculture applications. It is important to note here that all results depicted in figure \ref{fig:olives-results} were created from the data gathered in a single run, with the mission parameters described above.

Figure \ref{fig:presAgrPathOrtho} shows the area that was actually covered, along with the polygon ROI and the calculated paths. It is clear from the result that almost the whole user-defined ROI was successfully covered and that the data gathered are sufficient for a qualitative representation of the desired area. This 2D reconstruction of the ROI is called orthomosaic and it was was ceated to satisfy the objective (i) of this experiment. Orthomosaics are high-resolution maps created by aerial images, geometrically corrected (``orthorectified'') such that the scale is uniform. Orthomosaics can be used for an overall monitoring of the ROI, as the information on the existing map gets updated, and for making useful measurement (e.g., planting lines and distances, percentage of ground covered by vegetation etc.). In \ref{fig:olivesHealth} the created orthomosaic is used to produce a plant health map, to satisfy the objective (ii) of the experiment. Such maps, are specifically targeted towards agriculture. Their main purpose is to allow an exploration of the agricultural data even more deeply. Once the relevant plant health ranges have been identified, the color map provides a thresholding tool, able to quantify damage and predict yields by showing the areas within specific ranges. Finally, \ref{fig:olivesPointcloud} and  \ref{fig:olivesElevation} show a pointcloud and an elevation map of the ROI, respectively, that were created to satisfy the objective (iii) of the experiment. Pointclouds are created by the combination of visual and depth information in order to produce a 3D representation of the scenery. On the other hand, elevation maps use only depth information, in order to create a color map that visualizes the altitude differences in a ROI. Both of them are useful for the biomass estimation and a more efficient planning of tasks in the agricultural fields.

\begin{figure*}[!b]
	\centering
	\subfigure[Area covered and location of detected objects]{%
		\label{fig:SnRPathOrtho}%
		\includegraphics[height=6.28cm]{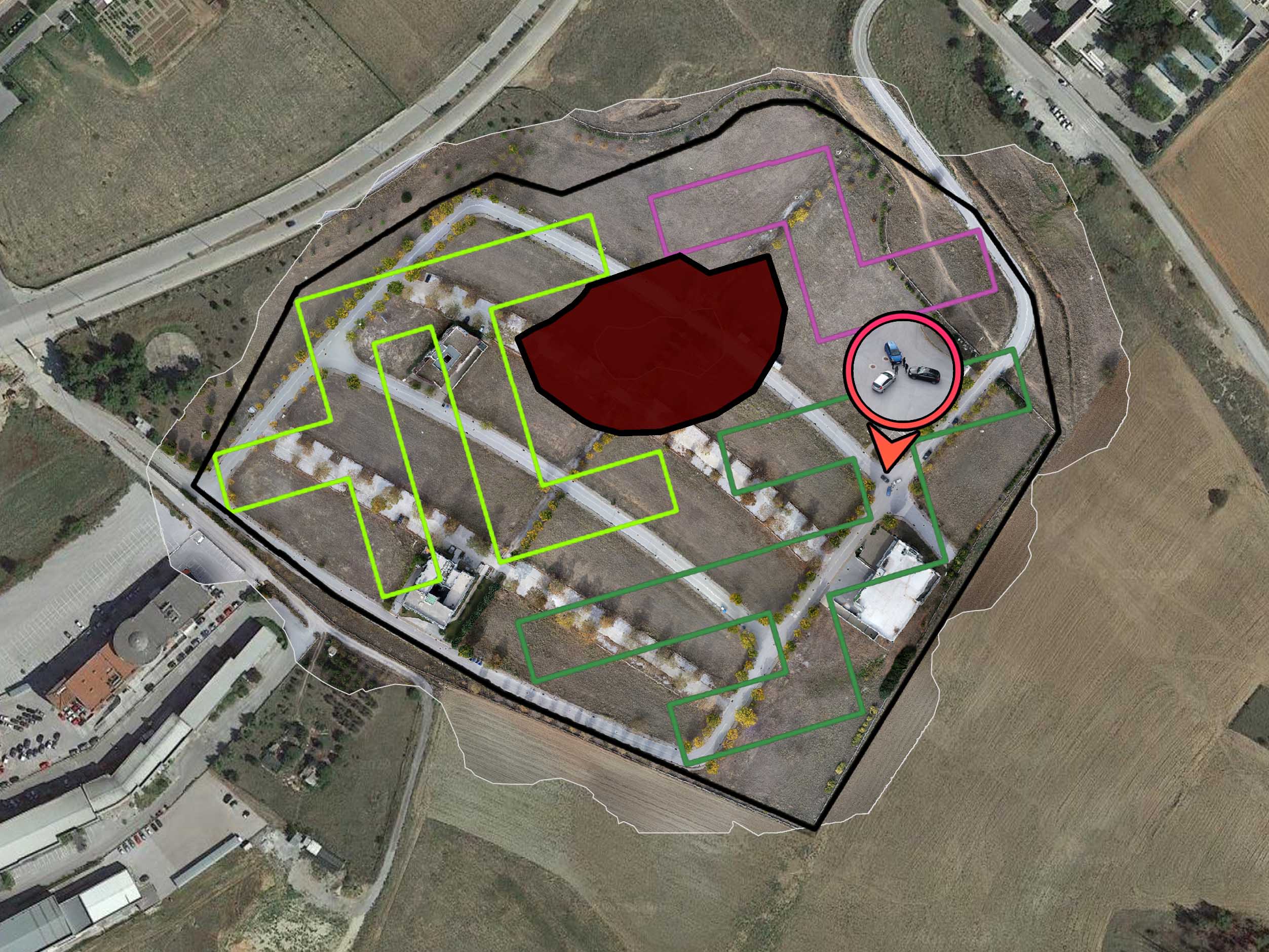}}%
	\qquad
	\subfigure[Cars of interest detected in the mission's ROI]{%
		\label{fig:carsYOLO}%
		\includegraphics[height=6.28cm]{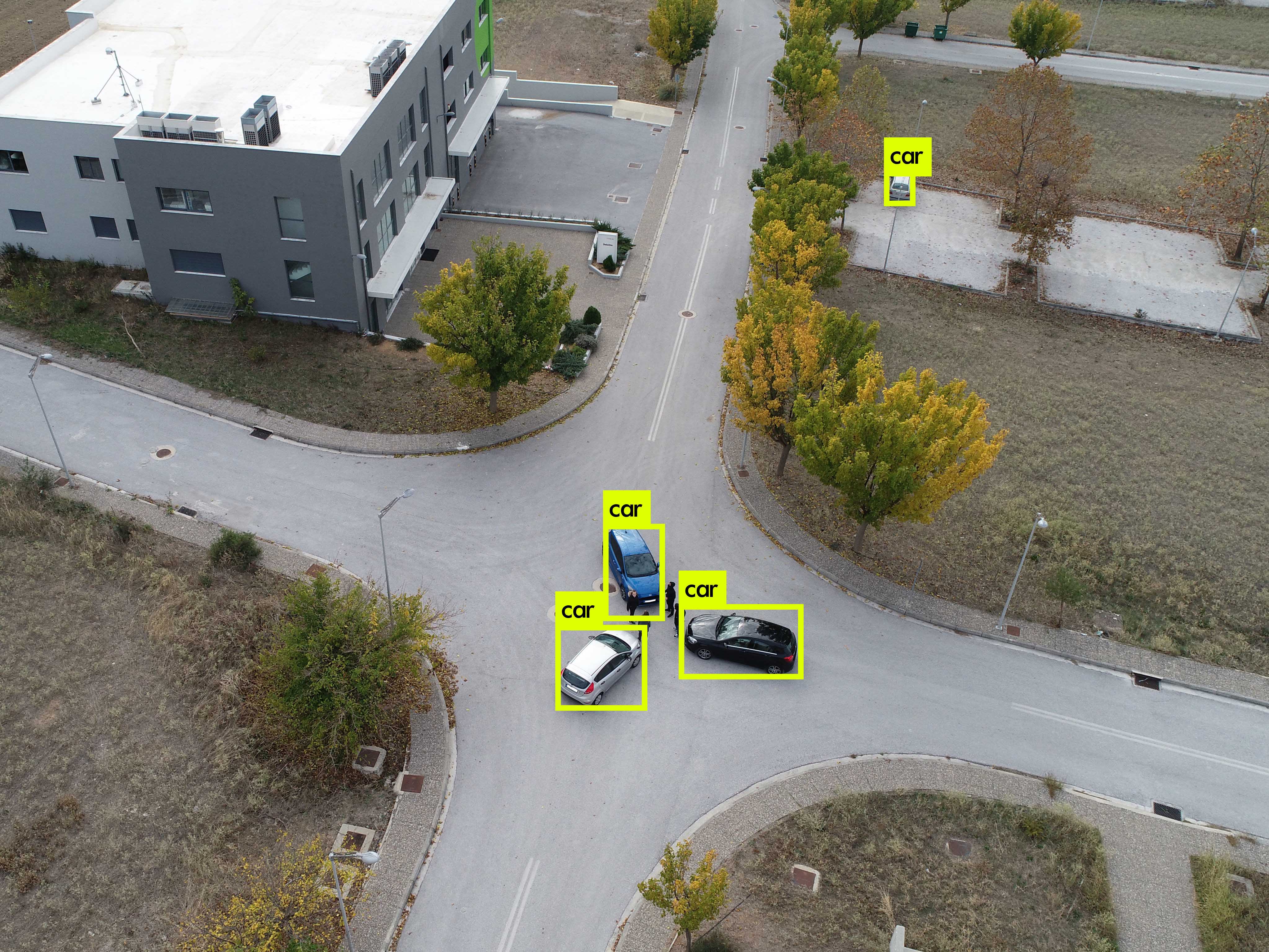}}%
	\caption{Search and rescue coverage mission}
	\label{fig:snrMission}
\end{figure*}

Overall, the results created for the scanned ROI are satisfactory and correspond to the actual condition and topology of the field. The user-defined ROI is covered almost completely and the gathered data are sufficient to produce a qualitative representation of the field, proving that the developed platform is appropriate for such operations. Finally, the ability to use multiple UAVs, allows (i) the faster conduction of tasks and (ii) the coverage of large areas with a single mission, something that is not always possible with a single UAV. For the mission presented here, the execution time was 7.56 minutes, while the estimated time to execute the same mission, with a single UAV, is about 14.6 minutes.

\subsection{Search and Rescue Application}
\label{subsec:searchExp}

According to \cite{bejiga2017convolutional}, the introduction of UAVs in search and rescue operations, has managed to face one of the very significant problems in this kind of missions. Specifically, the resources and time that were required to deploy rescue teams before, were major bottleneck that decreased the victim's chance of survival. Thanks to the advances in the field of UAVs, now the assessment of the situation, or the damage caused by natural or man-made disasters, and the location of victims in the debris, can be done with these small, unmanned vehicles, reducing both the deployment time and the operational risk to the the minimum. As reported in \cite{rudol2008human}, UAVs also offer increased precision and the capability of long duration missions, that are both critical in some cases. In addition, both equipment and operational costs are significantly reduced, making aerial search and rescue missions more affordable and available to a wider range incidents.

For the evaluation of the proposed platform in this kind of operations, was performed a mission with the following imaginary scenario: In a complex of business buildings, on a public holiday, there was an information about a suspicious meeting for smuggling, where a blue or a silver vehicle could be involved. A team of first responders deployed a swarm of UAVs in the region with the objective to completely cover the ROI and detect any possible vehicle in the area, that fits the available description. The images delivered in the GCS, in real time, were fed to an object detection algorithm (YOLO v3 \cite{redmon2018yolov3}) in order to detect the vehicles of interest, during the execution of the mission, and locate them on map. Finally, an orthomosaic was created with the collected photographs to update the image on map and provide better overall situational awareness.

For this mission, in an attempt to minimize the time from deployment, to the detection of vehicle(s) of interest, all of the available UAVs were utilized. This way, the mission was performed with 3 UAVs, however, due to the lower battery level on one of them, the tasks were allocated proportionally, with this UAV undertaking only the 15\% of the overall region, and the other two undertaking 40\% and 45\% respectively. In addition, due to the direct visual contact, the area around the point of deployment was excluded from the ROI. The mission altitude was selected to be 50 meters, in order to fly over the buildings in the field and the $p_o$ 160\%, so as to cover all points multiple times and reduce the risk of losing the detection of possible targets, because of their movement. The selected altitude and $p_o$ resulted in a $d_s$ of about 28.5 meters and the GSD in the collected images gets a value of 1.36cm/px. As in the previous experiment, the operational speed was selected to be 2 m/s and the gimbal pitch 45 degrees, to ease the detection of the vehicles from the object detection algorithm \cite{redmon2018yolov3}. From this mission were collected 432 images in total, that were used for the acquirement results presented bellow.

Figure \ref{fig:SnRPathOrtho} shows the defined ROI and the mission generated with these specifications, along with the area that was actually covered and the location of the detected objects of interest. As in the previous experiment, the whole ROI was successfully covered and the gathered data are sufficient for a qualitative representation of the area. Figure \ref{fig:carsYOLO} shows the output of the detection algorithm for the vehicles of interest (zoomed image).

Regarding the execution time, this mission needed 8.7 minutes to be completed, while for a single-UAV mission with the same specifications the estimated time for completion is more than 21. The utilization of multiple UAVs, also increased the probability to detect the objects of interest earlier. Indeed, the time needed from the beginning of the mission to the detection of the vehicles was just about 3.5 minutes.

Overall, the features and functionalities of the proposed mCPP approach and the end-to-end platform, seemed to be extremely useful in real-life operational conditions. The results achieved in both cases, presented in this section, were very satisfactory for this type of operations. In addition, the multi-robot capability and the energy efficiency features provided, are strong incentives to prefer the proposed solution over other mission planners, commercial or not, that are currently available.

\section{Conclusions and Future Work}
\label{sec:conclusions}
This paper aims at bridging the gap between the already existing, powerful UAVs platforms and the effective utilization of multiple UAVs, in the coverage path planning domain. Towards that purpose, first, the original grid-based STC algorithm \cite{gabriely2001spanning} was revisited to enable the grid's adjustment on the specific geometrical characteristics of the region of interest. To achieve that, an optimization approach, based on Simulated Annealing algorithm \cite{Kirkpatrick1983OptimizationBS}, was put in charge of the discretization strategy that ultimately controls the coverage performance. The efficiency of the optimization scheme is validated through an extensive series of simulations, in comparison with a state-of-the-art CPP alternative methodology \cite{bahnemann2019revisiting}. Then, DARP algorithm \cite{kapoutsis2017darp} takes over to divide the previously optimized grid into as-many-as-the-number-of-UAVs exclusive areas, with respect to their operational capabilities. 

The realization of the aforedescribed methodology is achieved through a production-ready software platform, that respects all the modern-software best practices. Two field experiments with different objectives were performed to prove the robustness of the platform: (i) a precision agriculture application, and (ii) a search and rescue mission. In both real-life demonstrations, the proposed platform preserved all the desired features of the commercial UAVs platforms, while at the same time, it achieved tremendous mission-duration improvements, that were directly proportional to the number of UAVs deployed in each mission. Such a feature is of paramount importance, as it enables the seamless utilization of more UAVs to trade-off their battery limitations, which is the one of the major factors that, at the moment, constrains the deployment of UAVs in specific types of missions.

Future research endeavors will focus on the integration of UAVs that have available on-board processing units. In such an approach, the proposed navigation platform should be enriched to be able to handle cases, where information about the environment (e.g., obstacles, other UAVs, etc.) comes also during the operation. This work may consider more complicated scenarios where UAVs cooperate with other autonomous devices or even humans in the same environment.

\section*{Acknowledgment}	
This project has received funding from the European Commission under the European Union’s Horizon 2020 research and innovation programme under grant agreement no 833805 (ARESIBO).


%
%

\bibliographystyle{unsrt}
\bibliography{references}   

\begin{thebibliography}{10}

\bibitem{deng2018uav}
Lei Deng, Zhihui Mao, Xiaojuan Li, Zhuowei Hu, Fuzhou Duan, and Yanan Yan.
\newblock Uav-based multispectral remote sensing for precision agriculture: A
  comparison between different cameras.
\newblock {\em ISPRS Journal of Photogrammetry and Remote Sensing},
  146:124--136, 2018.

\bibitem{maes2019perspectives}
Wouter~H Maes and Kathy Steppe.
\newblock Perspectives for remote sensing with unmanned aerial vehicles in
  precision agriculture.
\newblock {\em Trends in plant science}, 24(2):152--164, 2019.

\bibitem{comba2018unsupervised}
Lorenzo Comba, Alessandro Biglia, Davide~Ricauda Aimonino, and Paolo Gay.
\newblock Unsupervised detection of vineyards by 3d point-cloud uav
  photogrammetry for precision agriculture.
\newblock {\em Computers and Electronics in Agriculture}, 155:84--95, 2018.

\bibitem{ham2016visual}
Youngjib Ham, Kevin~K Han, Jacob~J Lin, and Mani Golparvar-Fard.
\newblock Visual monitoring of civil infrastructure systems via camera-equipped
  unmanned aerial vehicles (uavs): a review of related works.
\newblock {\em Visualization in Engineering}, 4(1):1, 2016.

\bibitem{al2017vbii}
Abdulla Al-Kaff, Francisco~Miguel Moreno, Luis~Javier San~Jos{\'e}, Fernando
  Garc{\'\i}a, David Mart{\'\i}n, Arturo de~la Escalera, Alberto Nieva, and
  Jos{\'e} Luis~Meana Garc{\'e}a.
\newblock Vbii-uav: Vision-based infrastructure inspection-uav.
\newblock In {\em World Conference on Information Systems and Technologies},
  pages 221--231. Springer, 2017.

\bibitem{renzaglia2020common}
Alessandro Renzaglia, Jilles Dibangoye, Vincent Le~Doze, and Olivier Simonin.
\newblock A common optimization framework for multi-robot exploration and
  coverage in 3d environments.
\newblock {\em Journal of Intelligent \& Robotic Systems}, pages 1--16, 2020.

\bibitem{sun2016camera}
Jingxuan Sun, Boyang Li, Yifan Jiang, and Chih-yung Wen.
\newblock A camera-based target detection and positioning uav system for search
  and rescue (sar) purposes.
\newblock {\em Sensors}, 16(11):1778, 2016.

\bibitem{qi2016search}
Juntong Qi, Dalei Song, Hong Shang, Nianfa Wang, Chunsheng Hua, Chong Wu, Xin
  Qi, and Jianda Han.
\newblock Search and rescue rotary-wing uav and its application to the lushan
  ms 7.0 earthquake.
\newblock {\em Journal of Field Robotics}, 33(3):290--321, 2016.

\bibitem{koutras2020autonomous}
Dimitrios Koutras, Athanasios Kapoutsis, and Elias Kosmatopoulos.
\newblock Autonomous and cooperative design of the monitor positions for a team
  of uavs to maximize the quantity and quality of detected objects.
\newblock {\em IEEE Robotics and Automation Letters}, 5(3):4986--4993, 2020.

\bibitem{kapoutsis2019distributed}
Athanasios~Ch Kapoutsis, Savvas~A Chatzichristofis, and Elias~B Kosmatopoulos.
\newblock A distributed, plug-n-play algorithm for multi-robot applications
  with a priori non-computable objective functions.
\newblock {\em The International Journal of Robotics Research}, 38(7):813--832,
  2019.

\bibitem{Galceran_2013}
Enric Galceran and Marc Carreras.
\newblock A survey on coverage path planning for robotics.
\newblock {\em Robotics and Autonomous Systems}, 61(12):1258--1276, dec 2013.

\bibitem{rekleitis2008efficient}
Ioannis Rekleitis, Ai~Peng New, Edward~Samuel Rankin, and Howie Choset.
\newblock Efficient boustrophedon multi-robot coverage: an algorithmic
  approach.
\newblock {\em Annals of Mathematics and Artificial Intelligence},
  52(2):109--142, 2008.

\bibitem{Cabreira_2019}
Tau{\~{a}} Cabreira, Lisane Brisolara, and Paulo R.~Ferreira Jr.
\newblock Survey on coverage path planning with unmanned aerial vehicles.
\newblock {\em Drones}, 3(1):4, jan 2019.

\bibitem{gabriely2001spanning}
Yoav Gabriely and Elon Rimon.
\newblock Spanning-tree based coverage of continuous areas by a mobile robot.
\newblock {\em Annals of mathematics and artificial intelligence},
  31(1-4):77--98, 2001.

\bibitem{Gower_1969}
J.~C. Gower and G.~J.~S. Ross.
\newblock Minimum spanning trees and single linkage cluster analysis.
\newblock {\em Applied Statistics}, 18(1):54, 1969.

\bibitem{choset1998coverage}
Howie Choset and Philippe Pignon.
\newblock Coverage path planning: The boustrophedon cellular decomposition.
\newblock In {\em Field and service robotics}, pages 203--209. Springer, 1998.

\bibitem{li2011coverage}
Yan Li, Hai Chen, Meng~Joo Er, and Xinmin Wang.
\newblock Coverage path planning for uavs based on enhanced exact cellular
  decomposition method.
\newblock {\em Mechatronics}, 21(5):876--885, 2011.

\bibitem{coombes2017boustrophedon}
Matthew Coombes, Wen-Hua Chen, and Cunjia Liu.
\newblock Boustrophedon coverage path planning for uav aerial surveys in wind.
\newblock In {\em 2017 International Conference on Unmanned Aircraft Systems
  (ICUAS)}, pages 1563--1571. IEEE, 2017.

\bibitem{lewis2017semi}
Jeremy~S Lewis, William Edwards, Kelly Benson, Ioannis Rekleitis, and Jason~M
  O'Kane.
\newblock Semi-boustrophedon coverage with a dubins vehicle.
\newblock In {\em 2017 IEEE/RSJ International Conference on Intelligent Robots
  and Systems (IROS)}, pages 5630--5637. IEEE, 2017.

\bibitem{bahnemann2019revisiting}
Rik B{\"a}hnemann, Nicholas Lawrance, Jen~Jen Chung, Michael Pantic, Roland
  Siegwart, and Juan Nieto.
\newblock Revisiting boustrophedon coverage path planning as a generalized
  traveling salesman problem.
\newblock {\em arXiv preprint arXiv:1907.09224}, 2019.

\bibitem{di2016coverage}
Carmelo Di~Franco and Giorgio Buttazzo.
\newblock Coverage path planning for uavs photogrammetry with energy and
  resolution constraints.
\newblock {\em Journal of Intelligent \& Robotic Systems}, 83(3-4):445--462,
  2016.

\bibitem{maza2007multiple}
Ivan Maza and Anibal Ollero.
\newblock Multiple uav cooperative searching operation using polygon area
  decomposition and efficient coverage algorithms.
\newblock In {\em Distributed Autonomous Robotic Systems 6}, pages 221--230.
  Springer, 2007.

\bibitem{guruprasad2019x}
KR~Guruprasad.
\newblock X-stc: An extended spanning tree-based coverage algorithm for mobile
  robots.
\newblock In {\em Proceedings of the Advances in Robotics 2019}, pages 1--6.
  2019.

\bibitem{hazon2005redundancy}
Noam Hazon and Gal~A Kaminka.
\newblock Redundancy, efficiency and robustness in multi-robot coverage.
\newblock In {\em Proceedings of the 2005 IEEE International Conference on
  Robotics and Automation}, pages 735--741. IEEE, 2005.

\bibitem{agmon2006constructing}
Noa Agmon, Noam Hazon, and Gal~A Kaminka.
\newblock Constructing spanning trees for efficient multi-robot coverage.
\newblock In {\em Proceedings 2006 IEEE International Conference on Robotics
  and Automation, 2006. ICRA 2006.}, pages 1698--1703. IEEE, 2006.

\bibitem{zheng2005multi}
Xiaoming Zheng, Sonal Jain, Sven Koenig, and David Kempe.
\newblock Multi-robot forest coverage.
\newblock In {\em 2005 IEEE/RSJ International Conference on Intelligent Robots
  and Systems}, pages 3852--3857. IEEE, 2005.

\bibitem{kapoutsis2017darp}
Athanasios~Ch Kapoutsis, Savvas~A Chatzichristofis, and Elias~B Kosmatopoulos.
\newblock Darp: Divide areas algorithm for optimal multi-robot coverage path
  planning.
\newblock {\em Journal of Intelligent \& Robotic Systems}, 86(3-4):663--680,
  2017.

\bibitem{huang2020multi}
Xiang Huang, Min Sun, Hang Zhou, and Shuai Liu.
\newblock A multi-robot coverage path planning algorithm for the environment
  with multiple land cover types.
\newblock {\em IEEE Access}, 2020.

\bibitem{barrientos2011aerial}
Antonio Barrientos, Julian Colorado, Jaime~del Cerro, Alexander Martinez,
  Claudio Rossi, David Sanz, and Jo{\~a}o Valente.
\newblock Aerial remote sensing in agriculture: A practical approach to area
  coverage and path planning for fleets of mini aerial robots.
\newblock {\em Journal of Field Robotics}, 28(5):667--689, 2011.

\bibitem{shah2020multidrone}
Kunal Shah, Grant Ballard, Annie Schmidt, and Mac Schwager.
\newblock Multidrone aerial surveys of penguin colonies in antarctica.
\newblock {\em Science Robotics}, 5(47), 2020.

\bibitem{choset2001coverage}
Howie Choset.
\newblock Coverage for robotics--a survey of recent results.
\newblock {\em Annals of mathematics and artificial intelligence},
  31(1-4):113--126, 2001.

\bibitem{ghaemi2009evaluation}
Sina Ghaemi, Payam Rahimi, and David~S Nobes.
\newblock Evaluation of digital image discretization error in droplet shape
  measurement using simulation.
\newblock {\em Particle \& Particle Systems Characterization},
  26(5-6):243--255, 2009.

\bibitem{cai2011coordinate}
Guowei Cai, Ben~M Chen, and Tong~Heng Lee.
\newblock Coordinate systems and transformations.
\newblock In {\em Unmanned rotorcraft systems}, pages 23--34. Springer, 2011.

\bibitem{Kirkpatrick1983OptimizationBS}
Scott Kirkpatrick, C.~D. Gelatt, and Mario~P. Vecchi.
\newblock Optimization by simulated annealing.
\newblock {\em Science}, 220 4598:671--80, 1983.

\bibitem{bochkarev2016minimizing}
Stanislav Bochkarev and Stephen~L Smith.
\newblock On minimizing turns in robot coverage path planning.
\newblock In {\em 2016 IEEE International Conference on Automation Science and
  Engineering (CASE)}, pages 1237--1242. IEEE, 2016.

\bibitem{stolfi2020cooperative}
Daniel~H Stolfi, Matthias~R Brust, Gr{\'e}goire Danoy, and Pascal Bouvry.
\newblock A cooperative coevolutionary approach to maximise surveillance
  coverage of uav swarms.
\newblock In {\em 2020 IEEE 17th Annual Consumer Communications \& Networking
  Conference (CCNC)}, pages 1--6. IEEE, 2020.

\bibitem{paradzik2016multi}
Matej Paradzik and G{\"o}khan {\.I}nce.
\newblock Multi-agent search strategy based on digital pheromones for uavs.
\newblock In {\em 2016 24th Signal Processing and Communication Application
  Conference (SIU)}, pages 233--236. IEEE, 2016.

\bibitem{kapoutsis2015real}
Athanasios~Ch Kapoutsis, Savvas~A Chatzichristofis, Lefteris Doitsidis,
  Jo{\~a}o~Borges de~Sousa, Jose Pinto, Jose Braga, and Elias~B Kosmatopoulos.
\newblock Real-time adaptive multi-robot exploration with application to
  underwater map construction.
\newblock {\em Autonomous Robots}, page 1–29, 2015.

\bibitem{blum2021learning}
Avrim Blum, Chen Dan, and Saeed Seddighin.
\newblock Learning complexity of simulated annealing.
\newblock In {\em International Conference on Artificial Intelligence and
  Statistics}, pages 1540--1548. PMLR, 2021.

\bibitem{gomez2020mission}
Jos{\'e}~Miguel G{\'o}mez-L{\'o}pez, Jos{\'e}~Luis P{\'e}rez-Garc{\'\i}a,
  Antonio~Tom{\'a}s Mozas-Calvache, and Jorge Delgado-Garc{\'\i}a.
\newblock Mission flight planning of rpas for photogrammetric studies in
  complex scenes.
\newblock {\em ISPRS International Journal of Geo-Information}, 9(6):392, 2020.

\bibitem{hunt2018good}
E~Raymond Hunt~Jr and Craig~ST Daughtry.
\newblock What good are unmanned aircraft systems for agricultural remote
  sensing and precision agriculture?
\newblock {\em International journal of remote sensing}, 39(15-16):5345--5376,
  2018.

\bibitem{karatzinis2020towards}
Georgios~D Karatzinis, Savvas~D Apostolidis, Athanasios~Ch Kapoutsis, Liza
  Panagiotopoulou, Yiannis~S Boutalis, and Elias~B Kosmatopoulos.
\newblock Towards an integrated low-cost agricultural monitoring system with
  unmanned aircraft system.
\newblock In {\em 2020 International Conference on Unmanned Aircraft Systems
  (ICUAS)}, pages 1131--1138. IEEE, 2020.

\bibitem{bejiga2017convolutional}
Mesay~Belete Bejiga, Abdallah Zeggada, Abdelhamid Nouffidj, and Farid Melgani.
\newblock A convolutional neural network approach for assisting avalanche
  search and rescue operations with uav imagery.
\newblock {\em Remote Sensing}, 9(2):100, 2017.

\bibitem{rudol2008human}
Piotr Rudol and Patrick Doherty.
\newblock Human body detection and geolocalization for uav search and rescue
  missions using color and thermal imagery.
\newblock In {\em 2008 IEEE aerospace conference}, pages 1--8. Ieee, 2008.

\bibitem{redmon2018yolov3}
Joseph Redmon and Ali Farhadi.
\newblock Yolov3: An incremental improvement.
\newblock {\em arXiv preprint arXiv:1804.02767}, 2018.

\end{thebibliography}


\end{document}